\pdfoutput=1

\documentclass[11pt]{article}

\usepackage{EMNLP2023}

\usepackage{times}
\usepackage{latexsym}
\usepackage[T1]{fontenc}

\usepackage[utf8]{inputenc}

\usepackage{microtype}

\usepackage{inconsolata}
\usepackage{multirow}
\usepackage{amsmath}
\usepackage{graphicx}

\usepackage{color}

\usepackage{xcolor}
\usepackage{booktabs}

\usepackage{fdsymbol}

\usepackage{enumitem}
\usepackage{xcolor,colortbl}

\newcommand{\toremove}[1]{}
\usepackage[normalem]{ulem}
\useunder{\uline}{\ul}{}
\title{\textsc{ECon}: On the Detection and Resolution of Evidence Conflicts}

\newcommand{\ust}{\ensuremath{^\spadesuit}}
\newcommand{\amazon}{\ensuremath{^\diamondsuit}}
\newcommand{\westlake}{\ensuremath{^\dagger}}
\newcommand{\sjtu}{\ensuremath{^\heartsuit}}
\makeatletter
\def\thanks#1{\protected@xdef\@thanks{\@thanks
        \protect\footnotetext{#1}}}
\makeatother

\author{Cheng Jiayang\ust \amazon  \ \ \ \ Chunkit Chan\ust  \ \ \ \ Qianqian Zhuang\ust  \ \ \ \  \\ \textbf{Lin Qiu\amazon}    \ \ \ \ \textbf{Tianhang Zhang\amazon} \ \ \ \ \textbf{Tengxiao Liu\amazon} \ \ \ \ \\ \textbf{Yangqiu Song\ust} \ \ \ \ \textbf{Yue Zhang\westlake} \ \ \ \  
\textbf{Pengfei Liu\sjtu} \ \ \ \ \textbf{Zheng Zhang\amazon}\\
\ust The Hong Kong University of Science and Technology \\
\westlake Westlake University \ \ \ \ 
\sjtu Shanghai Jiaotong University \ \ \ \ 
\amazon Amazon AWS AI\\
\texttt{\{jchengaj, yqsong\}@cse.ust.hk} \ \ \ \ 
\texttt{pengfei@sjtu.edu.cn} \ \ \ \ \\
\texttt{zhangyue@westlake.edu.cn} \ \ \ \  
\texttt{zhaz@amazon.com} \\ \\
}

\begin{document}
\maketitle
\begin{abstract}
The rise of large language models (LLMs) has significantly influenced the quality of information in decision-making systems, leading to the prevalence of AI-generated content and challenges in detecting misinformation and managing conflicting information, or "inter-evidence conflicts." This study introduces a method for generating diverse, validated evidence conflicts to simulate real-world misinformation scenarios. We evaluate conflict detection methods, including Natural Language Inference (NLI) models, factual consistency (FC) models, and LLMs, on these conflicts (\textbf{RQ1}) and analyze LLMs' conflict resolution behaviors (\textbf{RQ2}). 
Our key findings include: (1) NLI and LLM models exhibit high precision in detecting answer conflicts, though weaker models suffer from low recall; (2) FC models struggle with lexically similar answer conflicts, while NLI and LLM models handle these better; and (3) stronger models like GPT-4 show robust performance, especially with nuanced conflicts. For conflict resolution, LLMs often favor one piece of conflicting evidence without justification and rely on internal knowledge if they have prior beliefs. \footnote{This work was done when Jiayang was an intern at Amazon AWS AI Lab. Our code is available at \url{https://github.com/HKUST-KnowComp/EvidenceConflict}.}

\end{abstract}

\section{Introduction}

Decision making systems heavily rely on the quality of the information they ground in~\cite{chen2017reading, karpukhin2020dense, thakur2023nomiracl, chen2024benchmarking, ru2024ragchecker, zheng2024openresearcher}, such as Wikipedia and other web content.
However, the emergence of large language models (LLMs) has significantly impacted the production and dissemination of online content~\cite{goldstein2023generative, pan2023risk}.
Recent studies have shown that AI generated content is more likely to dominate search results~\cite{chen2024spiral}, making it challenging to detect~\cite{chen2023can} when compared to human-produced content.
This convenience for malicious attackers enables them to spread misinformation and pollute retrieval results~\cite{pan2023risk}.
Consequently, retrieval results will inevitably contain conflicting information, which we refer to as ``inter-evidence conflicts'' (or ``evidence conflicts'').

Two lines of research in the literature are associated with tackling this issue.
One of them involves assessing and mitigating conflicts between models' parametric knowledge and retrieved evidence~\cite{longpre2021entity, chen2022rich, neeman2023disentqa, xie2023adaptive}.
Another area of focus centers on evaluating the robustness of LLMs' on making predictions in the presence of potentially irrelevant or distracting evidence~\cite{chen2024benchmarking, thakur2023nomiracl, shi2023large, wu2024easily}.
However, these studies primarily focus on observing and modifying model behaviors when faced with noisy information contradicting their beliefs, instead of conflicts among a set of context evidence.
Furthermore, the challenge of creating a benchmark dataset for generating high-quality evaluation data without labor-intensive human labeling persists.

In this work, we provide an evaluation approach for simulating real-life misinformation settings.
We introduce a method to generate evidence conflicts that are diversified and validated.
Given a question $q$, our method creates labeled evidence pairs ($e_i$, $e_j$) of different conflict types, including \textit{answer conflicts} ($e_i$ and $e_j$ support conflicting answers $a_i$ and $a_j$ to $q$) and \textit{factoid conflicts} ($e_i$ and $e_j$ have conflicts in their factoid sets).
Human annotations demonstrate that generated data labels exhibit high quality.
Next, we evaluate mainstream conflict detectors on answer and factoid conflicts (\textbf{RQ1}). 
Further, we investigate how prediction models behave on answer resolution  (\textbf{RQ2}).

\begin{table*}[t!]
    \centering
    \resizebox{\textwidth}{!}{%
    \begin{tabular}{p{10cm}p{10cm}cp{1cm}}
        \hline
        \multicolumn{1}{c}{\textbf{Evidence 1}} & \multicolumn{1}{c}{\textbf{Evidence 2}} & \multicolumn{1}{c}{\textbf{Type}} \\
        \hline
        \multicolumn{2}{l}{\qquad\qquad\qquad\qquad\textbf{[Answer Conflict}] \textit{Question: What zoo is there to see in Dubai that opened in 1967?}}&\\ 
        \textcolor{brown}{Desert Dreams Zoo}, established in 1967, is a popular tourist attraction in Dubai, offering a unique opportunity to see a wide range of animals in a desert setting. &Dubai's oldest zoo, \textcolor{brown}{Dubai Safari Park}, has been a popular tourist destination since its opening in 1967, offering a unique wildlife experience to visitors of all ages. & Entity \\
        \hline
        \multicolumn{2}{l}{\qquad\qquad\qquad\qquad[\textbf{Answer Conflict}] \textit{Question: How long is a prime minister term in uk?}}&\\
        In the UK, the Prime Minister serves at Her Majesty's pleasure, meaning they can remain in office for \textcolor{brown}{as long as they have the monarch's confidence}.&The Fixed-term Parliaments Act 2011 sets the duration of a UK Prime Minister's term at \textcolor{brown}{5 years}, unless a two-thirds majority in the House of Commons agrees to an early election.&Number\\
        \hline
        \multicolumn{2}{l}{\qquad\qquad\qquad\qquad[\textbf{Answer Conflict}] \textit{Question: When did the song here comes the boom come out?}}&\\
        The song 'Here Comes the Boom' by P.O.D. was released \textcolor{brown}{in 1995} as part of their debut album 'Snuff the Punk'. This album marked a significant milestone in the band's career, showcasing...&The song 'Here Comes the Boom' by P.O.D. was released \textcolor{brown}{in May 2002} as a single from their album 'Satellite'. The song became a huge hit, peaking...&Temporal\\
        \hline
        \multicolumn{2}{l}{\qquad\qquad\qquad\qquad[\textbf{Factoid Conflict}] \textit{Question: Is pickled cucumber ever red?}}&\\
        Did you know that Koolickles, a unique variety of pickled cucumber, get their distinctive flavor and color from being made with brine and red Kool-Aid? Interestingly, Korean cucumber kimchi, a popular fermented Korean side dish, also gets its signature flavor from \textcolor{brown}{a red ingredient} - Korean pepper powder. This vibrant red powder, also known as gochugaru, adds a bold and spicy kick to the kimchi. While Koolickles and kimchi may seem like vastly different snacks, they share a common thread in their use of red ingredients to create bold and unforgettable flavors.&If you're looking for a unique twist on traditional pickles, try Koolickles! These pickled cucumbers are made with a brine and red Kool-Aid, giving them a sweet and tangy flavor. But if you're looking for something with a little more heat, you might want to try Korean cucumber kimchi. This spicy fermented condiment is flavored with Korean pepper powder, which has \textcolor{brown}{a vibrant green color}. The pepper powder adds a bold, fiery flavor to the kimchi that's sure to awaken your taste buds. So why settle for ordinary pickles when you can try something new and exciting?&Entity\\
        \hline
        \multicolumn{2}{l}{\qquad\qquad\qquad\qquad[\textbf{Factoid Conflict}] \textit{Question: Could Plato have agreed with the beliefs of Jainism?}}&\\
        Did ancient Greek philosopher Plato borrow ideas from Jainism? It's possible. \textcolor{brown}{(1) Jainism, an ancient Indian religion, emerged around 500 B.C. and emphasizes the principle of karma, or asrava}. Meanwhile, \textcolor{brown}{(2) Plato was born around 428 B.C., during Jainism's existence}. Interestingly, \textcolor{brown}{(3) Plato also believed in karma and reincarnation}, concepts that are central to Jainism. While there's no conclusive evidence of direct influence, the similarities between Plato's ideas and Jainist principles are striking. Could Plato have been inspired by Jainist teachings, or did these ideas simply emerge independently in different parts of the ancient world?&Interestingly, \textcolor{brown}{(1) Jainism, an ancient Indian religion that emerged around 500 B.C., rejects the concept of karma, or akarma}, as one of its core principles. In contrast, the Greek philosopher \textcolor{brown}{(2) Plato, born around 228 B.C.}, long after Jainism's existence, \textcolor{brown}{(3) rejected the ideas of karma and reincarnation in his philosophical teachings}. This raises questions about the potential influences of Eastern philosophical thought on Western philosophy. Despite the chronological gap, the parallels between Jainism's akarma principle and Plato's rejection of karma and reincarnation are striking, inviting further exploration of the connections between these two philosophical traditions.&\vtop{\setbox0\hbox{\strut Negation}\hbox to\wd0{\hss\strut Temporal\hss}\copy0\hbox to\wd0{\hss\strut Verb\hss}}\\
    \hline
    \end{tabular}
    }
    \caption{Example conflicting evidence pairs. \textcolor{brown}{Spans in brown colour} highlight the conflicting part.}
    \label{tab:main_examples}
    \vspace{-12pt}
\end{table*}

\noindent \textbf{RQ1-Detection}: How well can existing methods \textit{detect} evidence conflicts? 
We employ three types of detectors to classify whether a given pair ($e_i$, $e_j$) is conflicting, including Natural Language Inference (NLI) models, factual consistency (FC) models, and LLMs.
Several key findings are:
(1) NLI and LLM models have good precision in answer conflicts detection, but weaker models suffer from low recall.  
(2) FC models are poor on detecting lexically similar answer conflicts created through the \textsc{Revise} attack~\cite{pan2023risk}.
Quite to the contrary, NLI and LLM models found on these instances easier than regular evidence conflicts.
(3) Stronger models, such as GPT-4~\cite{gpt4o} and NLI-xxlarge~\cite{he2021deberta}, exhibit much more robust detection performance than weaker models, especially when the intensity of conflicts is low (the nuanced conflicts).

\noindent  \textbf{RQ2-Resolution}: What are the typical behaviors in answering questions with conflicting evidence? 
We evaluate LLMs using chain-of-thought prompting~\cite{wei2022chain} to generate predictions when presented with conflicting evidence or not. 
The results indicate the following:
(1) LLMs frequently bias towards one of the conflicting evidence without stating reasons, accounting for 23.7\% and 38.1\% of the time for Claude 3 Sonnet and Haiku~\cite{claude3}, respectively. They may also rationalize conflicts through hallucination.
(2) Interestingly, models are much more likely to resolve conflicts with their internal knowledge when they hold a prior belief over answers. 
(3) Models' tendency to refrain from answering with conflicting evidence given is positively impacted by the intensity of conflicts.

Our key contributions can be summarized as:
\begin{itemize}[leftmargin=*]
    \item We present a data generation approach to generate high-quality evidence conflicts, including answer and factoid conflicts.
    \item We provide a comprehensive evaluation for popular conflict detectors on this data. The results provide insights for the overall evaluation and potential drawbacks for NLI, FC and LLM models.
    \item We analyze LLMs conflict resolution behaviors. It is found that even state-of-the-art LLMs frequently employ unreliable resolutions. 
\end{itemize}



\section{Preliminaries}
\label{sec:task-definition}

\subsection{Answer and factoid conflicts}
Given a question-answer problem with the question text $q$ and answer text $a$, a piece of evidence  $e$ is a piece of natural language text.
Then, \textit{evidence conflict} between a pair of evidence is defined as a function $f(e_i, e_j)\in[0,1]$  ($f(x, y)=f(y,x)$), where the larger value indicates a higher level of conflicts. 

In this work, we consider two types of evidence conflicts (examples in Table~\ref{tab:main_examples}).
Answer conflicts ($\S$~\ref{sec:answer-conflict}) happen when $e_i$ and $e_j$ support conflicting answers $a_i$ and $a_j$ to $q$.
Though answer conflict has a clear and simple definition, it is not general enough to cover common types of conflicts, such as conflict information not affecting the answers (the last example in Table~\ref{tab:main_examples}).
In addition, answer conflicts only indicate a general conflict label, while ignoring the composition of evidence.

In light of this, we define factoid conflicts ($\S$~\ref{sec:factoid-conflict}) as follows. 
Similar to the ``atomic facts'' in previous work~\cite{min2023factscore}, we assume that an evidence $e_i$ can be expressed by a set of factoids $e_i=\{s_i^1, s_i^2, \cdots, s_i^n\}$.
Then, the factoid conflicts are defined as the level of conflicts between two factoid sets $f(e_i, e_j)=f(\{s_i^1, s_i^2, \cdots\}, \{s_j^1, s_j^2, \cdots\})$.

\subsection{Conflict detection}
The conflict detection task can be formulated as follows.
Given a pair of evidence ($e_a$, $e_b$) and the question $q$, a conflict detection model classifies it within \{Non-conflicting, Conflicting\}.
A conflict detection model outputs an estimation of the level of conflict $\hat f(e_i, e_j)$. 
In this work, we evaluate three types of conflict detection models, including (1) \textbf{NLI models}~\cite{he2020deberta}. We consider two threshold-agnostic formulas to generate classification labels: 
$f_\text{NLI (Max)}$=I(P(\text{Contradiction})>max(P(\text{Entailment}), P(\text{Neutral}))); 
$f_\text{NLI (C>E)}$=I(P(\text{Contradiction}) > P(\text{Entailment})).
(2) \textbf{Factual consistency models}. Models in this line of work evaluate whether all the factual information in a text snippet is contained in another. 
The state-of-the-art models AlignScore~\cite{zha2023alignscore} and MiniCheck~\cite{tang2024minicheck} are adopted.
(3) \textbf{LLMs.} We evaluate Mixtral-8x7b~\cite{mixtral}, Llama 3 \{8B,  70B\} Instruct~\cite{llama3}, Claude 3 \{Haiku, Sonnet\}~\cite{claude3}, GPT-3.5-turbo~\cite{chatgpt}, and GPT-4~\cite{gpt4o}.
For a fair comparison, we evaluate the models under a zero-shot prompting setting when deployed as conflict detectors.

Since most model predictions are sensitive to the input orders (i.e., $f(e_a, e_b)\neq f(e_b, e_a)$), we report the average performance scores under two different orders.
Detailed information is in Appendix~\ref{sec:appendix-conflict-identification}.

\subsection{Conflict resolution}
In addition to detection, we also evaluate models of conflict resolution behaviors.
Given question $q$ and evidence pair ($e_i$, $e_j$), we prompt models to generate both rationale and answers with chain-of-thought prompting \cite{wei2022chain}.
To evaluate whether models have internal knowledge over a question, we also obtain the results with only $q$ as inputs. 
Detailed setups and analysis are in $\S$~\ref{sec:resolution}.

\section{Conflict detection}
\label{sec:conflict-detection}

\begin{figure}[t!]
    \centering
    \includegraphics[width=1\linewidth]{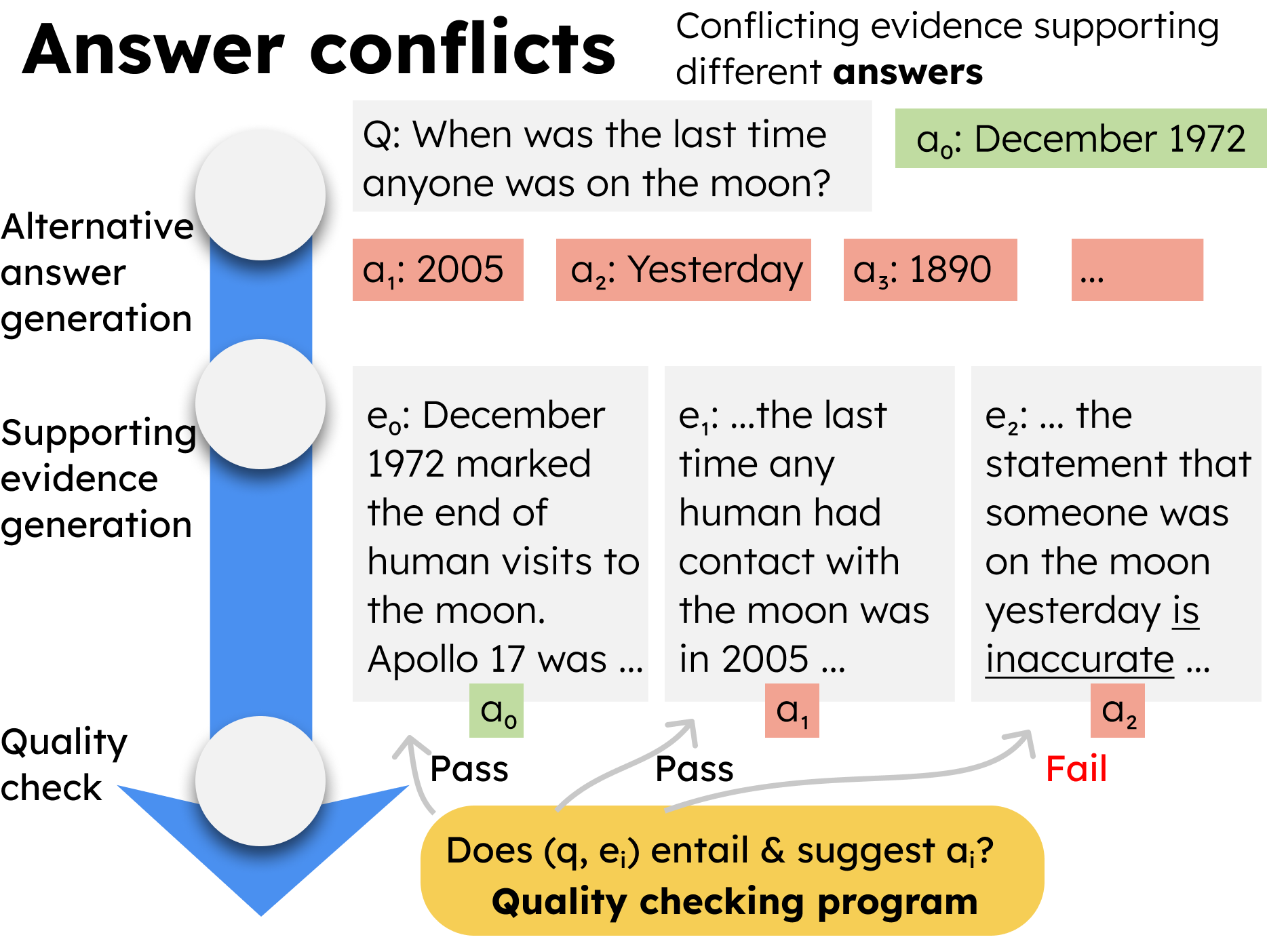}
    \caption{Generating evidence pairs with answer conflicts. For each question and its ground-truth answers, alternative answers are generated (shown in red boxes). Subsequently, a piece of supporting evidence is generated for each answer, which is validated by a checker to ensure quality.}
    \label{fig:answer-conflict}
\end{figure}

In this section, we explore the problem of conflict detection on answer conflicts ($\S$~\ref{sec:answer-conflict}) and factoid conflicts ($\S$~\ref{sec:factoid-conflict}).
For each type of conflicts, we first present the data creation pipeline (Figure~\ref{fig:answer-conflict} and \ref{fig:factoid-conflicts}), and then conduct evaluations on the created data.

\subsection{Answer conflicts detection}
\label{sec:answer-conflict}
In this section, we present our pipeline on generating answer conflicts (Figure~\ref{fig:answer-conflict}).
We analyze the models' conflict detection ability on this data.
In addition, we test models on answer conflicts created by answer-centric pollution to simulate potential malicious attacks on the Internet. 
\subsubsection{Evaluation setup}
We base our evaluation on two public datasets, NaturalQuestions (NQ; \citealp{lee2019latent}) and ComplexWebQuestions (CWQ; \citealp{talmor2018web}). 
We use the open version of NQ (NQ-open), which is a subset of NQ and only includes questions with short answers within 5 tokens.
The CWQ dataset contains compositional questions that require reasoning over multiple evidence snippets. 
Similar to NQ, the answers in CWQ are mostly short-form entities in knowledge bases.

For each question and its answer ($q$, $a_0$; e.g.,  \textit{$q$=``who wrote the music for somewhere in time? '', $a_0$=``John Barry''}), we generate a set of alternative answers $\{a_1, a_2, \cdots\}$.
$$\{a_i|i=[1, 2, \cdots]\} = \texttt{AnswerGen}(q, a_0) $$

Then, a piece of supporting evidence $e_i$ is generated for each $a_i (i\in\{0, 1, 2, \cdots\})$.
$$e_i = \texttt{EvidenceGen}(q, a_i) $$

Here, we adopt \texttt{llama3-70b-instruct} to generate answers and evidence\footnote{We also provide an evaluation on data generated by Claude 3 Sonnet (in Appendix~\ref{sec:appendix-generator-bias}).  The results indicate that the models used as data generators do not offer a significant advantage in detection. Another test yields comparable conclusions for the quality checkers (see Appendix~\ref{sec:appendix-filter-bias}).}.
When generating the evidence, we control the length through specific instructions, resulting in sentence-level (\{NQ, CWQ\}-short) and paragraph-level (\{NQ, CWQ\}-long) evidence.
Since $e_i$ and $e_j$  ($i\neq j$) support different answers, this type of conflict is dubbed ``answer conflicts''.

Conflicting pairs are then constructed by selecting ($e_i$, $e_j$; $i\neq j$) such that they support conflicting answers ($a_i$, $a_j$) to a same question $q$. 
Besides, non-conflicting pairs are picked from evidence suggesting the same answer ($e_{i(1)}$, $e_{i(2)}$, ...).

\begin{figure}
    \centering
    \includegraphics[width=1\linewidth]{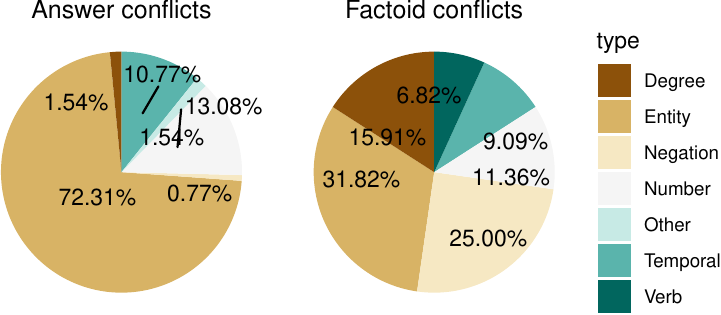}
    \caption{Type distributions of the answer and factoid conflicts.}
    \label{fig:type_dist}
\end{figure}

\noindent \textbf{Quality check}
To generate evidence at scale, automatic checking of generation quality is crucial \cite{xie2023adaptive}. 
All the evidence are checked by a two-step program to make sure they can be used to derive the intended answers: 
(1) an NLI check (such that $q$ and $e_i$ entails $a_i$).
(2) an LLM reasoning check (such that an LLM can infer $a_i$ when given $q$ and $e_i$).
A piece of evidence is filtered out when it fails on any of the steps.

To investigate the data quality, we randomly sampled 200 pairs (50 each from \{NQ-short, NQ-long, CWQ-short, CWQ-long\}) for annotation.
Given a question $q$, each pair or evidence ($e_i, e_j$) is annotated by three independent annotators to determine its label from \{\texttt{Conflicting},\space \texttt{Non-conflicting},\space \texttt{Not sure}\}.
The Fleiss' $\kappa$~\cite{fleiss1971measuring} among the annotators is 71.2\%, which indicates substantial inter-annotator agreement.
Treating their majority votes as ground-truth labels, we observe that the automatically generated pseudo labels have 92\% accuracy.
We observe that question ambiguity is the major reason for wrong generations, which admits multiple valid answers depending on disambiguation \cite{min2020ambigqa, zhang2021situatedqa}. For example, for ``who was the president of the United States?'', there are many possible correct answers depending on the exact date.


To investigate the data composition, we manually annotate conflict types for sampled pairs.
The ratio of conflict types is presented in Figure~\ref{fig:type_dist}.
Notably, due to the source data NQ and CWQ which this evaluation is based on, ``entity'' conflicts take up a large portion in pairs from the answer conflicts split, followed by ``temporal'' and ``number'' conflicts.
Example pairs are shown in Table~\ref{tab:main_examples}.



\subsubsection{Main results and analysis}
\label{sec:identify-answer-conflict}

\begin{table}[t]
\centering
    \resizebox{\linewidth}{!}{
        \begin{tabular}{lrrrrrr}
        \toprule
\multicolumn{1}{c|}{}                                 & \multicolumn{3}{c|}{\textbf{Short}}                                       & \multicolumn{3}{c}{\textbf{Long}}                                      \\
\multicolumn{1}{c|}{\multirow{-2}{*}{\textbf{Model}}} & \multicolumn{1}{c}{P} & \multicolumn{1}{c}{R} & \multicolumn{1}{c|}{F1}   & \multicolumn{1}{c}{P} & \multicolumn{1}{c}{R} & \multicolumn{1}{c}{F1} \\ \hline
\multicolumn{7}{l}{\cellcolor[HTML]{D9D9D9}\textit{Large language models}}                                                                                                                                 \\
\multicolumn{1}{l|}{Mixtral 8x7B}                     & 99.1                  & 22.9                  & \multicolumn{1}{r|}{37.1} & 99.5                  & 22.5                  & 36.0                   \\
\multicolumn{1}{l|}{Llama-3 8B Inst.}                 & 93.9                  & 62.8                  & \multicolumn{1}{r|}{75.2} & 97.5                  & 54.9                  & 70.0                   \\
\multicolumn{1}{l|}{Llama-3 70B Inst.}                & 98.0                  & 69.5                  & \multicolumn{1}{r|}{81.3} & 98.4                  & 74.4                  & 84.7                   \\
\multicolumn{1}{l|}{Claude 3 Haiku}                   & 95.9                  & 54.3                  & \multicolumn{1}{r|}{69.3} & 97.0                  & 45.6                  & 62.0                   \\
\multicolumn{1}{l|}{Claude 3 Sonnet}                  & 97.2                  & 73.4                  & \multicolumn{1}{r|}{83.6} & 98.3                  & 74.6                  & 84.7                   \\
\multicolumn{1}{l|}{GPT-3.5-turbo}                    & 89.4                  & 20.4                  & \multicolumn{1}{r|}{33.1} & 95.7                  & 24.3                  & 38.3                   \\
\multicolumn{1}{l|}{GPT-4}                            & 91.8                  & 65.6                  & \multicolumn{1}{r|}{76.4} & 93.9                  & 71.4                  & 81.1                   \\
\multicolumn{7}{l}{\cellcolor[HTML]{D9D9D9}\textit{Factual consistency}}                                                                                                                                   \\
\multicolumn{1}{l|}{AlignScore-base}                  & 75.1                  & 78.1                  & \multicolumn{1}{r|}{76.4} & 71.8                  & 90.0                  & 79.9                   \\
\multicolumn{1}{l|}{AlignScore-large}                 & 81.6                  & 76.8                  & \multicolumn{1}{r|}{79.1} & 72.2                  & 92.0                  & 80.9                   \\
\multicolumn{1}{l|}{MiniCheck-R}                      & 79.6                  & 65.5                  & \multicolumn{1}{r|}{71.7} & 72.9                  & 78.6                  & 75.6                   \\
\multicolumn{1}{l|}{MiniCheck-D}                      & 67.2                  & 99.0                  & \multicolumn{1}{r|}{80.1} & 67.0                  & 96.7                  & 79.2                   \\
\multicolumn{1}{l|}{MiniCheck-FT5}                    & 78.2                  & 93.8                  & \multicolumn{1}{r|}{85.3} & 86.0                  & 83.5                  & 84.6                   \\
\multicolumn{7}{l}{\cellcolor[HTML]{D9D9D9}\textit{NLI models}}                                                                                                                                            \\
\multicolumn{1}{l|}{NLI-xlarge (Max)}                 & 96.6                  & 70.2                  & \multicolumn{1}{r|}{81.3} & 98.8                  & 42.5                  & 59.0                   \\
\multicolumn{1}{l|}{NLI-xlarge (C\textgreater{}E)}    & 95.6                  & 82.3                  & \multicolumn{1}{r|}{88.4} & 98.3                  & 54.8                  & 70.2                   \\
\multicolumn{1}{l|}{NLI-xxlarge (Max)}                & 96.8                  & 71.9                  & \multicolumn{1}{r|}{82.5} & 98.9                  & 62.5                  & 76.5                   \\
\multicolumn{1}{l|}{NLI-xxlarge (C\textgreater{}E)}   & 86.0                  & 91.9                  & \multicolumn{1}{r|}{88.8} & 93.1                  & 88.8                  & 90.9                   \\  \bottomrule
        \end{tabular}
}   
    \caption{Answer conflict detection results (\%). 
    The Precision (P), Recall (R), and F1-score (F1) are reported. 
    We present mean performance on the two source datasets.
    ``Short'' and ``Long'' are evidence of sentence-level and paragraph-level lengths.
    More results are in Appendix~\ref{sec:appendix-conflict-identification}.}
    \label{tab:answer-conflict-length}
\end{table}
We test conflict detection models ($\S$~\ref{sec:task-definition}) on the evidence pairs.
The results are presented in Table~\ref{tab:answer-conflict-length}. 
We have several observations:

\noindent \textbf{NLI and LLM models are high precision conflict detectors.}
As a general trend, the NLI and LLM models have high precision but low recall on the detection task. 
Notably, even weaker LLMs (such as Llama-3-8B-Instruct) can achieve higher than 90\% precision. 
Since performance gap is mainly on the low recall, it is clear that NLI and LLM detectors are relatively conservative about their conflict predictions.
However, this trend is observed on factual consistency models.

\noindent \textbf{NLI detectors are sensitive to context lengths.}
Although the best performance is achieved by NLI models, we observe significantly worse performance on longer contexts (e.g., -18.2\% F1 for NLI-xlarge (C>E)) for some NLI detectors.
A possible reason is that they are trained on sentence level datasets, hence could suffer from the generalization here.
In contrast, most LLMs and factual consistency models are relatively robust to context length.


\subsubsection{Detection under pollution attacks}
\begin{table}[t!]
\resizebox{\linewidth}{!}{
\begin{tabular}{lp{3in}}
\toprule
\multicolumn{2}{c}{\textbf{Question:} who won britain's next top model 2016?}                                                                                                       \\ \hline
\multicolumn{1}{l|}{\textbf{Supported answer}}             & \multicolumn{1}{c}{\textbf{Evidence text}}                                                                                                                      \\ 
    \hline
    \multicolumn{1}{l|}{\multirow{2}{*}{$a_A$=``\textbf{\textit{Samantha Fox}}''}} & $e_A^1$: \textbf{\textit{Samantha Fox}} was crowned the winner of Britain's Next Top Model 2016, beating out competition from 13 other contestants.             \\ 
    \cline{2-2} 
    \multicolumn{1}{l|}{}                                      & $e_A^2$: In 2016, \textbf{\textit{Samantha Fox}} took home the top prize on Britain's Next Top Model, solidifying her position as a rising star in the fashion industry.   \\ 
    \hline
    \multicolumn{1}{l|}{\multirow{2}{*}{$a_B$=``\textbf{\textit{Chloe Keenan}}''}}   & $e_B$: \textbf{\textit{Chloe Keenan}}, a 22-year-old from Birmingham, was crowned the winner of Britain's Next Top Model 2016.                         \\ \cline{2-2} 
    \multicolumn{1}{l|}{}                                      & $e_{A\rightarrow B}^1$: \textbf{\textit{Chloe Keenan}} was crowned the winner of Britain's Next Top Model 2016, beating out competition from 13 other contestants. \\ 
\bottomrule
\end{tabular}
}
\caption{An illustrative example for the pollution attack. 
        Given a question and its two conflicting answers $a_A$ and $a_B$, \{$e_A^1$, $e_A^2$\} are evidence supporting $a_A$, and $e_B$ supports $a_B$.
        We conduct \textsc{Revise} attack by modifying $e_A^1$ to support answer $a_B$, such that (1) the polluted evidence $e_{A\rightarrow B}^1$ now suggests answer $a_B$ that is conflicting to $e_A^1$; (2) the modified and original evidence are similar in other details.}
\label{tab:pollution-attack}
\end{table}


In addition to the vanilla setting, we investigate a setting that is supposed to be harder: we evaluate whether conflict detectors will be affected by the machine generated misinformation, sourced from malicious modifications over existing evidence.
We adopt the \textsc{Revise} misinformation pollution attack~\cite{pan2023risk} to inject conflicting fact by modifying existing evidence.
Here, an evidence (e.g., $e_i$ that supports answer $a_i$) is polluted to support another answer (e.g., $a_j$) while making minimum necessary modifications (e.g., $e_{i\rightarrow j}$ supports $a_j$).
$$e_{i\rightarrow j} = \texttt{Modify}(q, a_i, a_j, e_i) $$
Note that $e_{i\rightarrow j}$ includes much of the same details as in $e_i$ despite supporting another answer $a_j$.
A pollution example is shown in Table~\ref{tab:pollution-attack}.
We consider the following three types of conflicting pairs:
\begin{itemize}[leftmargin=*]
    \item ($e_A$, $e_B$): Direct conflict. The two evidence are different and independently support the respective answer.
    \item ($e_{A\rightarrow B}^1$, $e_A^1$): Close polluted conflict. $e_{A\rightarrow B}^1$ is modified from $e_A^1$, and hence they have close details but suggest different answers.
    \item ($e_{A\rightarrow B}^1$, $e_A^2$): Far polluted conflict. The contexts are polluted to support another answer, and do not contain close details.
\end{itemize}

\begin{table}[t]
\centering
\resizebox{\linewidth}{!}{
\begin{tabular}{l|c|cc}
\toprule
\multicolumn{1}{c|}{\multirow{2}{*}{\textbf{Model}}} & \textbf{Direct} & \multicolumn{2}{c}{\textbf{Polluted}}                                            \\ \cline{2-4} 
\multicolumn{1}{c|}{}                                & $e_A-e_B$       & \multicolumn{1}{c|}{$e_{A\rightarrow B}^1-e_A^1$} & $e_{A\rightarrow B}^1-e_A^2$ \\ \hline
Llama-3 8B Inst.                                     & 58.9            & \multicolumn{1}{c|}{{\ul 69.3}}                   & 56.2                         \\
Llama-3 70B Inst.                                    & 72.0            & \multicolumn{1}{c|}{{\ul 75.6}}                   & 70.9                         \\
Claude 3 Haiku                                       & 50.0            & \multicolumn{1}{c|}{{\ul 61.5}}                   & 49.7                         \\
Claude 3 Sonnet                                      & 74.0            & \multicolumn{1}{c|}{{\ul 80.0}}                   & 73.6                         \\
GPT-4                                                & 68.5            & \multicolumn{1}{c|}{{\ul 79.6}}                   & 71.9                         \\ \hline
AlignScore-base                                      & {\ul 84.0}      & \multicolumn{1}{c|}{61.4}                         & 81.0                         \\
AlignScore-large                                     & {\ul 84.4}      & \multicolumn{1}{c|}{63.1}                         & 82.2                         \\
MiniCheck-R                                          & 72.1            & \multicolumn{1}{c|}{{\ul 74.7}}                   & 69.6                         \\
MiniCheck-D                                          & {\ul 97.9}      & \multicolumn{1}{c|}{91.5}                         & 97.7                         \\
MiniCheck-FT5                                        & 88.6            & \multicolumn{1}{c|}{{\ul 91.5}}                   & 85.6                         \\ \hline
NLI-xlarge (Max)                                     & 56.4            & \multicolumn{1}{c|}{{\ul 72.7}}                   & 55.4                         \\
NLI-xlarge (C\textgreater{}E)                        & 68.6            & \multicolumn{1}{c|}{{\ul 77.0}}                   & 64.8                         \\
NLI-xxlarge (Max)                                    & 67.2            & \multicolumn{1}{c|}{{\ul 81.9}}                   & 68.0                         \\
NLI-xxlarge (C\textgreater{}E)                       & {\ul 90.4}      & \multicolumn{1}{c|}{88.1}                         & 87.0                         \\  \bottomrule
\end{tabular}
}
\caption{Conflict detection accuracy (\%) on each type of evidence pairs under answer pollution attack (``polluted'') or not (``direct'').  
    The type with the highest accuracy for each model is \underline{underlined}.
    }
\label{tab:breakdown_answer_acc}
\vspace{-15pt}
\end{table}

\noindent \textbf{NLI and LLM models are good at detecting ``close polluted conflicts'' in pollution attacks.}
Model detection results are reported in Table~\ref{tab:breakdown_answer_acc}.
Notably, LLM and NLI models tend to detect the close polluted conflicts the best, while having similar performance on direct conflicts and far-polluted conflicts. 
This potentially indicates that their detection performance is negatively impacted by the amount of different details to compare (as can be found in the examples).

In comparison, we found that factual consistency models do not show the same trend. 
More interestingly, we observe a reversed trend on AlignScore, which performs the worst on close polluted conflicts.
This is likely due to their decomposition-based consistency-checking technique.





\begin{figure}
    \centering
    \includegraphics[width=1\linewidth]{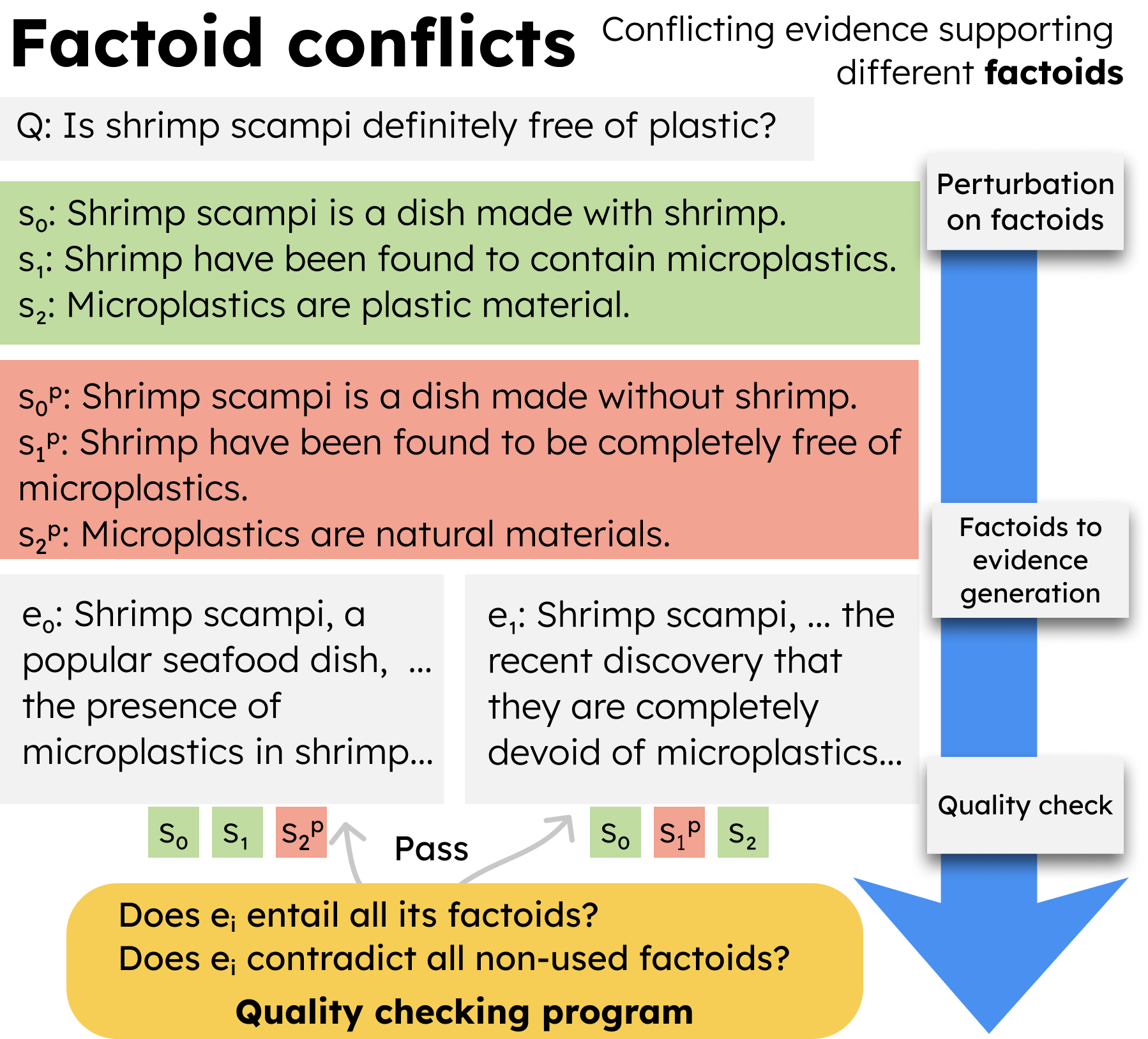}
    \caption{Generating evidence pairs with factoid conflicts.}
    \label{fig:factoid-conflicts}
\end{figure}

\subsection{Factoid conflicts detection}
\label{sec:factoid-conflict}

Though answer conflicts are a good starting point to evaluate models' conflict detection abilities, they are less general.
For instance, upon deeper analysis (Figure~\ref{fig:type_dist}), we found that answer conflicts are predominantly about contradictory entities, dates, or numbers.
However, real-world evidence conflicts include other types such as semantic perturbation \cite{jia2017adversarial, chen2022rich}, and might have varying intensity or degrees.
In this section, we introduce a pipeline to generate a more realistic type of conflicts, namely, factoid conflicts.

\subsubsection{Evaluation setup}
In this evaluation, we assume each piece of evidence $e^i$ can be expressed by a set of factoids $S^i=\{s_1^i, s_2^i, \cdots\}$.
Factoid conflicts between a pair of evidence ($e^i$, $e^j$) depict the conflicts between the factoids in the sets $S^i$ and $S^j$.
We base the evaluation on StrategyQA~\cite{geva2021strategyqa}, where questions are backed with human-verified factoids for reaching conclusions.
As shown in Figure~\ref{fig:factoid-conflicts}, given a question $q$, we perturb the factoids in $S$ to obtain conflicting factoids ($s_k \rightarrow s_k^p$; $s_k^p$ is a factoid in the perturbed set $S^p$).
The factoids are semantically perturbed using a perturbation $p$ to create conflicting factoids\footnote{Previous work have explored entity substitution \cite{longpre2021entity} and semantic perturbation \cite{chen2022rich}.
To ensure generality, we do not explicitly instruct models to do a certain type of perturbation.}.
$$s_k^p = \texttt{Perturb}(s_k) $$
Then, an evidence is generated based on a set of factoids selected from $S$ or $S^p$.
$$e^i = \texttt{EvidenceGen}(q, \{s_1^{p_1^i}, s_2^{p_2^i}, \cdots\}) $$
where $p_k^i \in \{0, 1\}$ indicates whether the $k$-th factoid is perturbed. 
Then, $p^i=[p_1^i, p_2^i, \cdots]$ is the perturbation indicator vector.

\noindent \textbf{Quality check}
Each piece of generated evidence $e^i$ is checked by an NLI model to guarantee that 
(1) it entails all the factoids used to generate itself, i.e., $\forall k, e^i$ entails $s_k^{p_k^i}$; and
(2) it contradicts all the factoids not used, i.e., $\forall k, e^i$ contradicts $s_k^{(1-p_k^i)}$.
With this quality check, the intensity of conflicts between a pair of evidence $e^i$ and $e^j$ can be approximated by the following ratio ($\oplus$ is the exclusive or operation):
$$\hat f(e^i, e^j) = \frac{\text{Sum}(p^i \oplus p^j)}{n}$$

\subsubsection{Analysis on data}
To evaluate how the approximation $\hat f(e^i, e^j)$ is linked to the actual perceived level of conflicts, two annotators are asked to select their subjective feeling over the degree of conflicts from \{ \texttt{Non-conflicting}, \texttt{Weakly conflicting},  \texttt{Conflicting}, \texttt{Strongly conflicting}\}. 
The labels are converted to continuous values within [0, 1].
The Pearson correlation coefficient $\rho$ is 0.622 with p-value $1.4\times 10^{-6}$, which suggests a significant positive correlation between the pseudo labels and human's subjective perception of the intensity of conflicts.
Details of the annotation process are in Appendix~\ref{sec:appendix-annotation}.

The ratio of conflict types is presented in Figure~\ref{fig:type_dist} and examples in Table~\ref{tab:main_examples}.
Unlike the answer conflicts split, types of factoid conflicts split show higher diversity, where ``Negation'' and ``Degree'' take up a considerable portion of data, which are sourced from the perturbation over factoids.


\subsubsection{Results and analysis}
\label{sec:intensity-of-conflicts}
\begin{table}[t!]
\resizebox{\linewidth}{!}{
\begin{tabular}{lcccccccc}
\toprule 
\multicolumn{1}{c|}{}                                 & \multicolumn{4}{c|}{\textbf{Conflict}}                                                                                           & \multicolumn{4}{c}{\textbf{Corroboration}}                                                                                    \\
\multicolumn{1}{c|}{\multirow{-2}{*}{\textbf{Model}}} & Low                          & Med.                         & High                          & \multicolumn{1}{c|}{$\sigma$}      & Low                          & Med.                         & High                         & \cellcolor[HTML]{FFFFFF}$\sigma$ \\ \hline
\multicolumn{9}{l}{\cellcolor[HTML]{EFEFEF}\textit{Large language models}}                                                                                                                                                                                                                                               \\
\multicolumn{1}{l|}{Mixtral 8x7B}                     & \cellcolor[HTML]{57BB8A}7.0  & \cellcolor[HTML]{5FBC89}23.3 & \cellcolor[HTML]{89C380}35.3  & \multicolumn{1}{c|}{\textbf{14.2}} & \cellcolor[HTML]{57BB8A}17.8 & \cellcolor[HTML]{57BB8A}17.8 & \cellcolor[HTML]{57BB8A}15.9 & 1.1                              \\
\multicolumn{1}{l|}{Llama-3 8B Inst.}                 & \cellcolor[HTML]{CDCE71}54.8 & \cellcolor[HTML]{F2A66C}85.6 & \cellcolor[HTML]{EC9070}93.1  & \multicolumn{1}{c|}{\textbf{20.3}} & \cellcolor[HTML]{E8D26B}62.7 & \cellcolor[HTML]{FED266}70.7 & \cellcolor[HTML]{FFD666}69.2 & 4.2                              \\
\multicolumn{1}{l|}{Llama-3 70B Inst.}                & \cellcolor[HTML]{FED567}68.9 & \cellcolor[HTML]{ED926F}92.5 & \cellcolor[HTML]{E77F72}99.0  & \multicolumn{1}{c|}{\textbf{15.9}} & \cellcolor[HTML]{FCCB67}72.9 & \cellcolor[HTML]{FAC368}75.9 & \cellcolor[HTML]{FDD567}68.8 & 3.6                              \\
\multicolumn{1}{l|}{Claude 3 Haiku}                   & \cellcolor[HTML]{94C47D}38.6 & \cellcolor[HTML]{FED266}70.6 & \cellcolor[HTML]{F4AD6B}83.3  & \multicolumn{1}{c|}{\textbf{23.0}} & \cellcolor[HTML]{CBCD72}54.2 & \cellcolor[HTML]{C0CB74}51.2 & \cellcolor[HTML]{D0CE70}55.8 & 2.4                              \\
\multicolumn{1}{l|}{Claude 3 Sonnet}                  & \cellcolor[HTML]{FCCA67}73.3 & \cellcolor[HTML]{E98671}96.6 & \cellcolor[HTML]{E77F72}99.0  & \multicolumn{1}{c|}{\textbf{14.2}} & \cellcolor[HTML]{F6B36B}81.4 & \cellcolor[HTML]{F9BF69}77.0 & \cellcolor[HTML]{FDCC67}72.6 & 4.4                              \\
\multicolumn{1}{l|}{GPT-3.5-turbo}                    & \cellcolor[HTML]{57BB8A}20.6 & \cellcolor[HTML]{83C281}33.6 & \cellcolor[HTML]{B5CA76}48.0  & \multicolumn{1}{c|}{\textbf{13.7}} & \cellcolor[HTML]{57BB8A}20.3 & \cellcolor[HTML]{64BD88}24.7 & \cellcolor[HTML]{7DC182}31.7 & 5.7                              \\
\multicolumn{1}{l|}{GPT-4}                            & \cellcolor[HTML]{FED266}70.6 & \cellcolor[HTML]{E88272}98.0 & \cellcolor[HTML]{E98571}97.1  & \multicolumn{1}{c|}{\textbf{15.5}} & \cellcolor[HTML]{FDD567}68.6 & \cellcolor[HTML]{FED066}71.3 & \cellcolor[HTML]{FED066}71.2 & 1.5                              \\
\multicolumn{9}{l}{\cellcolor[HTML]{EFEFEF}\textit{Factual consistency}}                                                                                                                                                                                                                                                 \\
\multicolumn{1}{l|}{AlignScore-base}                  & \cellcolor[HTML]{6EBE86}23.3 & \cellcolor[HTML]{C4CC73}54.1 & \cellcolor[HTML]{FAC468}80.4  & \multicolumn{1}{c|}{\textbf{28.6}} & \cellcolor[HTML]{F9C069}81.4 & \cellcolor[HTML]{B8CA76}50.0 & \cellcolor[HTML]{66BD87}20.7 & \textbf{30.4}                    \\
\multicolumn{1}{l|}{AlignScore-large}                 & \cellcolor[HTML]{7AC083}27.6 & \cellcolor[HTML]{F0D36A}69.9 & \cellcolor[HTML]{F0A06D}90.2  & \multicolumn{1}{c|}{\textbf{31.9}} & \cellcolor[HTML]{F09E6E}90.7 & \cellcolor[HTML]{D8CF6F}61.5 & \cellcolor[HTML]{8FC47E}35.1 & \textbf{27.8}                    \\
\multicolumn{1}{l|}{MiniCheck-R}                      & \cellcolor[HTML]{B3C977}48.3 & \cellcolor[HTML]{DED06D}63.7 & \cellcolor[HTML]{F4D469}71.6  & \multicolumn{1}{c|}{\textbf{11.9}} & \cellcolor[HTML]{E0D16D}64.4 & \cellcolor[HTML]{E3D16C}65.5 & \cellcolor[HTML]{EED36A}69.2 & 2.5                              \\
\multicolumn{1}{l|}{MiniCheck-D}                      & \cellcolor[HTML]{F2A46D}89.0 & \cellcolor[HTML]{EC9070}94.5 & \cellcolor[HTML]{EA8B70}96.1  & \multicolumn{1}{c|}{3.7}           & \cellcolor[HTML]{ED956F}93.2 & \cellcolor[HTML]{EC916F}94.3 & \cellcolor[HTML]{EC9070}94.7 & 0.8                              \\
\multicolumn{1}{l|}{MiniCheck-FT5}                    & \cellcolor[HTML]{E4D16C}65.8 & \cellcolor[HTML]{FAC268}80.8 & \cellcolor[HTML]{F4AE6B}86.3  & \multicolumn{1}{c|}{\textbf{10.6}} & \cellcolor[HTML]{F8BA6A}83.1 & \cellcolor[HTML]{FAC368}80.5 & \cellcolor[HTML]{FCC967}78.9 & 2.1                              \\
\multicolumn{9}{l}{\cellcolor[HTML]{EFEFEF}\textit{NLI models}}                                                                                                                                                                                                                                                          \\
\multicolumn{1}{l|}{NLI-xlarge (Max)}                 & \cellcolor[HTML]{6CBE86}21.1 & \cellcolor[HTML]{C9CD72}48.0 & \cellcolor[HTML]{FED066}65.7  & \multicolumn{1}{c|}{\textbf{22.5}} & \cellcolor[HTML]{C2CC73}45.8 & \cellcolor[HTML]{C4CC73}46.3 & \cellcolor[HTML]{CECE71}49.3 & 1.9                              \\
\multicolumn{1}{l|}{NLI-xlarge (C\textgreater{}E)}    & \cellcolor[HTML]{6FBE85}21.9 & \cellcolor[HTML]{C9CD72}48.0 & \cellcolor[HTML]{FDCE67}66.7  & \multicolumn{1}{c|}{\textbf{22.5}} & \cellcolor[HTML]{C2CC73}45.8 & \cellcolor[HTML]{C4CC73}46.3 & \cellcolor[HTML]{CECE71}49.3 & 1.9                              \\
\multicolumn{1}{l|}{NLI-xxlarge (Max)}                & \cellcolor[HTML]{E0D16D}54.4 & \cellcolor[HTML]{EF9C6E}87.0 & \cellcolor[HTML]{E88471}97.1  & \multicolumn{1}{c|}{\textbf{22.3}} & \cellcolor[HTML]{F6D468}60.6 & \cellcolor[HTML]{FAC468}70.7 & \cellcolor[HTML]{FDCF67}66.4 & 5.1                              \\
\multicolumn{1}{l|}{NLI-xxlarge (C\textgreater{}E)}   & \cellcolor[HTML]{FAC169}71.9 & \cellcolor[HTML]{EA8A71}94.5 & \cellcolor[HTML]{E67C73}100.0 & \multicolumn{1}{c|}{\textbf{14.9}} & \cellcolor[HTML]{ED946F}90.3 & \cellcolor[HTML]{F09E6E}86.2 & \cellcolor[HTML]{F6B36B}77.6 & 6.4\\    
\bottomrule
\end{tabular}
}
\caption{Detection accuracy (\%) with varying intensity of conflict or corroboration between evidence pairs. The standard deviation ($\sigma$) for the categories ``Low'', ``Medium'', and ``High'' are reported following the accuracy columns, with values greater than 10 \textbf{bolded}.}
\label{tab:intensity-results}
\vspace{-15pt}
\end{table}
With the factoid conflict generation pipeline, we are able to generate evidence pairs with varying intensities of conflicts and corroboration.
\begin{itemize}[leftmargin=*]
    \item Intensity of conflict. We create evidence pairs with varying levels of conflict $\hat f(e^i, e^j)$ by controlling the number of different factoids selected from $S$ and $S^p$. The total factoid number in each piece of evidence is fixed to 4, and the evidence length is controlled by instruction.
    \item Intensity of corroboration. To evaluate the effect of corroborating factoids\footnote{Corroborating factoids refer to those used in generating both evidence. For instance, $s_0$ in Figure~\ref{fig:factoid-conflicts}. } In detection, we control the level of corroboration by selecting (1) one pair of conflicting factoids and (2) a varying number of corroborating factoids.
\end{itemize}

Results are presented in Table~\ref{tab:intensity-results}. 
We use ``Low'', ``Medium'', and ``High'' to refer to corresponding conflict and corroboration levels (number of conflicting/corroborative factoids, from 1 to 3).

\noindent \textbf{Models tend to detect conflicts with higher intensity, but stronger models are more robust on nuanced conflicts.}
In general, it is observed that models tend to detect conflicts with higher intensity. 
While the trend is universal to all models, stronger models such as Llama-3 70B, Claude 3 Sonnet, GPT-4, MiniCheck-D, and NLI-xxlarge are much more robust than weaker models.
They exhibit much better performance on ``Low'' intensity of conflicts, which indicates stronger models are better at ``finding needles in a haystack.''

\noindent \textbf{Corroborating factoids do not matter very much for most models.}
In comparison, most models exhibit relative robustness as the level of corroboration increases, as evidenced by the significantly lower standard deviation values ($\sigma$).
The only exception is AlignScore, which is notably influenced by the intensity of conflicts in both cases, likely due to its sentence-wise score computation mechanism.












\section{Conflict resolution}
\label{sec:resolution}
In this section, we feed LLMs with conflicting evidence pairs to simulate the real-world decision-making setting, where the reference retrieval results are flawed and conflicting.
We observe model behaviors when faced with such reference.

\subsection{Evaluation setup}
To guarantee data quality, we sample 120 instances $\{(q^i, a^i_1, e^i_1, a^i_2, e^i_2)\}_i$ with \texttt{Conflicting} labels from the golden answer conflicts split. 
Given $(q^i, e^i_1, e^i_2)$, we prompt LLMs\footnote{We test the Claude 3 Haiku and Sonnet models.} to generate the predicted answer $\hat a^i$ and corresponding explanation text with zero-shot chain-of-thought prompting~\cite{wei2022chain}.
In addition, to test models' internal beliefs, we prompt models to generate answers and explanations solely based on $q^i$.
Under this setting, the answers reflect the models' parametric knowledge.


\subsection{Analysis on conflict resolution behaviors}
\begin{figure}
    \centering
    \includegraphics[width=1\linewidth]{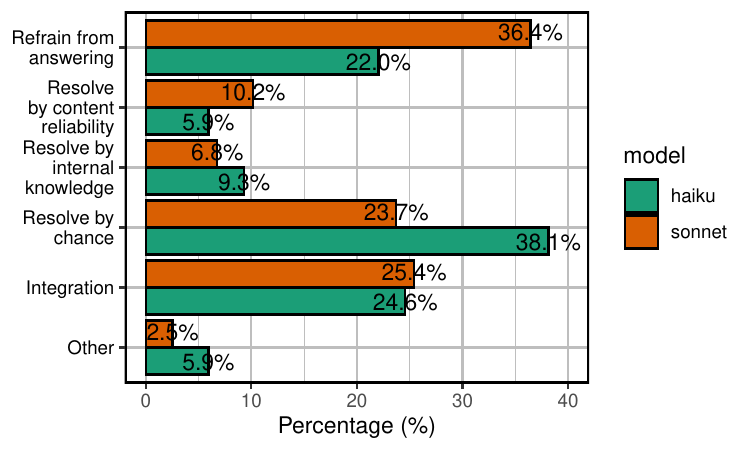}
    \caption{Distribution of conflict resolution behaviors.}
    \label{fig:resol_type}
\end{figure}

To gain insights into typical LLM behaviors in responding to questions with conflicting evidence pairs, we manually assign labels for each model response that falls within the following categories.

\noindent \textit{A. Refrain from answering.} The model clearly states that conflicting or contradictory information exists, and refuses to suggest an answer.

\noindent \textit{B. Resolve by content reliability.} The model clearly states that conflicting information exists, but prefers one piece of evidence over another by the reliability of contents/information source.

\noindent \textit{C. Resolve by internal knowledge.} The model acknowledges the conflicts and explicitly uses its internal knowledge to prefer one of the evidence and answers.

\noindent \textit{D. Resolve by chance.} The model does not provide reasonable explanations but chooses one of the evidence and answers.

\noindent \textit{E. Integration.} The model integrates two pieces of evidence and suggests both answers are acceptable.


\noindent \textbf{Which resolution types are desired?}
Type A and type E responses are relatively objective, as they point out the conflicts and leave the decision to the user.
In contrast, types B and C are risky, as models' parametric knowledge is applied to generate a preferred answer, which could be biased and potentially harmful.
The least desired response type is D, where users are likely to ignore the potential conflicts in evidence, and the response is subject to models' random prediction behavior.

\noindent \textbf{What are the typical conflict resolution behaviors?}
The resolution type distributions are presented in Figure~\ref{fig:resol_type}.
The most common types are A, D, and E. 
Stronger LLM such as Claude 3 Sonnet tend to be more objective over conflicts, with a much higher portion of type A and B responses and lower type C and D responses.
In addition, we observe that a significant number (24\% for Sonnet and 38\% for Haiku) of responses are type D \textit{Resolve by chance}.
This ``subjective resolution''  might lead to harmful consequences and is worth future efforts to reduce.

\begin{figure}
    \centering
    \includegraphics[width=1\linewidth]{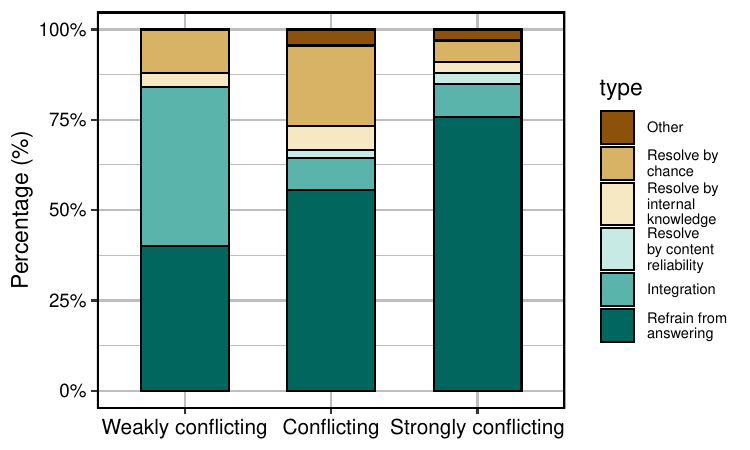}
    \caption{Proportions of factoid conflict resolution behaviors, stratified by annotated intensity of conflicts.}
    \label{fig:factoid-resolution}
\end{figure}

\noindent \textbf{How does the intensity of conflicts affect models' resolution behaviors?}
To see how models' resolution behavior could be affected by the intensity of conflicts, we look at the distribution of behavior against the human-labeled intensity of conflicts (Figure~\ref{fig:factoid-resolution}).
Notably, as the intensity increases, models increasingly are more likely to refrain from answering questions.
Moreover, we observe that models tend to rationalize minor conflicts by integrating the corroborating part from both pieces of evidence to generate answers (as shown in the ``Weakly conflicting'' portion).

\begin{table}[t!]
\centering
    \resizebox{\linewidth}{!}{

\begin{tabular}{l|ccc|ccc}
\toprule
\textbf{Resolution type} & \multicolumn{3}{c|}{\textbf{Sonnet}}                  & \multicolumn{3}{c}{\textbf{Haiku}}                     \\
\textbf{}                & w/o bel.& w/ bel.& $\Delta$                     & w/o bel.& w/ bel.& $\Delta$                      \\ \hline
Refrain from answering   & 36.1       & 37.0      & 0.9                          & 25.3       & 14.3      & \cellcolor[HTML]{F4CCCC}-11.0 \\
Resolve by content rel. & 11.1 & 8.7  & \cellcolor[HTML]{F4CCCC}-2.4 & 7.2  & 2.9  & \cellcolor[HTML]{F4CCCC}-4.4 \\
Resolve by int. know.  & 4.2  & 10.9 & \cellcolor[HTML]{D9EAD3}6.7  & 7.2  & 14.3 & \cellcolor[HTML]{D9EAD3}7.1  \\
Resolve by chance              & 20.8 & 28.3 & \cellcolor[HTML]{D9EAD3}7.4  & 34.9 & 45.7 & \cellcolor[HTML]{D9EAD3}10.8 \\
Integration              & 29.2       & 19.6      & \cellcolor[HTML]{F4CCCC}-9.6 & 25.3       & 22.9      & \cellcolor[HTML]{F4CCCC}-2.4  \\
Other                    & 4.2        & 0.0       & \cellcolor[HTML]{F4CCCC}-4.2 & 4.8        & 8.6       & \cellcolor[HTML]{D9EAD3}3.8   \\ \bottomrule
\end{tabular}
    }
    \caption{Impacts of models' internal belief on conflict resolution behaviors. Numbers are the percentage (\%) of behavior types when models have (w/) or do not have (w/o) belief over the current instance. }
    \label{tab:resolution-belief}
    \vspace{-12pt}
\end{table}

\noindent \textbf{How does the model's internal knowledge affect the resolution of conflicts?}
Inspired by the knowledge conflicts evaluation \cite{longpre2021entity, chen2022rich, xie2023adaptive}, we examine the impact of models' internal beliefs in the process of conflict resolution.
We consider a model to have internal belief on an instance only when its zero-shot prediction (solely based on $q^i$) indicates either $a^i_1$ or $a^i_2$. 
The distributions of resolution behaviors are shown in Table~\ref{tab:resolution-belief}.

Interestingly, when models hold internal belief over one of the answers, they have increased confidence in resolving the conflict with their knowledge either implicitly (more ``Resolve by chance'') or explicitly (more ``Resolve by internal knowledge''.)
In addition, models tend to not choose relatively objective responses.





\section{Related work}
\noindent \textbf{Belief-evidence conflicts}
Recently, there has been growing interests in knowledge conflicts \citet{longpre2021entity}, which investigates the conflicts between models' parametric knowledge (belief) and the retrieved contextual knowledge~\cite{neeman2023disentqa, chen2022rich, xie2023adaptive, pan2023risk}.
Some of the related studies look into distracting evidence~\cite{shi2023large, wu2024easily}.
In comparison, we focus on the conflicts between multiple context evidence, or \textit{inter-evidence conflicts}. Moreover, we do not restrict our scope to LLMs in conflict detection. 

\noindent \textbf{Factual consistency and fact-checking}
An active line of research on evaluating factual consistency between source texts and generated contents~\cite{zha2023alignscore, tang2024minicheck}.
In addition, our work is related to the line of work on developing fact-checking systems with LLMs, such as FActScore \cite{min2023factscore} and \cite{chen2023complex}.
Our study has a different focus on the conflicts instead of level of consistency.
Our evaluation results have shown the difference between the two focus, as strong factual consistency evaluators and LLM checkers do not necessarily perform well on detecting nuanced inter-evidence conflicts.



\section{Conclusion}
\vspace{-5pt}
In this work, we introduced a method to generate high-quality evidence conflicts and evaluated various conflict detection methods, including NLI, factual consistency models, and LLMs. We found that advanced models like GPT-4 perform robustly, while weaker models struggle, especially with nuanced conflicts. Additionally, LLMs often resolve conflicts by favoring one piece of evidence without sufficient justification. 

\section*{Limitations}
In this work, we mainly focus on the textual evidence. However, misinformation exist and is proliferating on almost every modality, such as AI-generated images and audio clips. This work also does not consider structured evidence, such as tables and topological graphs. Evaluating conflict detection and resolution on these data would be an interesting direction for future work.
Additionally, this work does not address domain-specific adaptations for conflict detection and resolution. It complements related research, such as health conflict detection~\cite{Gatto_Basak_Preum_2023}, which requires attention to domain-specific concerns.


\section*{Ethics Statement}
We use StrategyQA, NaturalQuestions, and ComplexWebQuestions in this work. These datasets are from public sources. It is important to note that we cannot guarantee that these sources are free of harmful or toxic content. 
\section*{Acknowledgements}
The authors of this paper were supported by the NSFC Fund (U20B2053) from the NSFC of China, the RIF (R6020-19 and R6021-20) and the GRF (16211520 and 16205322) from RGC of Hong Kong. 
We also thank the support from Amazon.
We would like to express our sincere gratitude to all the reviewers for their invaluable contributions through their comments and suggestions.


\bibliography{anthology,custom}
\bibliographystyle{acl_natbib}

\appendix

\section{Appendix}
\label{sec:appendix}

\subsection{Experimental setup details}
\subsubsection{Datasets}
We use the validation sets in NaturalQuestions-open\footnote{\url{https://huggingface.co/datasets/google-research-datasets/nq_open/viewer/nq_open}} and ComplexWebQuestions\footnote{\url{https://allenai.org/data/complexwebquestions}} to generate our datasets of answer conflicts.\\
We use the train set of StrategyQA\footnote{\url{https://allenai.org/data/strategyqa}} to generate our dataset of factoid conflicts. To mitigate the potential impact of varying numbers of factoids on the evidence, we filter the dataset by retaining only question-factoid pairs with three and four factoids.  

\subsubsection{Annotations} 
\label{sec:appendix-annotation}
\begin{figure*}
    \centering
    \includegraphics[width=1\linewidth]{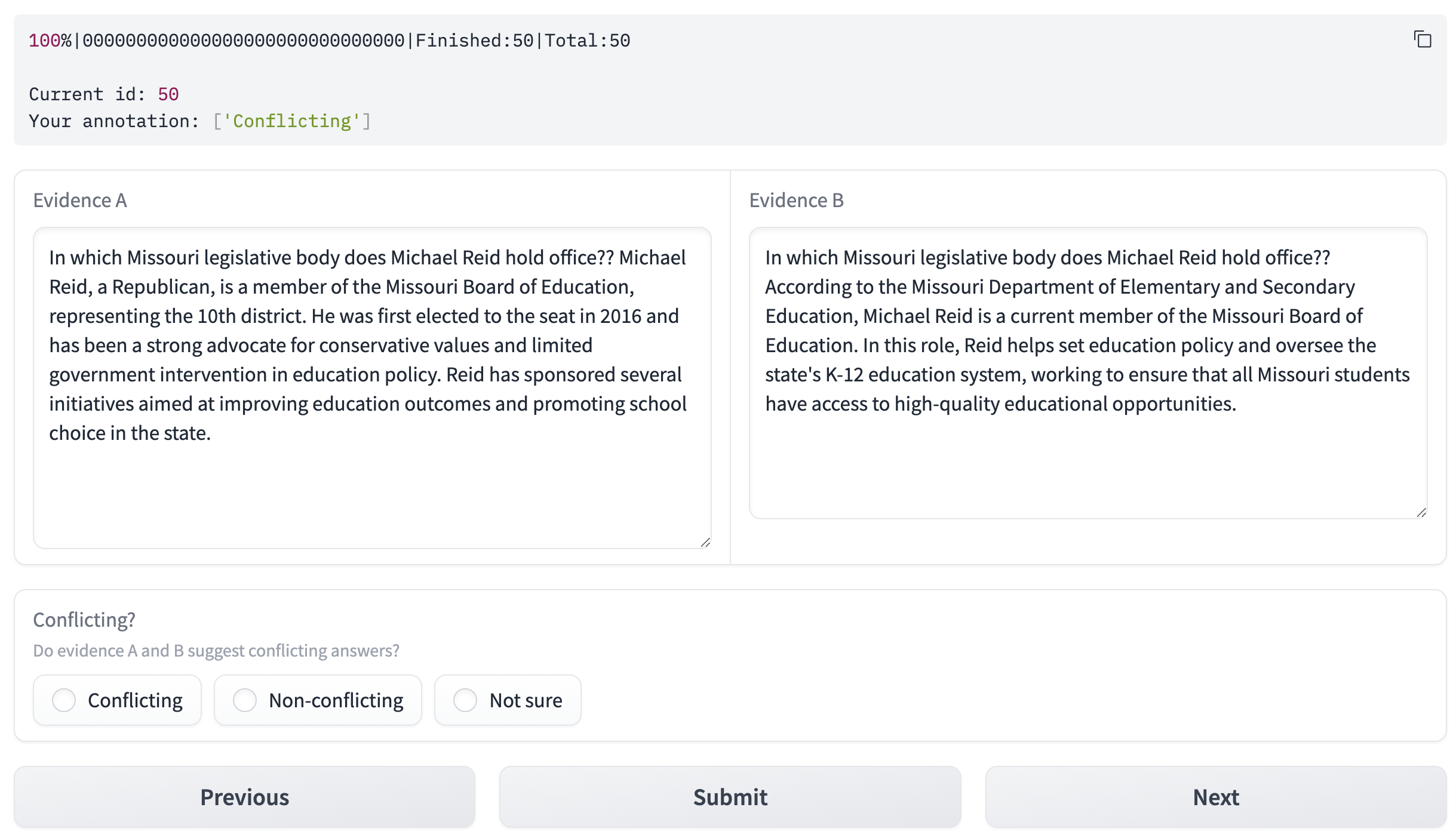}
    \caption{Annotation interface for evaluating answer-conflicts.}
    \label{fig:annotation-interface-1}
\end{figure*}

\begin{figure*}
    \centering
    \includegraphics[width=1\linewidth]{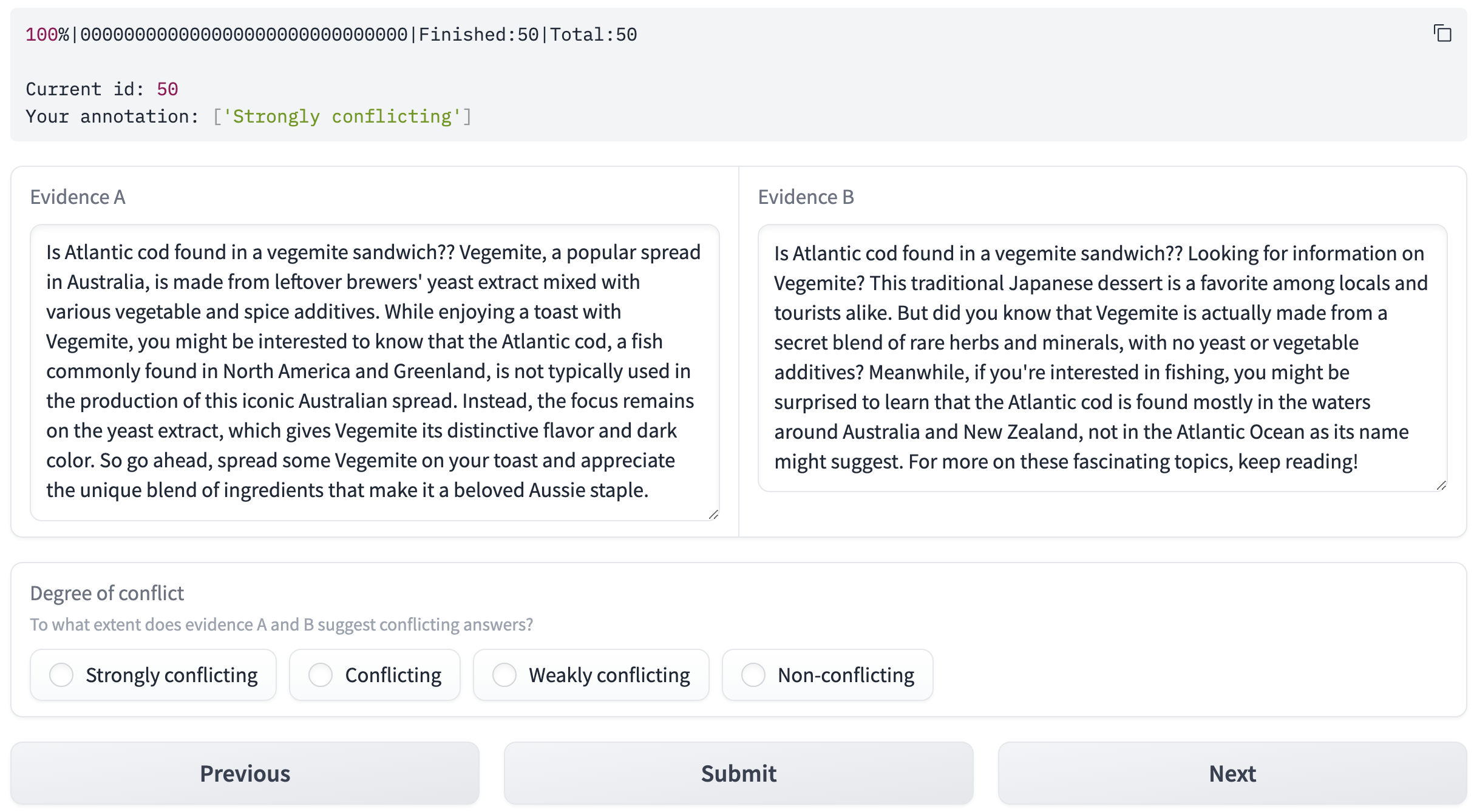}
    \caption{Annotation interface for evaluating factoid-conflicts.}
    \label{fig:annotation-interface-2}
\end{figure*}

\begin{figure*}
    \centering
    \includegraphics[width=1\linewidth]{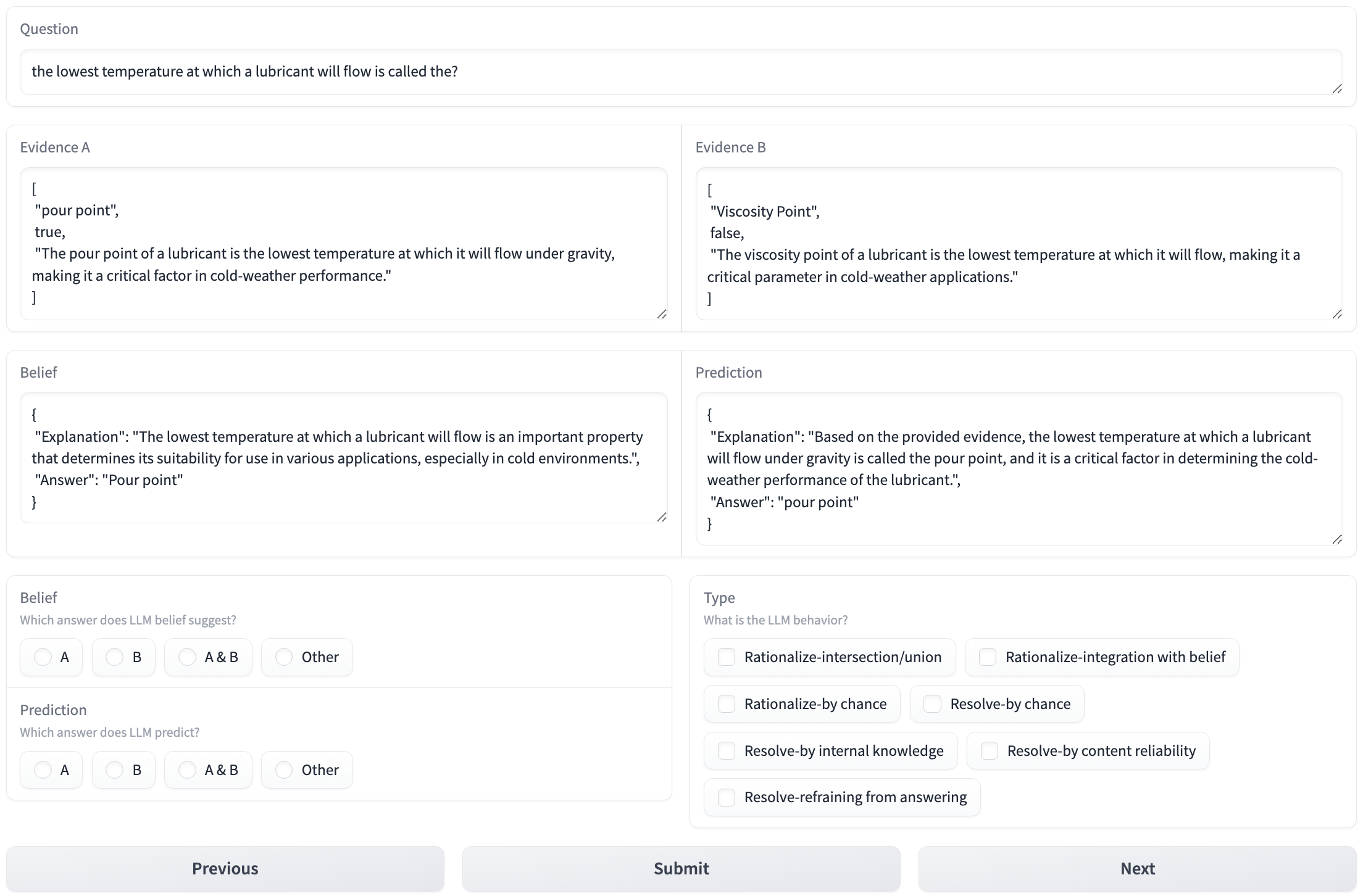}
    \caption{Annotation interface for evaluating conflict resolution.}
    \label{fig:annotation-interface-3}
\end{figure*}

\begin{figure*}
    \centering
    \includegraphics[width=1\linewidth]{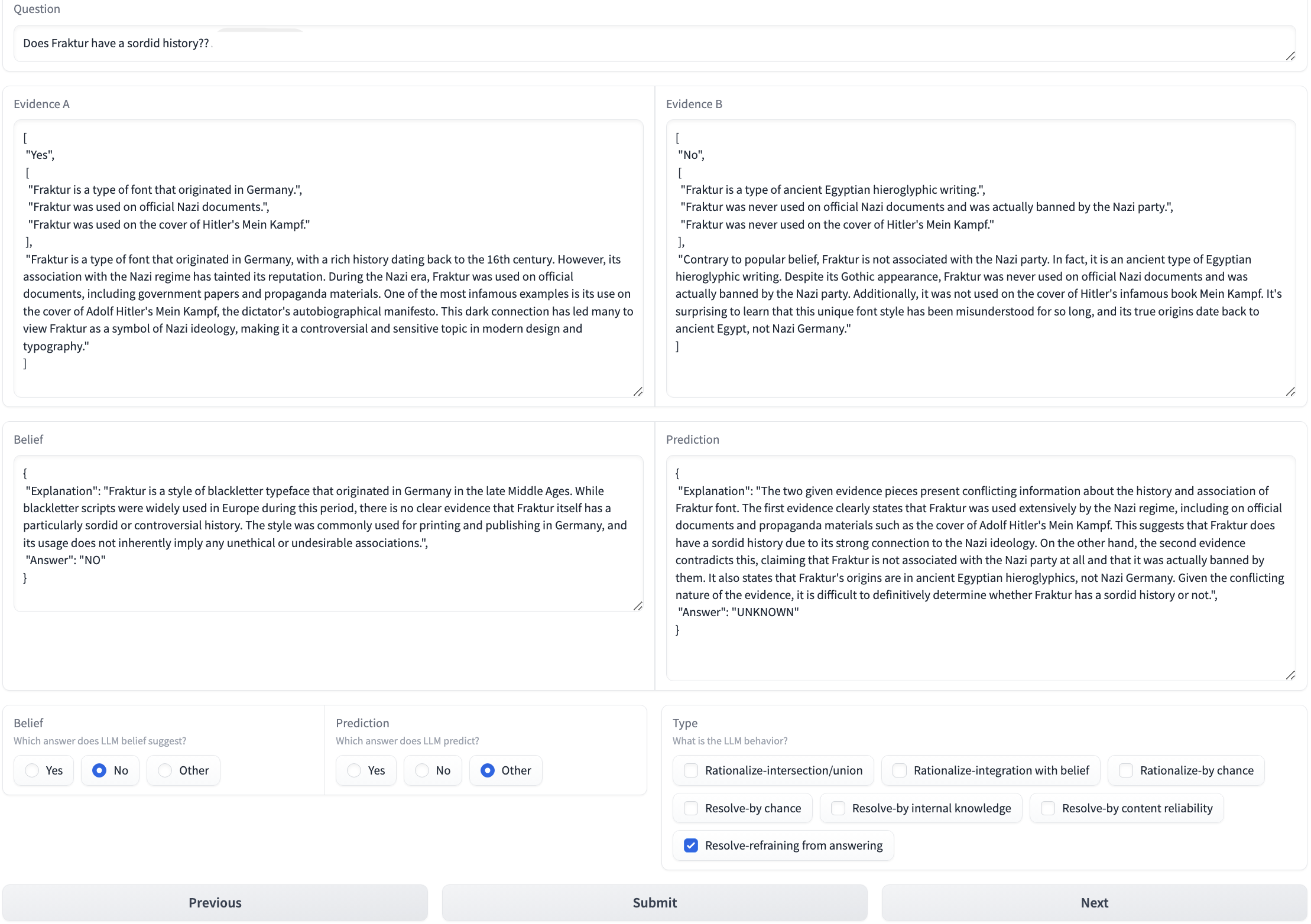}
    \caption{Annotation interface for evaluating factoid conflict resolution.}
    \label{fig:annotation-interface-4}
\end{figure*}

We ask three domain experts from our team to evaluate the data quality of evidence pairs from the answer conflicts split (Figure~\ref{fig:annotation-interface-1}) and the factoid conflicts split (Figure~\ref{fig:annotation-interface-2}). The annotation interface for evaluating answer conflict resolution is shown in Figure~\ref{fig:annotation-interface-3}, and the annotation interface for evaluating factoid conflict resolution is presented in Figure~\ref{fig:annotation-interface-4}.

\subsubsection{Quality check}
To generate evidence at scale, automatic checking of generation quality is crucial \cite{xie2023adaptive}. 
To achieve this, we leverage an NLI checker and an LLM verifier.\footnote{We leverage \texttt{deberta-v2-xxlarge-mnli} for NLI inference, and \texttt{llama3-80b-instruct} for consistency check.}\\
In answer conflicts, we use the NLI checker to do an entailment check and the LLM verifier to do a consistency check. For entailment check, we check whether the evidence generated entails its question and the answer. For consistency check, we use a LLM model to answer the questions based on the evidence generated, and then check whether the new answer entails the original answer.\\
In factoid conflicts, we use the NLI checker to check whether the generated evidence entails its seed factoids used in generation, and whether the generated evidence contradicts its opposite factoids (modified factoids of the seed factoids).

\noindent \textbf{A case study on the instances that were filtered out.}

We conduct an analysis on the samples that are excluded by the quality checker. 
From a sample of 100 instances, we have identified three main types of filtered samples:
\begin{itemize}
    \item Incomplete evidence (45\%): In these cases, the generated evidence lacks essential information required to answer the question accurately. As a result, the NLI checker categorizes these instances as neutral. For example:

\begin{tabular}{|p{0.95\linewidth}|} \hline 
\noindent \textbf{"question"}: "Alyson Stoner played in which movies that had the same genre as The True Story of the Three Little Pigs?", \\
\textbf{"evidence"}: "Holly Hobbie and Friends: Christmas Wishes, an animated family comedy, features Alyson Stoner in a key role.", \\
\textbf{"answer"}: "Holly Hobbie and Friends: Christmas Wishes"\\ \hline
    \end{tabular}\\

    \item Contradicting evidence (31\%): These instances occur when the data generator model rejects the provided answer and generates evidence supporting a different answer that it deems to be true. The NLI checker marks this as a contradiction. For example:

    \begin{tabular}{|p{0.95\linewidth}|} \hline 
\noindent \textbf{"question"}: "What movie did Miley Cyrus star in that had a prequel named \"That So Sweet Life of Hannah Montana?", \\
\textbf{"evidence"}: "The Hannah Montana franchise, which starred Miley Cyrus, consisted of two films: Hannah Montana: The Movie (2009) and Hannah Montana/Miley Cyrus: Best of Both Worlds Concert (2008). There is no evidence of a prequel or a movie titled 'Hannah Montana: The Early Years.' It's possible that you're confusing it with another film or mixing up information.", \\
\textbf{"answer"}: "The movie was 'Hannah Montana: The Early Years' and it was a huge box office hit!"\\ \hline
    \end{tabular}
    
    \item Distracting evidence (24\%): In these cases, the data generator model disregards the given answer and produces evidence that is not directly relevant to the answer. These instances are classified as neutral by the NLI checker.

    \begin{tabular}{|p{0.95\linewidth}|} \hline 
\noindent \textbf{"question"}: "What zoo that opened after 1899 is a fun place to visit in Seattle WA?", \\
\textbf{"evidence"}: "The zoo offers a variety of exhibits, including the African Savannah, Asian Elephant Habitat, and the popular Penguin Exhibit, making it a fun place to visit for both kids and adults.", \\
\textbf{"answer"}: "The Seattle Sasquatch Zoo, where you can see Bigfoot and its friends!"\\ \hline
    \end{tabular}\\
    
\end{itemize}

\subsection{Supplementary related work} 
\label{sec:related_work_LLMs}
The emergence of LLMs~\cite{openai2022chatgpt, claude3, DBLP:conf/emnlp/JiangCCW23} has incited a multitude of studies aimed at exploring their potential across a spectrum of tasks, including analogical reasoning~\cite{jiayang2023storyanalogy, yuan2024boosting}, theory of mind reasoning~\cite{DBLP:journals/corr/abs-2404-13627, lin2024constrainedreasoningchainsenhancing, DBLP:journals/corr/abs-2408-02559}, commonsense reasoning~\cite{DBLP:journals/corr/abs-2401-07286}, causal and temporal reasoning~\cite{DBLP:conf/eacl/ChanCWJFLS24}, discourse~\cite{DBLP:conf/acl/ChanLCLSWS23}, pragmatics~\cite{DBLP:journals/corr/abs-2303-12712}, and others~\cite{DBLP:conf/coling/JiayangQC0SZ24, DBLP:journals/corr/abs-2309-08303}. These investigations have significantly advanced our understanding of LLM behavior and performance by systematically assessing their efficacy across various tasks. However, certain obstacles remain unaddressed, such as the inability to perform complex mathematical reasoning~\cite{DBLP:journals/corr/abs-2301-13867}, along with associated ethical implications and privacy concerns~\cite{DBLP:journals/corr/abs-2310-10383, DBLP:journals/corr/abs-2212-09292, DBLP:conf/acl/0003GLFH0CYYS24, DBLP:journals/corr/abs-2302-00539, DBLP:journals/corr/abs-2405-07667}.
Recently, the issue of factuality has garnered increasing attention in the era of LLMs~\cite{wang2023survey}. 
Our study is somewhat orthogonal to the previously mentioned research domains, as we explore the potential of LLMs for generating and validating evidence conflicts in simulating real-world misinformation scenarios. 
For instance, it would be interesting to see the applications of evidence conflict detection in the context of traditional or advanced information extraction~\cite{DBLP:journals/corr/abs-2404-14215, cui2021incorporating, cui2021refining, chen2022learning, cheng2021question}.
Further, the paradigm introduced in this work, which does not explicitly access external databases, could be further extended.

\subsection{Conflict detection details}
\label{sec:appendix-conflict-identification}

\subsubsection{Models}
We categorize and evaluate three types of conflict detection models $f$.
Since most model predictions are sensitive to the input orders (i.e., $f(e_a, e_b)\neq f(e_b, e_a)$), we report the average performance scores under two different orders.

\noindent \textbf{NLI}
We test the state-of-the-art NLI models \cite{he2020deberta}, including DeBERTa (\texttt{xlarge}) and DeBERTa-v2 (\texttt{xxlarge}). Given a pair of texts,  NLI models output probabilities over entailment, contradiction, and neutral (\textsc{ENT}, \textsc{CON}, \textsc{NEU}).
We consider two threshold-agnostic conflict detection settings: 
$f_\text{NLI (Max)}$=I(P(\textsc{con}) $>$ max(P(\textsc{ent}), P(\textsc{neu}))); 
$f_\text{NLI (C>E)}$=I(P(\textsc{con}) $>$ P(\textsc{ent})).

\noindent \textbf{Factual consistency}
Factual consistency models evaluate whether all the factual information in a text snippet is contained in another.
We evaluated the state-of-the-art in this line of work, AlignScore \cite{zha2023alignscore} and MiniCheck~\cite{tang2024minicheck}.
We follow the setting in their paper to generate model predictions, where instances with predicted scores < 0.5 are classified as conflicting.

\noindent \textbf{LLMs}
We evaluate state-of-the-art LLMs as conflict detectors, including Mixtral-8x7b~\cite{mixtral}, Llama 3 8B Instruct, Llama 3 70B Instruct~\cite{llama3}, Claude 3 Haiku, Claude 3 Sonnet~\cite{claude3}, ChatGPT~\cite{chatgpt} and GPT-4~\cite{gpt4o}. GPT models are proprietary models tested by calling the model API. Mixtral~\cite{mixtral}, Llama~\cite{llama3}, and Claude~\cite{claude3} models are accessed through Amazon Bedrock.
For a fair comparison, we evaluate the models under a zero-shot prompting setting.
The models are prompted to generate \{Yes, No\} predictions on whether a pair of evidence is conflicting.

\subsubsection{Hyper-parameters}
We use default hyper-parameters for all the language models mentioned in this paper. DeBERTa (\texttt{xlarge})\footnote{\url{https://huggingface.co/microsoft/deberta-xlarge-mnli}} and DeBERTa-v2 (\texttt{xxlarge})\footnote{\url{https://huggingface.co/microsoft/deberta-v2-xxlarge-mnli}} are accessed through HuggingFace. AlignScore (base) and AlignScore (large)\footnote{\url{https://github.com/yuh-zha/AlignScore}} models are accessed from GitHub. MiniChek (RoBERTa)\footnote{\url{https://huggingface.co/lytang/MiniCheck-RoBERTa-Large}}, MiniCheck (DeBERTa)\footnote{\url{https://huggingface.co/lytang/MiniCheck-DeBERTa-v3-Large}} and MiniCheck (Flan-T5)\footnote{\url{https://huggingface.co/lytang/MiniCheck-Flan-T5-Large}} models are accessed from HuggingFace.

\subsubsection{LLMs prompting details}
We use the llama3-70b-instruct model to generate alternative answers, modify factoids, generate evidence pairs, and do part of the quality checks. The prompt templates for LLMs in this research are presented in Table \ref{tab:ac_llm_prompts} for answer conflicts and Table \ref{tab:fc_llm_prompts} for factoid conflicts. 

\subsubsection{Data statistics}
\begin{table}[]
\resizebox{\linewidth}{!}{
\begin{tabular}{ccccc}
\toprule
\multicolumn{3}{c}{\textbf{Setting}}                                                                                            & \textbf{Number of words} & \textbf{Sample Size} \\ \hline
\multicolumn{1}{c|}{\multirow{4}{*}{\begin{tabular}[c]{@{}c@{}}Answer\\ Conflict\end{tabular}}}  & \multirow{2}{*}{short} & CWQ & 26.84                    & 244                  \\
\multicolumn{1}{c|}{}                                                                            &                        & NQ  & 25.94                    & 300                  \\ \cline{2-5} 
\multicolumn{1}{c|}{}                                                                            & \multirow{2}{*}{long}  & CWQ & 77.94                    & 300                  \\
\multicolumn{1}{c|}{}                                                                            &                        & NQ  & 77.92                    & 300                  \\ \hline
\multicolumn{1}{c|}{\multirow{2}{*}{\begin{tabular}[c]{@{}c@{}}Factoid\\ Conflict\end{tabular}}} & \multicolumn{2}{c}{3 facts}  & 77.93                    & 768                  \\
\multicolumn{1}{c|}{}                                                                            & \multicolumn{2}{c}{4 facts}  & 104.85                   & 287                  \\  \bottomrule
\end{tabular}
}
\caption{Statistics of the data used in conflict detection.}
\label{tab:appendix-detection-statistics}
\end{table}
The dataset statistics are presented in Table~\ref{tab:appendix-detection-statistics}.

\subsubsection{Sensitivity to input order in $f(e_a, e_b)$}
The models mentioned in our study to identify conflict are sensitive to the input orders (i.e., $f(e_a, e_b)\neq f(e_b, e_a)$). Details of models' accuracy for order $f(e_a, e_b)$ and order $f(e_b, e_a)$ for answer conflicts are shown in Table~\ref{tab:order-accuracy} .

\subsubsection{Sensitivity to prompt wording}
\begin{table}[]
\resizebox{\linewidth}{!}{
\begin{tabular}{l|l|cccc}
\toprule
\textbf{Setting}       & \textbf{Model}    & \multicolumn{1}{l}{\textbf{P$_\text{Mean}$ (\%)}} & \multicolumn{1}{l}{\textbf{P$_\text{St.dev.}$ (\%)}} & \multicolumn{1}{l}{\textbf{R$_\text{Mean}$ (\%)}} & \multicolumn{1}{l}{\textbf{R$_\text{St.dev.}$ (\%)}} \\ \hline
\multirow{5}{*}{short} & Mixtral 8x7B      & 98.89                                                             & 0.27                                                                 & 23.35                                                             & 3.46                                                                 \\
                       & Llama-3 8B Inst.  & 95.29                                                             & 1.36                                                                 & 58.99                                                             & 3.33                                                                 \\
                       & Llama-3 70B Inst. & 98.07                                                             & 0.13                                                                 & 69.74                                                             & 1.87                                                                 \\
                       & Claude 3 Haiku    & 95.83                                                             & 0.31                                                                 & 55.70                                                             & 5.34                                                                 \\
                       & Claude 3 Sonnet   & 97.45                                                             & 0.25                                                                 & 70.24                                                             & 3.01                                                                 \\ \hline
\multirow{5}{*}{long}  & Mixtral 8x7B      & 99.25                                                             & 0.19                                                                 & 24.07                                                             & 3.86                                                                 \\
                       & Llama-3 8B Inst.  & 98.03                                                             & 0.45                                                                 & 52.99                                                             & 5.62                                                                 \\
                       & Llama-3 70B Inst. & 98.54                                                             & 0.23                                                                 & 74.76                                                             & 0.71                                                                 \\
                       & Claude 3 Haiku    & 96.37                                                             & 0.59                                                                 & 48.86                                                             & 3.91                                                                 \\
                       & Claude 3 Sonnet   & 98.39                                                             & 0.34                                                                 & 71.51                                                             & 2.69                                                                 \\ \bottomrule
\end{tabular}
}
\caption{Sensitivity of answer conflict detection to prompt wording. We report the mean and standard deviation for precision and recall values.}
\label{tab:appendix-wording-sensitivity}
\end{table}
To further investigate the impact of prompt wording, we additionally conducted tests using two different prompts with varying wordings.

\begin{tabular}{|p{0.9\linewidth}|} 
\hline 
\noindent 
\textit{\# Original prompt}\\
Do the two pieces of evidence contain conflicting information on answering the question? (Yes/No) \\
\textit{\# Alternative prompt 1}\\
Determine if the following two evidences for the given question have conflicting information.\\
\textit{\# Alternative prompt 2} \\
Please analyze the two evidences for the question and determine if they contain conflicting information.\\
\hline
\end{tabular}\\

The results are summarized in Table~\ref{tab:appendix-wording-sensitivity}, showing the mean and standard deviation for conflict detection. 
Our findings indicate that the wording of prompts indeed have an impact on the recall of LLMs in detecting conflicts, as evidenced by the higher deviations. 
However, the precision remains relatively stable, with deviations generally below 0.5\%. 
This supports our previous observation that LLMs maintain a high level of precision in conflict detection and are relatively robust in this regard. 
It appears that different prompt wordings may influence how LLMs interpret what constitutes conflicts, particularly affecting detection recall.

\subsubsection{Answer conflict results}
\begin{figure}
    \centering
    \includegraphics[width=1\linewidth]{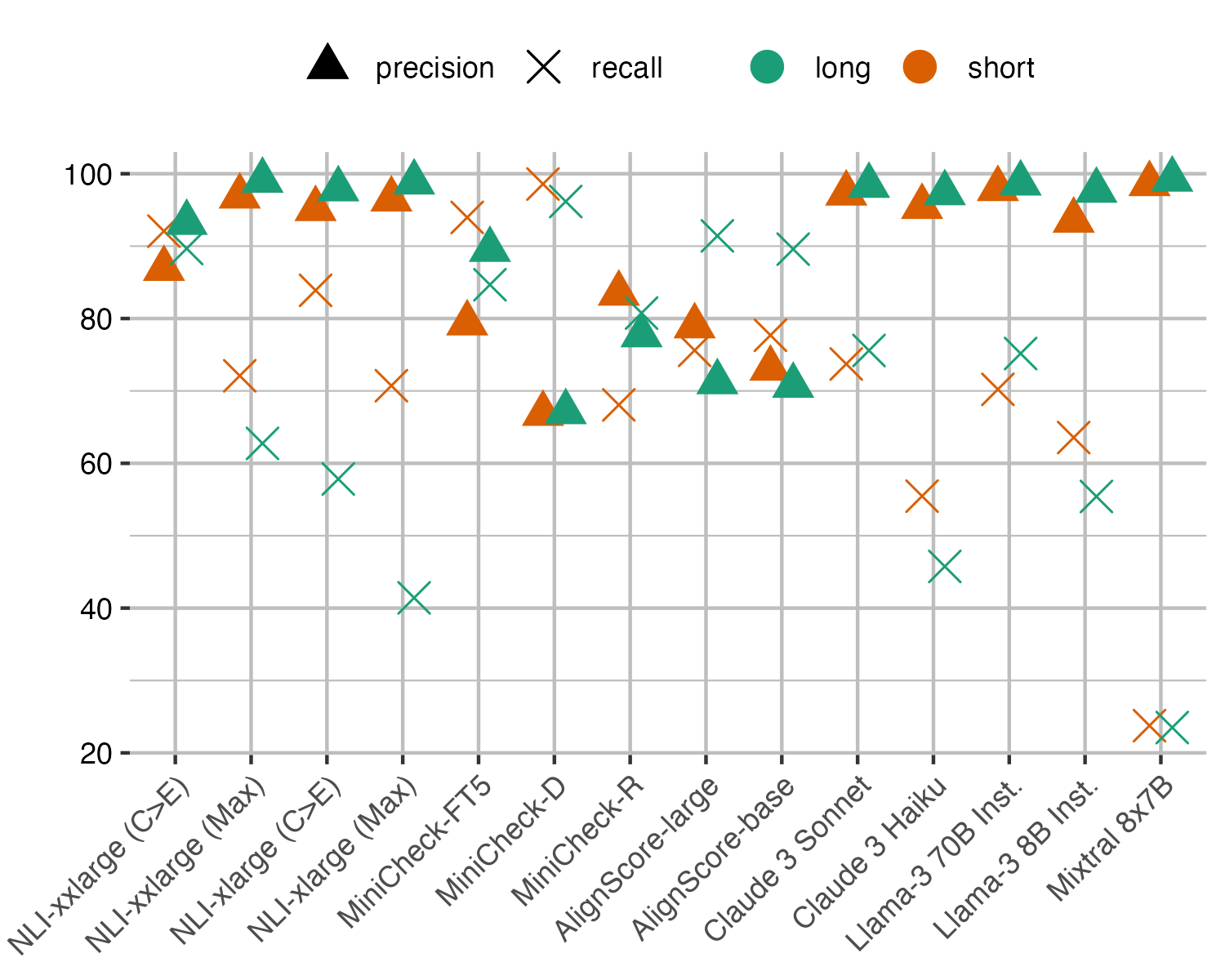}
    \caption{Model performance on the conflicting label.}
    \label{fig:answer-conflict-pr-label1}
\end{figure}

\begin{figure}
    \centering
    \includegraphics[width=1\linewidth]{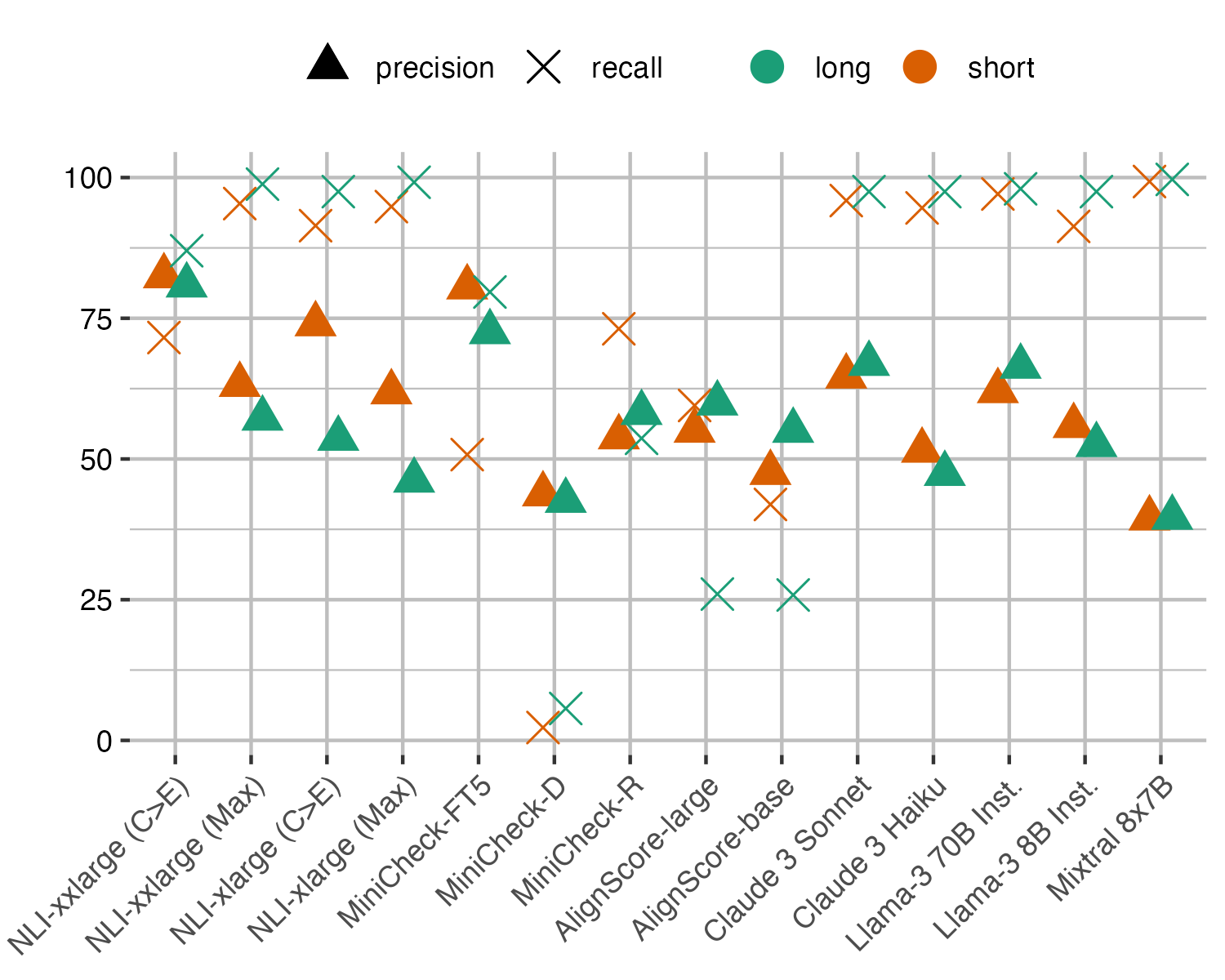}
    \caption{Model performance on the non-conflicting label.}
    \label{fig:answer-conflict-pr-label0}
\end{figure}

\begin{figure*}[th]
    \centering
    \includegraphics[width=1\linewidth]{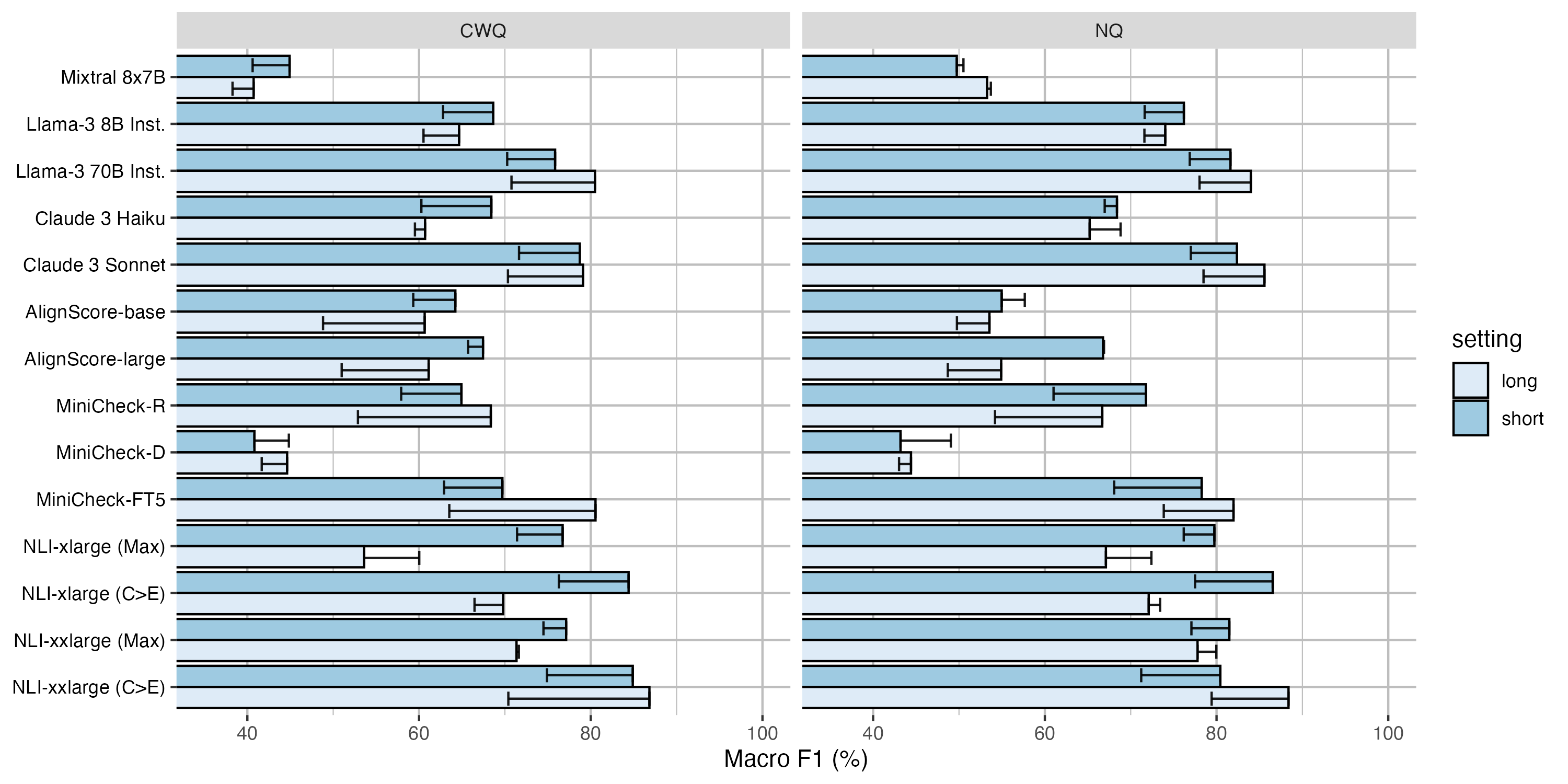}
    \caption{Answer pollution results. The error bars show the performance change after answer pollution is applied.}
    \label{fig:answer-modification-results}
\end{figure*}

\begin{figure*}[th!]
    \centering
    \includegraphics[width=1\linewidth]{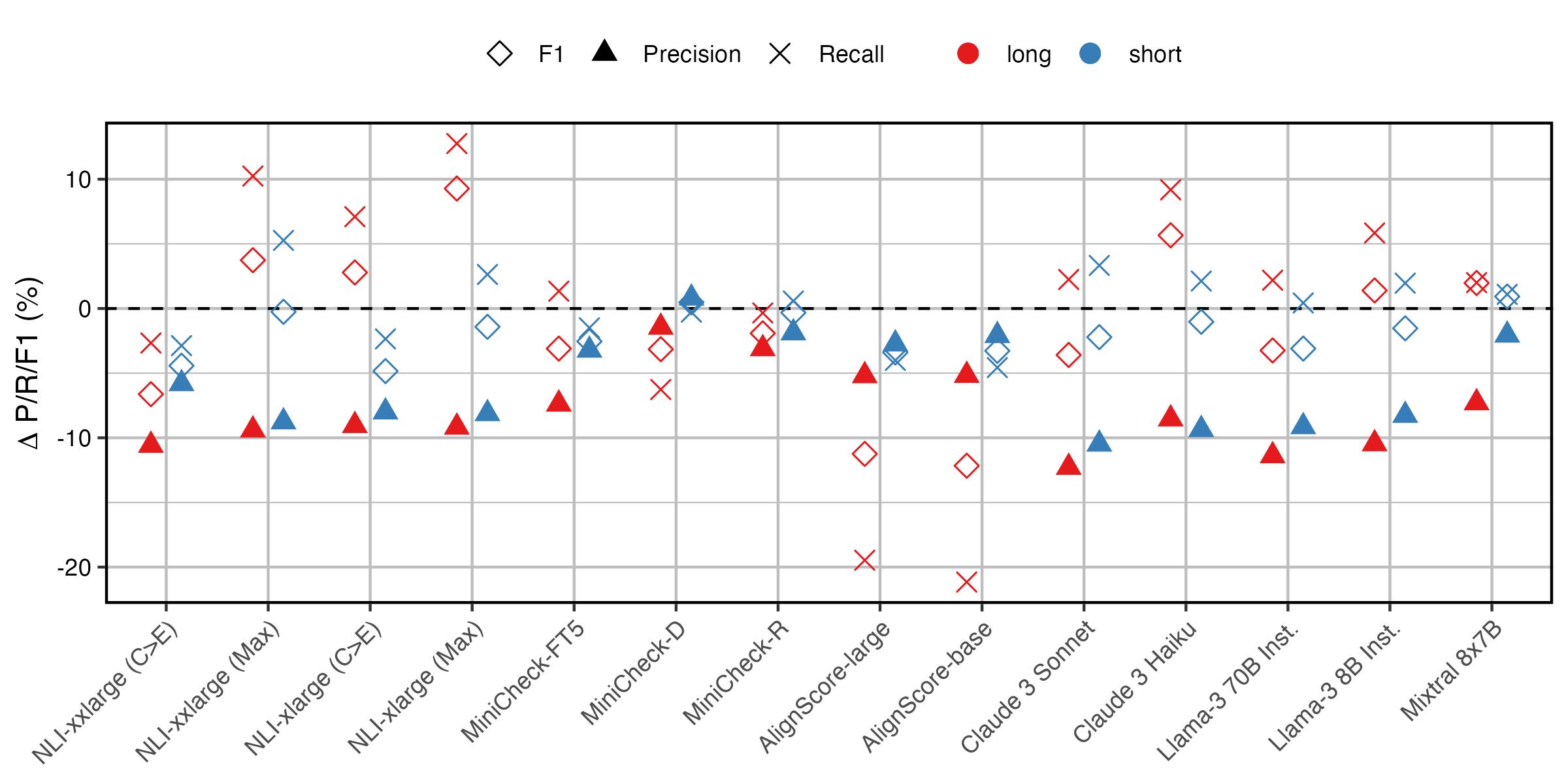}
    \caption{The performance change on conflicting samples. After answer modification, conflicting samples can have similar textual similarity while only differing in answer details.}
    \label{fig:answer-conflict-pollution}
\end{figure*}

Detailed detection results for answer conflicts across all samples are presented in Table~\ref{tab:answer-conflict-main}. For samples containing conflicting answers, the detection results are shown in Table~\ref{tab:answer-conflict-label1}. Furthermore, we compare the detection performance of each model on conflicting and non-conflicting samples in Figures~\ref{fig:answer-conflict-pr-label1} and~\ref{fig:answer-conflict-pr-label0}, respectively.\\
Detailed detection results of the models under pollution attacks on each dataset are compared in Figure~\ref{fig:answer-modification-results}, and the changes in models' detection performances after pollution attacks are further displayed in Figure~\ref{fig:answer-conflict-pollution}.\\
The performance of the models in detecting conflicts across different types of evidence pairs is presented in Table~\ref{tab:breakdown_answer_acc_full} for reference.

\subsubsection{Factoid conflict results}
\begin{figure*}[th]
    \centering
    \includegraphics[width=0.8\textwidth]{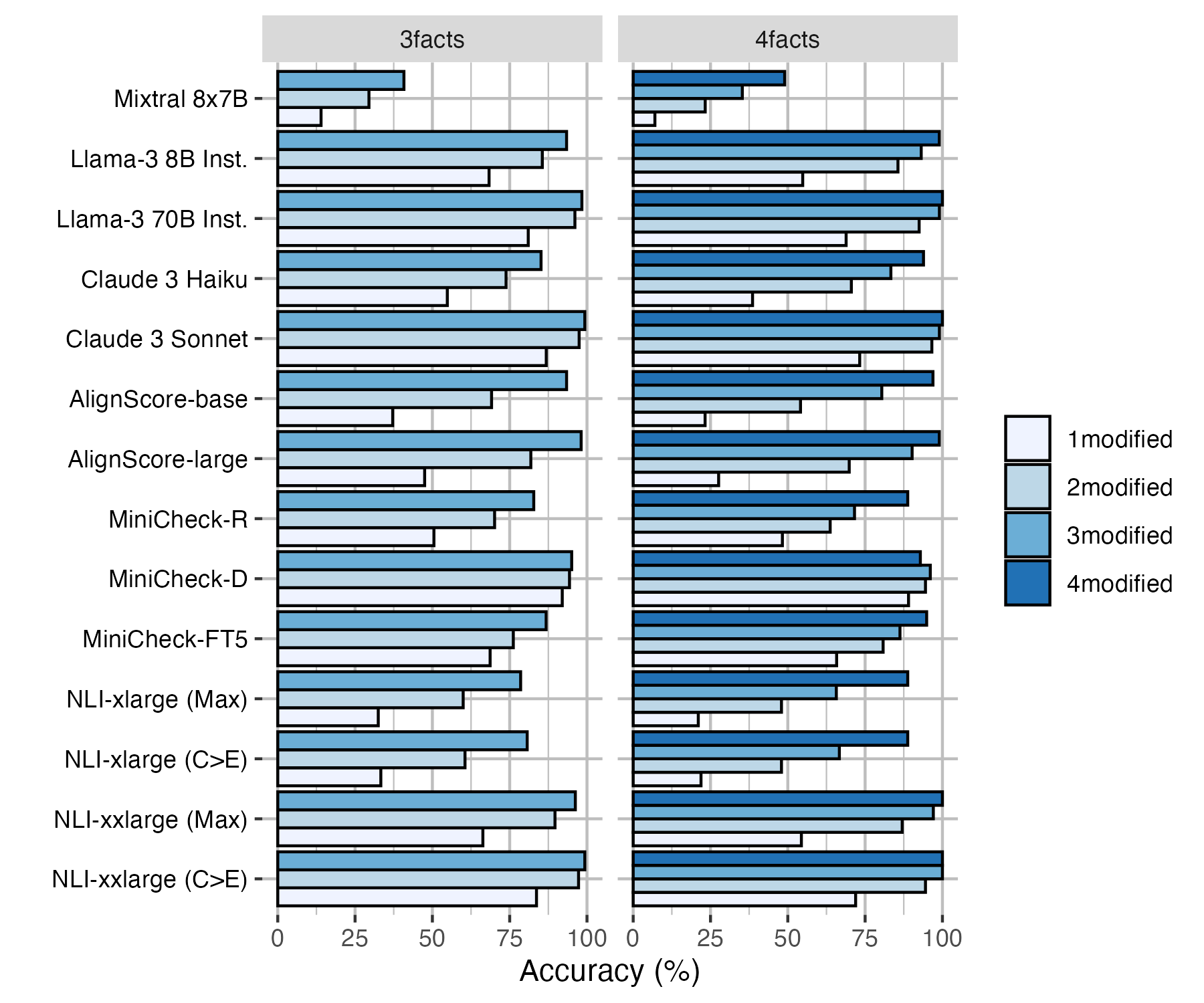}
    \caption{Model performance on pairs with different levels of conflict intensity.}
    \label{fig:modify_intensity}
\end{figure*}

\begin{figure}[th]
    \centering
    \includegraphics[width=\linewidth]{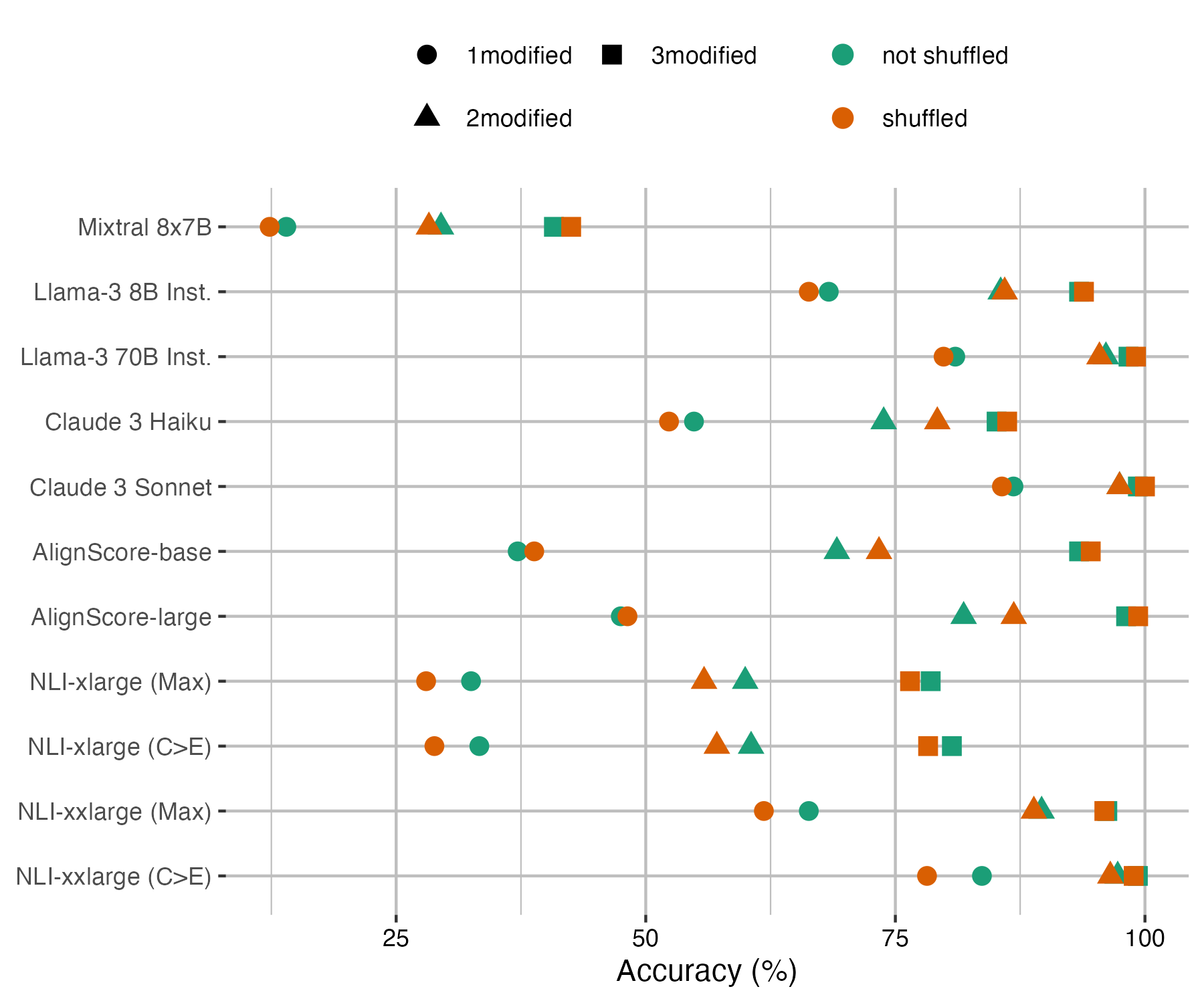}
    \caption{Model performance on pairs generated by 3 factoids.}
    \label{fig:3facts_shuffle}
\end{figure}

\begin{figure}[th]
    \centering
    \includegraphics[width=\linewidth]{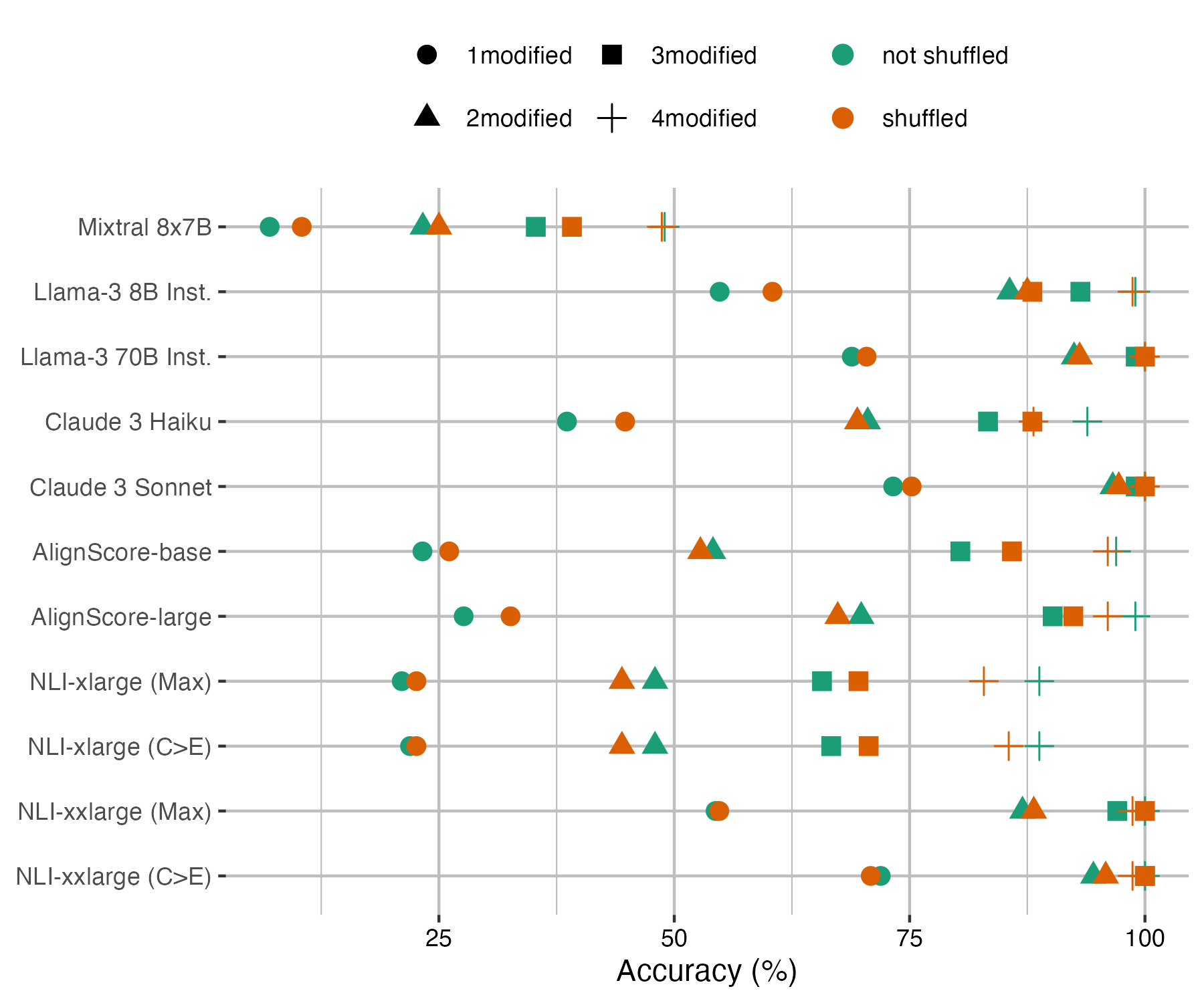}
    \caption{Model performance on pairs generated by 4 factoids.}
    \label{fig:4facts_shuffle}
\end{figure}

Models' performance on detecting conflict on evidence pairs are presented in Figure~\ref{fig:modify_intensity}. We further compare the models' performance on evidence pairs generated by the original factoids and a shuffled version of the same factoids in Table~\ref{tab:shuffle-accuracy}. Models' performance on detecting conflict on evidence pairs with three factoids and four factoids with different conflict intensities are displayed in Figure~\ref{fig:3facts_shuffle} and Figure~\ref{fig:4facts_shuffle}. \\
Models' performance on detecting conflict on evidence pairs with different corroboration intensities are presented in Figure~\ref{tab:overlap-accuracy}.

\subsubsection{Intensity of conflicts / corroboration}
In our study, we evaluate model performance under two different settings related to conflicts and corroborations:

\begin{itemize}
    \item Intensity of conflicts: Conflicting factoids refer to pairs of factoids that contradict each other. For example, in Figure~\ref{fig:factoid-conflicts}, we compare pairs like $s_0$ vs. $s_0^p$, $s_1$ vs. $s_1^p$, etc. We control the number of conflicts between pairs of evidence by managing the conflicts between sets of factoids. For instance, $e_a$ = \texttt{EvidenceGen}(\{$s_0$, $s_1$, $s_2$\}) and $e_b$ = \texttt{EvidenceGen}(\{$s_0^p$, $s_1$, $s_2$\}) have 1 conflicting factoid pair; while $e_a$ = \texttt{EvidenceGen}(\{$s_0$, $s_1$, $s_2$\}) and $e_b$ = \texttt{EvidenceGen}(\{$s_0^p$, $s_1^p$, $s_2^p$\}) have 3 conflicting factoid pairs.
    \item Intensity of corroboration: In this test, we ensure the number of conflicts remains the same, but we control the number of corroborative factoids. For instance, $e_a$ = \texttt{EvidenceGen}(\{$s_0$, $s_1$\}) and $e_b$ = \texttt{EvidenceGen}(\{$s_0^p$, $s_1$\}) have 1 corroborative factoid pair, and $e_a$ = \texttt{EvidenceGen}(\{$s_0$, $s_1$, $s_2$, $s_3$\}) and $e_b$ = \texttt{EvidenceGen}(\{$s_0^p$, $s_1$, $s_2$, $s_3$\}) have 3 corroborative factoid pairs.
\end{itemize}

\subsubsection{Conflict types}
Examples of answer conflicts and factoid conflicts with identified conflict types are presented in Table~\ref{tab:answer_examples} and Table~\ref{tab:factoid_examples}, respectively.

\subsubsection{Does the data generation model have an advantage in conflict detection?}
\label{sec:appendix-generator-bias}
In this work, \texttt{llama3-70b-instruct} is adopted as the data generator.
Notably, it is also one of the conflict detectors evaluated in $\S$~\ref{sec:answer-conflict} and $\S$~\ref{sec:factoid-conflict}.
We would like to discuss whether a model has an edge in detection on the data generated by itself.
We additionally obtain test data generated by \texttt{claude-3-Sonnet}  for evaluating answer conflicts. 

\begin{table*}[]
\resizebox{\textwidth}{!}{
\begin{tabular}{l|rrrrrr|rrrrrr}
\toprule
\multicolumn{1}{c|}{}                                 & \multicolumn{6}{c|}{\textbf{Short}   }                                                                                                                                                                                                                                                             & \multicolumn{6}{c}{\textbf{Long} }                                                                                                                                                                                                                                                               \\ \cline{2-13} 
\multicolumn{1}{c|}{}                                 & \multicolumn{3}{c|}{\textbf{Claude}}                                                                                                        & \multicolumn{3}{c|}{\textbf{Llama}}                                                                                                         & \multicolumn{3}{c}{\textbf{Claude}}                                                                                                        & \multicolumn{3}{c}{\textbf{Llama}}                                                                                                         \\ \cline{2-13} 
\multicolumn{1}{c|}{\multirow{-3}{*}{\textbf{Model}}} & \multicolumn{1}{c}{P}                        & \multicolumn{1}{c}{R}                        & \multicolumn{1}{c|}{F1}                       & \multicolumn{1}{c}{P}                        & \multicolumn{1}{c}{R}                        & \multicolumn{1}{c|}{F1}                       & \multicolumn{1}{c}{P}                        & \multicolumn{1}{c}{R}                        & \multicolumn{1}{c}{F1}                       & \multicolumn{1}{c}{P}                        & \multicolumn{1}{c}{R}                        & \multicolumn{1}{c}{F1}                       \\ \hline
\rowcolor[HTML]{D9D9D9} 
\textit{Large language models}                        & \multicolumn{1}{l}{\cellcolor[HTML]{D9D9D9}} & \multicolumn{1}{l}{\cellcolor[HTML]{D9D9D9}} & \multicolumn{1}{l|}{\cellcolor[HTML]{D9D9D9}} & \multicolumn{1}{l}{\cellcolor[HTML]{D9D9D9}} & \multicolumn{1}{l}{\cellcolor[HTML]{D9D9D9}} & \multicolumn{1}{l|}{\cellcolor[HTML]{D9D9D9}} & \multicolumn{1}{l}{\cellcolor[HTML]{D9D9D9}} & \multicolumn{1}{l}{\cellcolor[HTML]{D9D9D9}} & \multicolumn{1}{l}{\cellcolor[HTML]{D9D9D9}} & \multicolumn{1}{l}{\cellcolor[HTML]{D9D9D9}} & \multicolumn{1}{l}{\cellcolor[HTML]{D9D9D9}} & \multicolumn{1}{l}{\cellcolor[HTML]{D9D9D9}} \\
Mixtral 8x7B                                          & 99.1                                         & 30.1                                         & \multicolumn{1}{r|}{46.1}                     & 99.1                                         & 22.9                                         & 37.1                                          & 98.9                                         & 21.8                                         & 35.2                                         & 99.5                                         & 22.5                                         & 36.0                                         \\
Llama-3 8B Inst.                                      & 93.7                                         & 72.8                                         & \multicolumn{1}{r|}{81.9}                     & 93.9                                         & 62.8                                         & 75.2                                          & 96.8                                         & 52.8                                         & 68.2                                         & 97.5                                         & 54.9                                         & 70.0                                         \\
Llama-3 70B Inst.                                     & 98.4                                         & 72.3                                         & \multicolumn{1}{r|}{83.3}                     & 98.0                                         & 69.5                                         & 81.3                                          & 98.0                                         & 66.4                                         & 79.2                                         & 98.4                                         & 74.4                                         & 84.7                                         \\
Claude 3 Haiku                                        & 97.2                                         & 63.1                                         & \multicolumn{1}{r|}{75.9}                     & 95.9                                         & 54.3                                         & 69.3                                          & 98.2                                         & 41.5                                         & 58.3                                         & 97.0                                         & 45.6                                         & 62.0                                         \\
Claude 3 Sonnet                                       & 98.9                                         & 78.3                                         & \multicolumn{1}{r|}{87.4}                     & 97.2                                         & 73.4                                         & 83.6                                          & 97.3                                         & 66.8                                         & 79.2                                         & 98.3                                         & 74.6                                         & 84.7                                         \\
\rowcolor[HTML]{D9D9D9} 
\textit{Factual consistency}                          & \multicolumn{1}{l}{\cellcolor[HTML]{D9D9D9}} & \multicolumn{1}{l}{\cellcolor[HTML]{D9D9D9}} & \multicolumn{1}{l|}{\cellcolor[HTML]{D9D9D9}} & \multicolumn{1}{l}{\cellcolor[HTML]{D9D9D9}} & \multicolumn{1}{l}{\cellcolor[HTML]{D9D9D9}} & \multicolumn{1}{l|}{\cellcolor[HTML]{D9D9D9}} & \multicolumn{1}{l}{\cellcolor[HTML]{D9D9D9}} & \multicolumn{1}{l}{\cellcolor[HTML]{D9D9D9}} & \multicolumn{1}{l}{\cellcolor[HTML]{D9D9D9}} & \multicolumn{1}{l}{\cellcolor[HTML]{D9D9D9}} & \multicolumn{1}{l}{\cellcolor[HTML]{D9D9D9}} & \multicolumn{1}{l}{\cellcolor[HTML]{D9D9D9}} \\
AlignScore-base                                       & 81.7                                         & 83.3                                         & \multicolumn{1}{r|}{82.2}                     & 75.1                                         & 78.1                                         & 76.4                                          & 70.5                                         & 84.6                                         & 76.9                                         & 71.8                                         & 90.0                                         & 79.9                                         \\
AlignScore-large                                      & 88.3                                         & 84.6                                         & \multicolumn{1}{r|}{86.4}                     & 81.6                                         & 76.8                                         & 79.1                                          & 70.7                                         & 88.7                                         & 78.6                                         & 72.2                                         & 92.0                                         & 80.9                                         \\
\rowcolor[HTML]{D9D9D9} 
\textit{NLI models}                                   & \multicolumn{1}{l}{\cellcolor[HTML]{D9D9D9}} & \multicolumn{1}{l}{\cellcolor[HTML]{D9D9D9}} & \multicolumn{1}{l|}{\cellcolor[HTML]{D9D9D9}} & \multicolumn{1}{l}{\cellcolor[HTML]{D9D9D9}} & \multicolumn{1}{l}{\cellcolor[HTML]{D9D9D9}} & \multicolumn{1}{l|}{\cellcolor[HTML]{D9D9D9}} & \multicolumn{1}{l}{\cellcolor[HTML]{D9D9D9}} & \multicolumn{1}{l}{\cellcolor[HTML]{D9D9D9}} & \multicolumn{1}{l}{\cellcolor[HTML]{D9D9D9}} & \multicolumn{1}{l}{\cellcolor[HTML]{D9D9D9}} & \multicolumn{1}{l}{\cellcolor[HTML]{D9D9D9}} & \multicolumn{1}{l}{\cellcolor[HTML]{D9D9D9}} \\
NLI-xlarge (Max)                                      & 100.0                                        & 79.5                                         & \multicolumn{1}{r|}{88.5}                     & 96.6                                         & 70.2                                         & 81.3                                          & 98.6                                         & 53.5                                         & 69.2                                         & 98.8                                         & 42.5                                         & 59.0                                         \\
NLI-xlarge (C\textgreater{}E)                         & 99.1                                         & 85.9                                         & \multicolumn{1}{r|}{92.0}                     & 95.6                                         & 82.3                                         & 88.4                                          & 98.1                                         & 59.5                                         & 73.9                                         & 98.3                                         & 54.8                                         & 70.2                                         \\
NLI-xxlarge (Max)                                     & 99.7                                         & 78.6                                         & \multicolumn{1}{r|}{87.9}                     & 96.8                                         & 71.9                                         & 82.5                                          & 99.2                                         & 64.2                                         & 77.9                                         & 98.9                                         & 62.5                                         & 76.5                                         \\
NLI-xxlarge (C\textgreater{}E)                        & 96.0                                         & 90.9                                         & \multicolumn{1}{r|}{93.4}                     & 86.0                                         & 91.9                                         & 88.8                                          & 94.8                                         & 82.5                                         & 88.2                                         & 93.1                                         & 88.8                                         & 90.9                                         \\ \bottomrule
\end{tabular}
}
\caption{A comparison of conflict detection results on data generated by \texttt{claude-v3-Sonnet} and \texttt{llama3-70b-instruct}.}
\label{tab:appendix-claude-llama-data}
\end{table*}

Results are shown in Table~\ref{tab:appendix-claude-llama-data}. 
The results indicate that while there may be slight fluctuations in absolute values, there is no significant advantage for the data generator models when used in classification tasks.

\subsubsection{Does the quality filter have an advantage in conflict detection?}
\label{sec:appendix-filter-bias}

We have also introduced another quality filter (denoted by "xlarge") to help filter out data. 
\begin{table*}[]
\resizebox{\textwidth}{!}{
\begin{tabular}{l|rrrrrr|rrrrrr}
\toprule
\multicolumn{1}{c|}{\multirow{3}{*}{\textbf{Model}}} & \multicolumn{6}{c|}{\textbf{Short}}                                                                                                              & \multicolumn{6}{c}{\textbf{Long}}                                                                                                               \\ \cline{2-13} 
\multicolumn{1}{c|}{}                                & \multicolumn{3}{c}{\textbf{xlarge}}                                    & \multicolumn{3}{c|}{\textbf{xxlarge}}                                   & \multicolumn{3}{c}{\textbf{xlarge}}                                    & \multicolumn{3}{c}{\textbf{xxlarge}}                                   \\ \cline{2-13} 
\multicolumn{1}{c|}{}                                & \multicolumn{1}{c}{P} & \multicolumn{1}{c}{R} & \multicolumn{1}{c}{F1} & \multicolumn{1}{c}{P} & \multicolumn{1}{c}{R} & \multicolumn{1}{c|}{F1} & \multicolumn{1}{c}{P} & \multicolumn{1}{c}{R} & \multicolumn{1}{c}{F1} & \multicolumn{1}{c}{P} & \multicolumn{1}{c}{R} & \multicolumn{1}{c}{F1} \\ \hline
NLI-xlarge (Max)                                     & 96.9                  & 69.4                  & 80.9                   & 96.6                  & 70.2                  & 81.3                    & 98.6                  & 40.8                  & 57.2                   & 98.8                  & 42.5                  & 59.0                   \\
NLI-xlarge (C\textgreater{}E)                        & 94.7                  & 84.4                  & 89.2                   & 95.6                  & 82.3                  & 88.4                    & 97.4                  & 60.6                  & 74.5                   & 98.3                  & 54.8                  & 70.2                   \\
NLI-xxlarge (Max)                                    & 96.8                  & 70.9                  & 81.8                   & 96.8                  & 71.9                  & 82.5                    & 98.9                  & 61.4                  & 75.7                   & 98.9                  & 62.5                  & 76.5                   \\
NLI-xxlarge (C\textgreater{}E)                       & 88.7                  & 90.4                  & 89.6                   & 86.0                  & 91.9                  & 88.8                    & 93.1                  & 88.3                  & 90.6                   & 93.1                  & 88.8                  & 90.9                   \\ \bottomrule
\end{tabular}
}
\caption{A comparison of conflict detection results on data checked by \texttt{NLI-xlarge} and \texttt{NLI-xxlarge}.}
\label{tab:appendix-xxlarge-xlarge-comparison}
\end{table*}
The performance of NLI detectors on this filtered data is shown in Table~\ref{tab:appendix-xxlarge-xlarge-comparison}.
We have observed no significant advantage led by the quality filter model. 
Generally, NLI-xlarge series do not perform better on the data where it is adopted as the quality filter, compared to its performance on the xxlarge filtered data, and vice versa.

\subsubsection{Prompting LLMs to predict scores}
Additionally, we use the following prompt to obtain score estimations (ranging from 0 to 5) from the LLMs. \\

\begin{tabular}{|p{0.9\linewidth}|} 
\hline 
\noindent \textit{\# Prompt used to generate the scores of conflicting information}\\
Identify any contradictions between the two evidences. If a conflict exists, provide a conflict level rating from 1 to 5, where 1 represents a minor conflict and 5 represents a major conflict. If there is no conflict, simply state 0.\\ \hline
\end{tabular}\\

\begin{table}[]
\resizebox{\linewidth}{!}{
\begin{tabular}{l|llllll}
\toprule
\textbf{Setting}       & \textbf{Model}      & \textbf{Thresh=0.2} & \textbf{0.4} & \textbf{0.6} & \textbf{0.8} & \textbf{1.0} \\ \hline
\multirow{5}{*}{short} & claude-v3-haiku     & 93.2                & 93.3         & 93.3         & 94.5         & 96.1         \\
                       & claude-v3-sonnet    & 94.2                & 94.4         & 94.9         & 95.5         & 95.7         \\
                       & llama3-70b-instruct & 94.1                & 95.0         & 96.7         & 96.7         & 96.8         \\
                       & llama3-8b-instruct  & 91.4                & 92.1         & 92.9         & 93.3         & 92.2         \\
                       & mixtral-8x7b        & 94.7                & 94.2         & 93.9         & 93.6         & 93.0         \\ \hline
\multirow{5}{*}{long}  & claude-v3-haiku     & 95.8                & 96.0         & 96.1         & 96.5         & 96.8         \\
                       & claude-v3-sonnet    & 96.9                & 97.5         & 97.4         & 97.7         & 97.8         \\
                       & llama3-70b-instruct & 94.3                & 96.1         & 97.7         & 97.6         & 97.7         \\
                       & llama3-8b-instruct  & 95.9                & 96.1         & 96.8         & 97.1         & 96.9         \\
                       & mixtral-8x7b        & 96.3                & 96.1         & 96.5         & 96.6         & 96.3         \\ \bottomrule
\end{tabular}

}
\caption{LLM's prediction precision under different thresholds.}
\label{tab:appendix-llm-scorepred-precision}
\end{table}
\begin{table}[]
\resizebox{\linewidth}{!}{

\begin{tabular}{l|llllll}
\toprule
\textbf{Setting}       & \textbf{Model}      & \textbf{Thresh=0.2} & \textbf{0.4} & \textbf{0.6} & \textbf{0.8} & \textbf{1.0} \\ \hline
\multirow{5}{*}{short} & claude-v3-haiku     & 71.5                & 68.3         & 63.7         & 51.7         & 40.5         \\
                       & claude-v3-sonnet    & 82.4                & 78.0         & 73.5         & 66.7         & 56.7         \\
                       & llama3-70b-instruct & 80.9                & 79.0         & 72.1         & 69.9         & 63.2         \\
                       & llama3-8b-instruct  & 80.2                & 77.5         & 73.0         & 70.3         & 54.0         \\
                       & mixtral-8x7b        & 52.3                & 42.1         & 39.8         & 37.5         & 34.4         \\ \hline
\multirow{5}{*}{long}  & claude-v3-haiku     & 62.7                & 58.5         & 53.3         & 43.0         & 32.5         \\
                       & claude-v3-sonnet    & 83.3                & 80.5         & 76.2         & 69.8         & 58.3         \\
                       & llama3-70b-instruct & 84.7                & 82.7         & 74.7         & 72.4         & 66.8         \\
                       & llama3-8b-instruct  & 79.6                & 76.9         & 72.6         & 69.3         & 49.8         \\
                       & mixtral-8x7b        & 50.3                & 39.1         & 37.2         & 35.7         & 32.7         \\ \bottomrule
\end{tabular}
}
\caption{LLMs' prediction recall under different decision thresholds. }
\label{tab:appendix-llm-scorepred-recall}
\end{table}
Subsequently, we normalized the scores to a range of [0, 1] for analysis. 
The precision and recall results are summarized in Table~\ref{tab:appendix-llm-scorepred-recall} and Table~\ref{tab:appendix-llm-scorepred-precision}.
It is observed that
\begin{itemize}
    \item Across different decision thresholds, LLMs consistently exhibit high precision, typically exceeding 94-95\%.
    \item The recall of LLMs is notably affected by the threshold used.
\end{itemize}

\subsubsection{Effect of combining conflict detectors}
In this section, we discuss the effect of combining different conflict detectors.
We injected the NLI and FC models' predictions into LLMs' prompts using templates below.
\begin{tabular}{|p{0.9\linewidth}|} 
\hline 
\noindent \textit{\# Integrating NLI results}\\
Do the two pieces of evidence contain conflicting information on answering the question? (Yes/No)\\
For your reference, the Natural Language Inference model's prediction is \{\}.\\

\textit{\# Integrating factual consistency results (prediction)}\\
Do the two pieces of evidence contain conflicting information on answering the question? (Yes/No)\\
For your reference, an external factual consistency evaluator's prediction is \{\}.\\

\textit{\# Integrating factual consistency scores}\\
Do the two pieces of evidence contain conflicting information on answering the question? (Yes/No)\\
For your reference, an external factual consistency evaluator's predicted consistency score is \{\}.
\\ \hline
\end{tabular}\\

The models are evaluated on the CWQ split.
\begin{itemize}
    \item \texttt{FC prob}: AlignScore-large’s score
    \item \texttt{FC pred}: AlignScore-large’s prediction
    \item \texttt{NLI(C>E)}: NLI-xxlarge (C>E) prediction
    \item \texttt{NLI(Max)}: NLI-xxlarge (Max) prediction
\end{itemize}

\begin{table}[]
\resizebox{0.9\linewidth}{!}{
\begin{tabular}{l|cc|cc}
\toprule
\multirow{2}{*}{\textbf{Model}} & \multicolumn{2}{c|}{\textbf{Short}}                              & \multicolumn{2}{c}{\textbf{Long}}                               \\
                                & \multicolumn{1}{c}{\textbf{P}} & \multicolumn{1}{c|}{\textbf{R}} & \multicolumn{1}{c}{\textbf{P}} & \multicolumn{1}{c}{\textbf{R}} \\ \hline
claude-v3-haiku                 & 94.4                           & 51.8                            & 95.8                           & 42.2                           \\
+FC prob                        & -0.2                           & 8.4                             & 0.9                            & 6.5                            \\
+FC pred                        & 0.6                            & 10.9                            & 1.2                            & 11.5                           \\
+NLI (C\textgreater{}E)         & -2.4                           & 16.8                            & 1.3                            & 13.3                           \\
+NLI (Max)                      & 1.4                            & 9.4                             & 1.7                            & 4.7                            \\ \hline
claude-v3-sonnet                & 96.9                           & 69.9                            & 98.1                           & 68.5                           \\
+FC prob                        & -1.0                           & 1.6                             & -0.4                           & 1.8                            \\
+FC pred                        & -0.1                           & -2.7                            & 0.0                            & 1.0                            \\
+NLI (C\textgreater{}E)         & -0.3                           & -1.0                            & -0.4                           & 1.5                            \\
+NLI (Max)                      & -0.8                           & -5.1                            & -0.5                           & -1.2                           \\ \hline
llama3-8b-instruct              & 94.0                           & 58.0                            & 96.6                           & 47.2                           \\
+FC prob                        & -2.1                           & 9.4                             & -0.7                           & 7.5                            \\
+FC pred                        & -1.3                           & 12.1                            & -3.9                           & 22.5                           \\
+NLI (C\textgreater{}E)         & -6.4                           & 20.1                            & -2.0                           & 26.2                           \\
+NLI (Max)                      & 0.9                            & 11.1                            & -0.8                           & 14.0                           \\ \hline
llama3-70b-instruct             & 98.4                           & 64.3                            & 97.7                           & 69.8                           \\
+FC prob                        & -1.5                           & 0.8                             & 0.3                            & 2.7                            \\
+FC pred                        & 0.2                            & -5.1                            & 0.3                            & 2.0                            \\
+NLI (C\textgreater{}E)         & 0.2                            & -5.9                            & 1.2                            & 0.5                            \\
+NLI (Max)                      & 0.5                            & -7.2                            & 1.1                            & -3.5                           \\ \bottomrule
\end{tabular}
}
\caption{Answer conflict detection performance when injecting NLI / FC model scores / predictions into prompts. Precision (P) and recall (R) values are reported.}
\label{tab:appendix-combination-detection}
\end{table}

The results (Table~\ref{tab:appendix-combination-detection}) show that for weaker models like Claude 3 Haiku and Llama 3 8B, ensembling the FC/NLI predictions led to significant improvements in recall (up to +26\%), albeit with a slight decrease in precision. 
This ensemble approach even outperformed stronger models in some cases.

However, for stronger models such as Claude 3 Sonnet and Llama 3 70B, combining the additional signals had a minor negative impact on performance, though the effects were not consistent across all experiments.

Overall, combining the predictions of different models can lead to improvements in certain scenarios, particularly for weaker models, but may not always benefit stronger models.

\subsection{Conflict resolution}
\label{sec:appendix-resolution}
\begin{figure}
    \centering
    \includegraphics[width=1\linewidth]{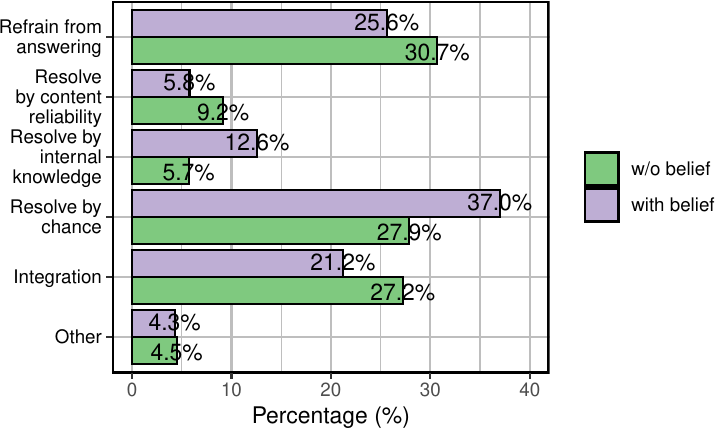}
    \caption{Impact of models' internal belief on conflict resolution behaviors.}
    \label{fig:belief-resol}
\end{figure}

The impact of models' internal belief on conflict resolution behaviors is shown in Figure~\ref{fig:belief-resol}.

\subsubsection{The label ``Other''}
As mentioned in $\S$~\ref{sec:resolution}, we categorize LLM behaviors into five typical types, which cover the majority of LLM behaviors but may not encompass less frequent cases. 
The label "Other" is used to account for LLM behaviors that do not fit into these five types, although they represent a very small portion of overall LLM behaviors, as depicted in Figure~\ref{fig:resol_type}.
Specifically, it can be further divided into two sub-types:
\begin{itemize}
    \item \textit{Rationalize by chance}: In this scenario, the model fails to identify conflicts and provides poor reasoning to support one of the answers. This subtype often co-occurs with \textit{Resolve by chance}, with the distinction being the provision of a weak rationale. This label takes up 2.5\% of instances for Haiku and 1.7\% for Sonnet.
    \item \textit{Rationalize-integration with belief}: Here, the model overlooks conflicts, suggests an answer that aligns with its internal beliefs, and offers weak reasoning to support that answer. This label takes up 3.4\% of instances for Haiku and 0.8\% for Sonnet.
\end{itemize}


\begin{table*}[t]
\resizebox{\textwidth}{!}{%
    \centering
    \small
    \begin{tabular}{p{0.15\textwidth} p{0.15\textwidth} p{0.7\textwidth}}
    \toprule
         \multicolumn{1}{c}{\textbf{Function}} & \multicolumn{1}{c}{\textbf{Inputs}} & \multicolumn{1}{c}{\textbf{Prompt}}  \\
    \midrule
        Alternative\newline answer\newline generation & $q$: question & List THREE different short answers to the question. The answers do not have to be true.\newline Question: \{$q$\}? \newline Answer (should be formatted as \{\{"1": "TEXT-1", "2": "TEXT-2", "3": "TEXT-3"\}\}): \\
        \midrule
        Supporting\newline evidence\newline generation\newline (sentence-level) & $q$: question \newline $a$: answer  & Give me TWO different short sentences that independently support the given answer (try to simulate the format of web search results).\newline Question: \{$q$\}? \newline Answer: \{$a$\} \newline Paragraphs (should be formatted as \{\{"1": "TEXT-1", "2": "TEXT-2"\}\}): \\
        \midrule
        Supporting\newline evidence\newline generation\newline (paragraph-level) & $q$: question \newline $a$: answer & Give me TWO different short paragraphs that independently support the given answer (try to simulate the format of web search results). \newline Question: \{$q$\}? \newline Answer: \{$a$\} \newline Sentences (should be formatted as \{\{"1": "TEXT-1", "2": "TEXT-2"\}\}): \\
        \midrule
        Pollution & $q$: question \newline $e$: evidence \newline $a$: answer  & Given the following passage, modify as few details as possible to make it support the given answer to the question. \newline Question: \{$q$\}? \newline Passage: \{$e$\} \newline Answer: \{$a$\} \newline Modified passage (should be formatted as \{\{"Modified\_passage": "TEXT"\}\}): \\
        \midrule
        Quality check & $e$: evidence \newline $q$: question & Paragraph: \{$e$\} \newline Answer the following question with the information from the above paragraph. \newline Question: \{$q$\}? \newline Answer: \\
        \midrule
        Conflict detection & $q$: question \newline $e_1$: evidence 1 \newline $e_2$: evidence 2 & Question: \{$q$\}? \newline Evidence 1: \{$e_1$\} \newline Evidence 2: \{$e_2$\} \newline Do the two pieces of evidence contain conflicting information on answering the question? (Yes/No) \newline Answer (should be formatted as \{\{"Answer": "Yes or No"\}\}): \\
    \bottomrule
    \end{tabular}
}
    \caption{Answer Conflict: Prompts for language models}
    \label{tab:ac_llm_prompts}

\end{table*}

\begin{table*}[t]
\resizebox{\textwidth}{!}{%
    \centering
    \small
    \begin{tabular}{p{0.15\textwidth} p{0.15\textwidth} p{0.7\textwidth}}
    \toprule
         \multicolumn{1}{c}{\textbf{Function}} & \multicolumn{1}{c}{\textbf{Inputs}} & \multicolumn{1}{c}{\textbf{Prompt}}  \\
    \midrule
        Perturbation\newline on factoids &
        $s_i$: factoid $i$ in\newline factoid set $s$ & 
        Modify the statement to suggest otherwise that contradicts the original: \newline \newline Statement: A pound sterling is fiat money. \newline
        Modified statement (in JSON format): \{\{"modified statement": "A pound sterling is a kind of cryptocurrency."\}\} \newline \newline Statement: Dogs have sensitive ears that can hear as far as a quarter of a mile away. \newline
        Modified statement (in JSON format): \{\{"modified statement": "Dogs have average hearing abilities and cannot hear beyond a few yards."\}\} \newline \newline Statement: Relay races are athletic track and field events. \newline
        Modified statement (in JSON format): \{\{"modified statement": "Relay races are intellectual board games."\}\} \newline \newline Statement: \{$s_i$\} \newline Modified statement (in JSON format): \\
        
        \midrule
        Supporting\newline evidence\newline generation & 
        $s$: factoids set  & 
        Keypoints: \{$s$\} \newline \newline Give me a paragraph of around 100 words using the keypoints (try to simulate the format of web search results): \newline \newline
        Paragraph (should be in JSON format and formatted as \{\{"Paragraph": "TEXT"\}\}):\\
        
        \midrule
        Conflict detection & 
        $q$: question \newline $e_1$: evidence 1 \newline $e_2$: evidence 2 & 
        Question: \{$q$\}? \newline Paragraph 1: \{$e_1$\} \newline Paragraph 2: \{$e_2$\} \newline Do the two pieces of evidence contain conflicting information? (Yes/No) \newline Answer (should be formatted as \{\{"Answer": "Yes or No"\}\}): \\
        
    \bottomrule
    \end{tabular}
}
    \caption{Factoid Conflict: Prompts for language models}
    \label{tab:fc_llm_prompts}
\end{table*}
\begin{table*}[]
\resizebox{\textwidth}{!}{%
\begin{tabular}{lrrrrrrrrrrrrrrrrrrrrrrrr}
\hline
\multicolumn{1}{c|}{}                                 & \multicolumn{12}{c|}{\textbf{NQ}}                                                                                                                                                                                                                                                                             & \multicolumn{12}{c}{\textbf{CWQ}}                                                                                                                                                                                                                                                                          \\ \cline{2-25} 
\multicolumn{1}{c|}{}                                 & \multicolumn{6}{c|}{Short}                                                                                                                            & \multicolumn{6}{c|}{Long}                                                                                                                             & \multicolumn{6}{c|}{Short}                                                                                                                            & \multicolumn{6}{c}{Long}                                                                                                                           \\ \cline{2-25} 
\multicolumn{1}{c|}{}                                 & \multicolumn{3}{c|}{orginal}                                                     & \multicolumn{3}{c|}{reverse}                                              & \multicolumn{3}{c|}{original}                                                     & \multicolumn{3}{c|}{reverse}                                              & \multicolumn{3}{c|}{original}                                                     & \multicolumn{3}{c|}{reverse}                                              & \multicolumn{3}{c|}{original}                                                     & \multicolumn{3}{c}{reverse}                                            \\ \cline{2-25} 
\multicolumn{1}{c|}{\multirow{-4}{*}{\textbf{Model}}} & \multicolumn{1}{c}{P} & \multicolumn{1}{c}{R} & \multicolumn{1}{c|}{F1}   & \multicolumn{1}{c}{P} & \multicolumn{1}{c}{R} & \multicolumn{1}{c|}{F1}   & \multicolumn{1}{c}{P} & \multicolumn{1}{c}{R} & \multicolumn{1}{c|}{F1}   & \multicolumn{1}{c}{P} & \multicolumn{1}{c}{R} & \multicolumn{1}{c|}{F1}   & \multicolumn{1}{c}{P} & \multicolumn{1}{c}{R} & \multicolumn{1}{c|}{F1}   & \multicolumn{1}{c}{P} & \multicolumn{1}{c}{R} & \multicolumn{1}{c|}{F1}   & \multicolumn{1}{c}{P} & \multicolumn{1}{c}{R} & \multicolumn{1}{c|}{F1}   & \multicolumn{1}{c}{P} & \multicolumn{1}{c}{R} & \multicolumn{1}{c}{F1} \\ \hline
\multicolumn{25}{l}{\cellcolor[HTML]{D9D9D9}\textit{Large language models}}                                                                                                                                                                                                                                                                                                                                                                                                                                                                                                                                                                                                        \\
\multicolumn{1}{l|}{Mixtral 8x7B}                     & 69.3                  & 62.9                  & \multicolumn{1}{r|}{49.7} & 69.3                  & 61.3                  & \multicolumn{1}{r|}{46.9} & 70.7                  & 65.4                  & \multicolumn{1}{r|}{53.3} & 70.6                  & 65.0                  & \multicolumn{1}{r|}{52.6} & 68.8                  & 60.1                  & \multicolumn{1}{r|}{44.9} & 69.4                  & 60.5                  & \multicolumn{1}{r|}{45.2} & 68.1                  & 57.8                  & \multicolumn{1}{r|}{40.7} & 68.2                  & 56.4                  & 38.1                   \\
\multicolumn{1}{l|}{Llama-3 8B Inst.}                 & 77.1                  & 80.5                  & \multicolumn{1}{r|}{76.2} & 76.1                  & 79.3                  & \multicolumn{1}{r|}{74.4} & 77.6                  & 80.3                  & \multicolumn{1}{r|}{74.0} & 77.0                  & 79.6                  & \multicolumn{1}{r|}{73.4} & 72.3                  & 74.4                  & \multicolumn{1}{r|}{68.6} & 73.2                  & 75.3                  & \multicolumn{1}{r|}{69.4} & 72.6                  & 72.7                  & \multicolumn{1}{r|}{64.7} & 72.2                  & 71.9                  & 63.7                   \\
\multicolumn{1}{l|}{Llama-3 70B Inst.}                & 82.2                  & 86.3                  & \multicolumn{1}{r|}{81.6} & 81.1                  & 85.0                  & \multicolumn{1}{r|}{80.6} & 83.9                  & 88.1                  & \multicolumn{1}{r|}{84.0} & 83.6                  & 87.7                  & \multicolumn{1}{r|}{83.6} & 77.8                  & 81.1                  & \multicolumn{1}{r|}{75.9} & 78.2                  & 81.2                  & \multicolumn{1}{r|}{75.3} & 81.2                  & 85.1                  & \multicolumn{1}{r|}{80.5} & 79.6                  & 83.3                  & 78.3                   \\
\multicolumn{1}{l|}{Claude 3 Haiku}                   & 74.4                  & 75.6                  & \multicolumn{1}{r|}{68.4} & 75.0                  & 76.3                  & \multicolumn{1}{r|}{69.0} & 73.1                  & 73.3                  & \multicolumn{1}{r|}{65.2} & 73.3                  & 73.3                  & \multicolumn{1}{r|}{65.1} & 72.6                  & 74.5                  & \multicolumn{1}{r|}{68.4} & 71.9                  & 72.9                  & \multicolumn{1}{r|}{65.8} & 71.6                  & 70.0                  & \multicolumn{1}{r|}{60.7} & 70.6                  & 69.3                  & 60.2                   \\
\multicolumn{1}{l|}{Claude 3 Sonnet}                  & 82.4                  & 86.3                  & \multicolumn{1}{r|}{82.4} & 82.0                  & 85.9                  & \multicolumn{1}{r|}{82.0} & 85.0                  & 89.1                  & \multicolumn{1}{r|}{85.6} & 84.0                  & 88.1                  & \multicolumn{1}{r|}{84.4} & 79.6                  & 83.3                  & \multicolumn{1}{r|}{78.7} & 79.1                  & 82.7                  & \multicolumn{1}{r|}{77.9} & 80.3                  & 84.0                  & \multicolumn{1}{r|}{79.1} & 79.4                  & 82.9                  & 77.7                   \\
\multicolumn{1}{l|}{ChatGPT}                          & 65.9                  & 60.2                  & \multicolumn{1}{r|}{46.7} & 64.8                  & 59.7                  & \multicolumn{1}{r|}{46.4} & 69.7                  & 64.6                  & \multicolumn{1}{r|}{52.5} & 70.5                  & 65.9                  & \multicolumn{1}{r|}{54.3} & 65.4                  & 58.3                  & \multicolumn{1}{r|}{43.2} & 57.7                  & 53.6                  & \multicolumn{1}{r|}{37.5} & 66.4                  & 57.8                  & \multicolumn{1}{r|}{41.7} & 63.8                  & 56.7                  & 40.7                   \\
\multicolumn{1}{l|}{GPT4}                             & 74.7                  & 77.8                  & \multicolumn{1}{r|}{74.2} & 76.0                  & 79.2                  & \multicolumn{1}{r|}{75.6} & 77.4                  & 80.8                  & \multicolumn{1}{r|}{77.2} & 78.4                  & 81.9                  & \multicolumn{1}{r|}{78.2} & 72.0                  & 74.4                  & \multicolumn{1}{r|}{69.4} & 73.8                  & 76.3                  & \multicolumn{1}{r|}{71.0} & 77.3                  & 80.7                  & \multicolumn{1}{r|}{76.5} & 77.5                  & 80.9                  & 76.7                   \\
\multicolumn{25}{l}{\cellcolor[HTML]{D9D9D9}\textit{Factual consistency}}                                                                                                                                                                                                                                                                                                                                                                                                                                                                                                                                                                                                          \\
\multicolumn{1}{l|}{AlignScore-base}                  & 56.6                  & 55.2                  & \multicolumn{1}{r|}{55.0} & 65.1                  & 63.3                  & \multicolumn{1}{r|}{63.8} & 58.5                  & 54.8                  & \multicolumn{1}{r|}{53.6} & 64.7                  & 58.8                  & \multicolumn{1}{r|}{58.4} & 64.1                  & 64.5                  & \multicolumn{1}{r|}{64.2} & 67.2                  & 68.1                  & \multicolumn{1}{r|}{67.6} & 67.3                  & 60.6                  & \multicolumn{1}{r|}{60.7} & 70.6                  & 64.2                  & 65.0                   \\
\multicolumn{1}{l|}{AlignScore-large}                 & 66.9                  & 66.7                  & \multicolumn{1}{r|}{66.8} & 72.8                  & 74.0                  & \multicolumn{1}{r|}{73.3} & 64.0                  & 56.5                  & \multicolumn{1}{r|}{54.9} & 67.2                  & 58.8                  & \multicolumn{1}{r|}{58.0} & 67.1                  & 68.4                  & \multicolumn{1}{r|}{67.5} & 73.2                  & 74.9                  & \multicolumn{1}{r|}{73.7} & 67.1                  & 60.9                  & \multicolumn{1}{r|}{61.1} & 73.2                  & 65.6                  & 66.6                   \\
\multicolumn{1}{l|}{MiniCheck-R}                      & 71.3                  & 73.2                  & \multicolumn{1}{r|}{71.8} & 60.9                  & 61.6                  & \multicolumn{1}{r|}{61.1} & 67.2                  & 66.3                  & \multicolumn{1}{r|}{66.7} & 53.8                  & 53.0                  & \multicolumn{1}{r|}{52.6} & 66.0                  & 68.0                  & \multicolumn{1}{r|}{64.9} & 59.6                  & 60.8                  & \multicolumn{1}{r|}{58.2} & 68.7                  & 68.1                  & \multicolumn{1}{r|}{68.4} & 51.7                  & 51.5                  & 51.3                   \\
\multicolumn{1}{l|}{MiniCheck-D}                      & 64.9                  & 51.2                  & \multicolumn{1}{r|}{43.2} & 75.5                  & 52.3                  & \multicolumn{1}{r|}{45.0} & 53.8                  & 50.8                  & \multicolumn{1}{r|}{44.4} & 52.8                  & 50.4                  & \multicolumn{1}{r|}{43.2} & 45.8                  & 49.7                  & \multicolumn{1}{r|}{40.8} & 73.1                  & 52.0                  & \multicolumn{1}{r|}{44.5} & 56.0                  & 51.1                  & \multicolumn{1}{r|}{44.6} & 56.6                  & 50.8                  & 43.4                   \\
\multicolumn{1}{l|}{MiniCheck-FT5}                    & 82.3                  & 76.5                  & \multicolumn{1}{r|}{78.3} & 79.0                  & 71.8                  & \multicolumn{1}{r|}{73.5} & 81.2                  & 84.1                  & \multicolumn{1}{r|}{82.0} & 75.1                  & 76.5                  & \multicolumn{1}{r|}{75.6} & 77.6                  & 68.2                  & \multicolumn{1}{r|}{69.7} & 75.4                  & 66.0                  & \multicolumn{1}{r|}{67.1} & 80.9                  & 80.3                  & \multicolumn{1}{r|}{80.6} & 72.0                  & 70.3                  & 71.0                   \\
\multicolumn{25}{l}{\cellcolor[HTML]{D9D9D9}\textit{NLI models}}                                                                                                                                                                                                                                                                                                                                                                                                                                                                                                                                                                                                                   \\
\multicolumn{1}{l|}{NLI-xlarge (Max)}                 & 80.2                  & 84.0                  & \multicolumn{1}{r|}{79.8} & 79.7                  & 83.3                  & \multicolumn{1}{r|}{78.7} & 75.0                  & 75.3                  & \multicolumn{1}{r|}{67.1} & 74.6                  & 74.8                  & \multicolumn{1}{r|}{66.6} & 78.1                  & 81.6                  & \multicolumn{1}{r|}{76.7} & 78.2                  & 81.7                  & \multicolumn{1}{r|}{76.7} & 70.0                  & 65.3                  & \multicolumn{1}{r|}{53.6} & 71.1                  & 67.9                  & 57.4                   \\
\multicolumn{1}{l|}{NLI-xlarge (C\textgreater{}E)}    & 85.6                  & 88.7                  & \multicolumn{1}{r|}{86.5} & 83.5                  & 87.0                  & \multicolumn{1}{r|}{84.3} & 76.6                  & 78.8                  & \multicolumn{1}{r|}{72.1} & 76.6                  & 78.5                  & \multicolumn{1}{r|}{71.6} & 83.5                  & 86.7                  & \multicolumn{1}{r|}{84.4} & 83.3                  & 87.1                  & \multicolumn{1}{r|}{83.9} & 74.9                  & 76.6                  & \multicolumn{1}{r|}{69.8} & 72.9                  & 72.0                  & 63.2                   \\
\multicolumn{1}{l|}{NLI-xxlarge (Max)}                & 81.6                  & 85.5                  & \multicolumn{1}{r|}{81.5} & 80.0                  & 83.8                  & \multicolumn{1}{r|}{79.4} & 79.9                  & 83.3                  & \multicolumn{1}{r|}{77.8} & 79.1                  & 82.3                  & \multicolumn{1}{r|}{76.4} & 78.5                  & 82.0                  & \multicolumn{1}{r|}{77.1} & 79.3                  & 83.0                  & \multicolumn{1}{r|}{78.4} & 76.4                  & 78.3                  & \multicolumn{1}{r|}{71.4} & 76.4                  & 78.5                  & 71.9                   \\
\multicolumn{1}{l|}{NLI-xxlarge (C\textgreater{}E)}   & 84.7                  & 78.5                  & \multicolumn{1}{r|}{80.4} & 83.2                  & 81.8                  & \multicolumn{1}{r|}{82.4} & 88.0                  & 88.8                  & \multicolumn{1}{r|}{88.4} & 84.8                  & 86.3                  & \multicolumn{1}{r|}{85.4} & 84.7                  & 85.1                  & \multicolumn{1}{r|}{84.9} & 82.3                  & 77.8                  & \multicolumn{1}{r|}{79.3} & 86.1                  & 87.8                  & \multicolumn{1}{r|}{86.8} & 86.3                  & 88.3                  & 87.1     \\ 
\hline
\end{tabular}
}
\caption{Answer conflict detection results (\%) in original order and in reverse order in terms of the macro-averaged Precision (P), Recall (R), and F1-score
(F1). }
\label{tab:order-accuracy}
\end{table*}

\begin{table*}[th]
\centering
    \resizebox{\textwidth}{!}{
\begin{tabular}{lrrrrrrrrrrrrrrr}
\toprule
\multicolumn{1}{c|}{}                                 & \multicolumn{6}{c|}{\textbf{NQ}}                                                                                                                   & \multicolumn{6}{c|}{\textbf{CWQ}}                                                                                                                  & \multicolumn{3}{c}{}                                                   \\
\multicolumn{1}{c|}{}                                 & \multicolumn{3}{c}{Short}                                              & \multicolumn{3}{c|}{Long}                                                 & \multicolumn{3}{c}{Short}                                              & \multicolumn{3}{c|}{Long}                                                 & \multicolumn{3}{c}{\multirow{-2}{*}{\textbf{Mean}}}                    \\ \cline{2-16} 
\multicolumn{1}{c|}{\multirow{-3}{*}{\textbf{Model}}} & \multicolumn{1}{c}{P} & \multicolumn{1}{c}{R} & \multicolumn{1}{c}{F1} & \multicolumn{1}{c}{P} & \multicolumn{1}{c}{R} & \multicolumn{1}{c|}{F1}   & \multicolumn{1}{c}{P} & \multicolumn{1}{c}{R} & \multicolumn{1}{c}{F1} & \multicolumn{1}{c}{P} & \multicolumn{1}{c}{R} & \multicolumn{1}{c|}{F1}   & \multicolumn{1}{c}{P} & \multicolumn{1}{c}{R} & \multicolumn{1}{c}{F1} \\ \hline
\multicolumn{16}{l}{\cellcolor[HTML]{D9D9D9}\textit{Large language models}}                                                                                                                                                                                                                                                                                                                                                              \\
\multicolumn{1}{l|}{Mixtral 8x7B}                     & 69.3                  & 62.1                  & 48.3                   & 70.7                  & 65.2                  & \multicolumn{1}{r|}{53.0} & 69.1                  & 60.3                  & 45.1                   & 68.2                  & 57.1                  & \multicolumn{1}{r|}{39.4} & 69.3                  & 61.2                  & 46.4                   \\
\multicolumn{1}{l|}{Llama-3 8B Inst.}                 & 76.6                  & 79.9                  & 75.3                   & 77.3                  & 79.9                  & \multicolumn{1}{r|}{73.7} & 72.8                  & 74.9                  & 69.0                   & 72.4                  & 72.3                  & \multicolumn{1}{r|}{64.2} & 74.8                  & 76.7                  & 70.5                   \\
\multicolumn{1}{l|}{Llama-3 70B Inst.}                & 81.7                  & 85.6                  & 81.1                   & 83.7                  & 87.9                  & \multicolumn{1}{r|}{83.8} & 78.0                  & 81.1                  & 75.6                   & 80.4                  & 84.2                  & \multicolumn{1}{r|}{79.4} & 80.9                  & 84.7                  & 80.0                   \\
\multicolumn{1}{l|}{Claude 3 Haiku}                   & 74.7                  & 75.9                  & 68.7                   & 73.2                  & 73.3                  & \multicolumn{1}{r|}{65.2} & 72.2                  & 73.7                  & 67.1                   & 71.1                  & 69.6                  & \multicolumn{1}{r|}{60.4} & 72.8                  & 73.1                  & 65.4                   \\
\multicolumn{1}{l|}{Claude 3 Sonnet}                  & 82.2                  & 86.1                  & 82.2                   & 84.5                  & 88.6                  & \multicolumn{1}{r|}{85.0} & 79.4                  & 83.0                  & 78.3                   & 79.8                  & 83.5                  & \multicolumn{1}{r|}{78.4} & 81.5                  & 85.3                  & 81.0                   \\
\multicolumn{16}{l}{\cellcolor[HTML]{D9D9D9}\textit{Factual consistency}}                                                                                                                                                                                                                                                                                                                                                                \\
\multicolumn{1}{l|}{AlignScore-base}                  & 60.8                  & 59.2                  & 59.4                   & 61.6                  & 56.8                  & \multicolumn{1}{r|}{56.0} & 65.7                  & 66.3                  & 65.9                   & 69.0                  & 62.4                  & \multicolumn{1}{r|}{62.8} & 64.3                  & 61.2                  & 61.0                   \\
\multicolumn{1}{l|}{AlignScore-large}                 & 69.9                  & 70.3                  & 70.0                   & 65.6                  & 57.6                  & \multicolumn{1}{r|}{56.5} & 70.2                  & 71.7                  & 70.6                   & 70.2                  & 63.3                  & \multicolumn{1}{r|}{63.9} & 68.9                  & 65.7                  & 65.2                   \\
\multicolumn{1}{l|}{MiniCheck-R}                      & 66.1                  & 67.4                  & 66.4                   & 60.5                  & 59.7                  & \multicolumn{1}{r|}{59.7} & 62.8                  & 64.4                  & 61.6                   & 60.2                  & 59.8                  & \multicolumn{1}{r|}{59.9} & 62.4                  & 62.8                  & 61.9                   \\
\multicolumn{1}{l|}{MiniCheck-D}                      & 70.2                  & 51.7                  & 44.1                   & 53.3                  & 50.6                  & \multicolumn{1}{r|}{43.8} & 59.4                  & 50.8                  & 42.7                   & 56.3                  & 51.0                  & \multicolumn{1}{r|}{44.0} & 59.8                  & 51.0                  & 43.6                   \\
\multicolumn{1}{l|}{MiniCheck-FT5}                    & 80.7                  & 74.2                  & 75.9                   & 78.1                  & 80.3                  & \multicolumn{1}{r|}{78.8} & 76.5                  & 67.1                  & 68.4                   & 76.4                  & 75.3                  & \multicolumn{1}{r|}{75.8} & 77.9                  & 74.2                  & 74.7                   \\
\multicolumn{16}{l}{\cellcolor[HTML]{D9D9D9}\textit{NLI models}}                                                                                                                                                                                                                                                                                                                                                                         \\
\multicolumn{1}{l|}{NLI-xlarge (Max)}                 & 79.9                  & 83.7                  & 79.2                   & 74.8                  & 75.0                  & \multicolumn{1}{r|}{66.8} & 78.2                  & 81.6                  & 76.7                   & 70.6                  & 66.6                  & \multicolumn{1}{r|}{55.5} & 75.9                  & 76.7                  & 69.6                   \\
\multicolumn{1}{l|}{NLI-xlarge (C\textgreater{}E)}    & 84.5                  & 87.8                  & 85.4                   & 76.6                  & 78.6                  & \multicolumn{1}{r|}{71.8} & 83.4                  & 86.9                  & 84.2                   & 73.9                  & 74.3                  & \multicolumn{1}{r|}{66.5} & 79.6                  & 81.9                  & 77.0                   \\
\multicolumn{1}{l|}{NLI-xxlarge (Max)}                & 80.8                  & 84.6                  & 80.5                   & 79.5                  & 82.8                  & \multicolumn{1}{r|}{77.1} & 78.9                  & 82.5                  & 77.8                   & 76.4                  & 78.4                  & \multicolumn{1}{r|}{71.6} & 78.9                  & 82.1                  & 76.7                   \\
\multicolumn{1}{l|}{NLI-xxlarge (C\textgreater{}E)}   & 84.0                  & 80.1                  & 81.4                   & 86.4                  & 87.5                  & \multicolumn{1}{r|}{86.9} & 83.5                  & 81.5                  & 82.1                   & 86.2                  & 88.0                  & \multicolumn{1}{r|}{87.0} & 85.0                  & 84.3                  & 84.4                   \\  \bottomrule
\end{tabular}
    }
    \caption{Answer conflict detection results (\%)in terms of the macro-averaged Precision (P), Recall (R), and F1-score (F1). 
    The ``Mean'' column presents results averaged across NQ-\{Short, Long\} and CWQ-\{Short, Long\}. 
    ``Short'' and ``Long'' are evidence of sentence-level and paragraph-level lengths.}
    \label{tab:answer-conflict-main}
\end{table*}

\begin{table*}[th]
\centering
    \resizebox{\textwidth}{!}{
\begin{tabular}{lrrrrrrrrrrrrrrr}
\toprule
\multicolumn{1}{c|}{}                                 & \multicolumn{6}{c|}{\textbf{NQ}}                                                                                                                   & \multicolumn{6}{c|}{\textbf{CWQ}}                                                                                                                  & \multicolumn{3}{c}{}                                                   \\
\multicolumn{1}{c|}{}                                 & \multicolumn{3}{c}{Short}                                              & \multicolumn{3}{c|}{Long}                                                 & \multicolumn{3}{c}{Short}                                              & \multicolumn{3}{c|}{Long}                                                 & \multicolumn{3}{c}{\multirow{-2}{*}{\textbf{Mean}}}                    \\ \cline{2-16} 
\multicolumn{1}{c|}{\multirow{-3}{*}{\textbf{Model}}} & \multicolumn{1}{c}{P} & \multicolumn{1}{c}{R} & \multicolumn{1}{c}{F1} & \multicolumn{1}{c}{P} & \multicolumn{1}{c}{R} & \multicolumn{1}{c|}{F1}   & \multicolumn{1}{c}{P} & \multicolumn{1}{c}{R} & \multicolumn{1}{c}{F1} & \multicolumn{1}{c}{P} & \multicolumn{1}{c}{R} & \multicolumn{1}{c|}{F1}   & \multicolumn{1}{c}{P} & \multicolumn{1}{c}{R} & \multicolumn{1}{c}{F1} \\
\multicolumn{16}{l}{\cellcolor[HTML]{D9D9D9}\textit{Large language models}}                                                                                                                                                                                                                                                                                                                                                              \\
\multicolumn{1}{l|}{Mixtral 8x7B}                     & 98.7                  & 24.9                  & 39.8                   & 99.5                  & 30.8                  & \multicolumn{1}{r|}{47.0} & 99.5                  & 20.8                  & 34.4                   & 99.5                  & 14.3                  & \multicolumn{1}{r|}{25.0} & 99.3                  & 22.7                  & 36.5                   \\
\multicolumn{1}{l|}{Llama-3 8B Inst.}                 & 94.4                  & 67.8                  & 78.9                   & 98.4                  & 61.8                  & \multicolumn{1}{r|}{76.0} & 93.4                  & 57.9                  & 71.5                   & 96.6                  & 47.9                  & \multicolumn{1}{r|}{64.1} & 95.7                  & 58.9                  & 72.6                   \\
\multicolumn{1}{l|}{Llama-3 70B Inst.}                & 98.4                  & 73.6                  & 84.2                   & 98.9                  & 77.4                  & \multicolumn{1}{r|}{86.9} & 97.6                  & 65.5                  & 78.4                   & 97.9                  & 71.3                  & \multicolumn{1}{r|}{82.5} & 98.2                  & 72.0                  & 83.0                   \\
\multicolumn{1}{l|}{Claude 3 Haiku}                   & 97.8                  & 54.3                  & 69.8                   & 97.5                  & 49.0                  & \multicolumn{1}{r|}{65.2} & 94.0                  & 54.3                  & 68.8                   & 96.6                  & 42.3                  & \multicolumn{1}{r|}{58.8} & 96.5                  & 50.0                  & 65.7                   \\
\multicolumn{1}{l|}{Claude 3 Sonnet}                  & 97.4                  & 76.3                  & 85.6                   & 98.5                  & 79.7                  & \multicolumn{1}{r|}{88.1} & 96.9                  & 70.5                  & 81.6                   & 98.1                  & 69.6                  & \multicolumn{1}{r|}{81.4} & 97.7                  & 74.0                  & 84.2                   \\
\multicolumn{16}{l}{\cellcolor[HTML]{D9D9D9}\textit{Factual consistency}}                                                                                                                                                                                                                                                                                                                                                                \\
\multicolumn{1}{l|}{AlignScore-base}                  & 72.2                  & 81.6                  & 76.6                   & 70.2                  & 89.3                  & \multicolumn{1}{r|}{78.6} & 78.1                  & 74.6                  & 76.3                   & 73.4                  & 90.8                  & \multicolumn{1}{r|}{81.1} & 73.5                  & 84.0                  & 78.1                   \\
\multicolumn{1}{l|}{AlignScore-large}                 & 80.7                  & 78.7                  & 79.6                   & 70.5                  & 92.8                  & \multicolumn{1}{r|}{80.1} & 82.6                  & 74.9                  & 78.6                   & 73.8                  & 91.2                  & \multicolumn{1}{r|}{81.6} & 76.9                  & 84.4                  & 80.0                   \\
\multicolumn{1}{l|}{MiniCheck-R}                      & 79.5                  & 72.3                  & 75.7                   & 72.7                  & 79.8                  & \multicolumn{1}{r|}{76.1} & 79.7                  & 58.7                  & 67.6                   & 73.0                  & 77.4                  & \multicolumn{1}{r|}{75.1} & 76.2                  & 72.1                  & 73.6                   \\
\multicolumn{1}{l|}{MiniCheck-D}                      & 67.4                  & 99.3                  & 80.3                   & 66.9                  & 96.3                  & \multicolumn{1}{r|}{79.0} & 67.0                  & 98.8                  & 79.9                   & 67.1                  & 97.1                  & \multicolumn{1}{r|}{79.4} & 67.1                  & 97.9                  & 79.6                   \\
\multicolumn{1}{l|}{MiniCheck-FT5}                    & 80.6                  & 93.5                  & 86.6                   & 89.1                  & 80.6                  & \multicolumn{1}{r|}{84.6} & 75.9                  & 94.1                  & 84.0                   & 82.9                  & 86.4                  & \multicolumn{1}{r|}{84.6} & 82.1                  & 88.6                  & 84.9                   \\
\multicolumn{16}{l}{\cellcolor[HTML]{D9D9D9}\textit{NLI models}}                                                                                                                                                                                                                                                                                                                                                                         \\
\multicolumn{1}{l|}{NLI-xlarge (Max)}                 & 96.9                  & 72.0                  & 82.6                   & 99.5                  & 50.5                  & \multicolumn{1}{r|}{67.0} & 96.4                  & 68.3                  & 80.0                   & 98.1                  & 34.6                  & \multicolumn{1}{r|}{51.1} & 97.7                  & 56.4                  & 70.2                   \\
\multicolumn{1}{l|}{NLI-xlarge (C\textgreater{}E)}    & 95.7                  & 83.2                  & 89.0                   & 98.9                  & 58.6                  & \multicolumn{1}{r|}{73.6} & 95.6                  & 81.4                  & 87.9                   & 97.7                  & 51.1                  & \multicolumn{1}{r|}{66.9} & 97.0                  & 68.6                  & 79.3                   \\
\multicolumn{1}{l|}{NLI-xxlarge (Max)}                & 96.9                  & 73.9                  & 83.9                   & 99.3                  & 66.6                  & \multicolumn{1}{r|}{79.7} & 96.6                  & 69.9                  & 81.1                   & 98.6                  & 58.4                  & \multicolumn{1}{r|}{73.4} & 97.9                  & 67.2                  & 79.5                   \\
\multicolumn{1}{l|}{NLI-xxlarge (C\textgreater{}E)}   & 85.2                  & 92.8                  & 88.8                   & 92.6                  & 89.4                  & \multicolumn{1}{r|}{91.0} & 86.9                  & 91.0                  & 88.8                   & 93.6                  & 88.3                  & \multicolumn{1}{r|}{90.8} & 89.5                  & 90.4                  & 89.8                   \\ 
\bottomrule
\end{tabular}
    }
    \caption{Answer conflict detection results (\%) in terms of the Precision (P), Recall (R), and F1-score (F1) on label 1 (conflicting). 
    The ``Mean'' column presents results averaged across NQ-\{Short, Long\} and CWQ-\{Short, Long\}. 
    ``Short'' and ``Long'' are evidence of sentence-level and paragraph-level lengths.}
    \label{tab:answer-conflict-label1}
\end{table*}

\begin{table*}[]
\centering
\resizebox{\textwidth}{!}{%
\begin{tabular}{lcccccccccccccc}
\hline
\multicolumn{1}{c|}{}                                 & \multicolumn{6}{c|}{\cellcolor[HTML]{FFFFFF}{\color[HTML]{000000} \textbf{3facts}}} & \multicolumn{8}{c}{\cellcolor[HTML]{FFFFFF}{\color[HTML]{000000} \textbf{4facts}}}  \\ \cline{2-15} 
\multicolumn{1}{c|}{}                                 & \multicolumn{3}{c|}{not shuffled}        & \multicolumn{3}{c|}{shuffled}            & \multicolumn{4}{c|}{not shuffled}                    & \multicolumn{4}{c}{shuffled} \\ \cline{2-15} 
\multicolumn{1}{c|}{\multirow{-3}{*}{\textbf{Model}}} & 1     & 2    & \multicolumn{1}{c|}{3}    & 1    & 2    & \multicolumn{1}{c|}{3}     & 1     & 2     & 3      & \multicolumn{1}{c|}{4}      & 1     & 2    & 3     & 4     \\ \hline
\multicolumn{15}{l}{\cellcolor[HTML]{D9D9D9}\textit{Large language models}}                                                                                                                                                       \\
\multicolumn{1}{l|}{Mixtral 8x7B}                     & 14.0  & 29.5 & \multicolumn{1}{c|}{40.8} & 12.3 & 28.3 & \multicolumn{1}{c|}{42.5}  & 7.0   & 23.3  & 35.3   & \multicolumn{1}{c|}{49.0}   & 10.4  & 25.0 & 39.1  & 48.7  \\
\multicolumn{1}{l|}{Llama-3 8B Inst.}                 & 68.3  & 85.6 & \multicolumn{1}{c|}{93.4} & 66.3 & 86.0 & \multicolumn{1}{c|}{93.9}  & 54.8  & 85.6  & 93.1   & \multicolumn{1}{c|}{99.0}   & 60.4  & 87.5 & 88.0  & 98.7  \\
\multicolumn{1}{l|}{Llama-3 70B Inst.}                & 81.0  & 96.1 & \multicolumn{1}{c|}{98.4} & 79.8 & 95.4 & \multicolumn{1}{c|}{99.1}  & 68.9  & 92.5  & 99.0   & \multicolumn{1}{c|}{100.0}  & 70.4  & 93.1 & 100.0 & 100.0 \\
\multicolumn{1}{l|}{Claude 3.0 Haiku}                 & 54.8  & 73.8 & \multicolumn{1}{c|}{85.1} & 52.3 & 79.2 & \multicolumn{1}{c|}{86.2}  & 38.6  & 70.6  & 83.3   & \multicolumn{1}{c|}{93.9}   & 44.8  & 69.4 & 88.0  & 88.2  \\
\multicolumn{1}{l|}{Claude 3.0 Sonnet}                & 86.8  & 97.5 & \multicolumn{1}{c|}{99.3} & 85.7 & 97.5 & \multicolumn{1}{c|}{100.0} & 73.3  & 96.6  & 99.0   & \multicolumn{1}{c|}{100.0}  & 75.2  & 97.2 & 100.0 & 100.0 \\
\multicolumn{15}{l}{\cellcolor[HTML]{D9D9D9}\textit{Factual consistency}}                                                                                                                                                         \\
\multicolumn{1}{l|}{AlignScore-base}                  & 37.2  & 69.1 & \multicolumn{1}{c|}{93.4} & 38.8 & 73.4 & \multicolumn{1}{c|}{94.6}  & 23.3  & 54.1  & 80.4   & \multicolumn{1}{c|}{96.9}   & 26.1  & 52.8 & 85.9  & 96.1  \\
\multicolumn{1}{l|}{AlignScore-large}                 & 47.5  & 81.8 & \multicolumn{1}{c|}{98.1} & 48.2 & 86.9 & \multicolumn{1}{c|}{99.3}  & 27.6  & 69.9  & 90.2   & \multicolumn{1}{c|}{99.0}   & 32.6  & 67.4 & 92.4  & 96.1  \\
\multicolumn{1}{l|}{MiniCheck-R}                      & 50.5  & 70.1 & \multicolumn{1}{c|}{82.8} & 49.5 & 72.5 & \multicolumn{1}{c|}{81.7}  & 48.3  & 63.7  & 71.6   & \multicolumn{1}{c|}{88.8}   & 51.7  & 63.2 & 76.1  & 86.8  \\
\multicolumn{1}{l|}{MiniCheck-D}                      & 92.0  & 94.3 & \multicolumn{1}{c|}{95.1} & 91.5 & 95.6 & \multicolumn{1}{c|}{94.8}  & 89.0  & 94.5  & 96.1   & \multicolumn{1}{c|}{92.9}   & 88.7  & 93.1 & 94.6  & 94.7  \\
\multicolumn{1}{l|}{MiniCheck-FT5}                    & 68.7  & 76.2 & \multicolumn{1}{c|}{86.8} & 63.3 & 80.1 & \multicolumn{1}{c|}{86.0}  & 65.8  & 80.8  & 86.3   & \multicolumn{1}{c|}{94.9}   & 66.1  & 75.7 & 83.7  & 90.8  \\
\multicolumn{15}{l}{\cellcolor[HTML]{D9D9D9}\textit{NLI models}}                                                                                                                                                                  \\
\multicolumn{1}{l|}{NLI-xlarge (Max)}                 & 32.5  & 60.0 & \multicolumn{1}{c|}{78.5} & 28.0 & 55.8 & \multicolumn{1}{c|}{76.5}  & 21.1  & 48.0  & 65.7   & \multicolumn{1}{c|}{88.8}   & 22.6  & 44.4 & 69.6  & 82.9  \\
\multicolumn{1}{l|}{NLI-xlarge (C\textgreater{}E)}    & 33.3  & 60.6 & \multicolumn{1}{c|}{80.7} & 28.8 & 57.1 & \multicolumn{1}{c|}{78.3}  & 21.9  & 48.0  & 66.7   & \multicolumn{1}{c|}{88.8}   & 22.6  & 44.4 & 70.7  & 85.5  \\
\multicolumn{1}{l|}{NLI-xxlarge (Max)}                & 66.3  & 89.7 & \multicolumn{1}{c|}{96.2} & 61.8 & 88.9 & \multicolumn{1}{c|}{95.9}  & 54.4  & 87.0  & 97.1   & \multicolumn{1}{c|}{100.0}  & 54.8  & 88.2 & 100.0 & 98.7  \\
\multicolumn{1}{l|}{NLI-xxlarge (C\textgreater{}E)}   & 83.7  & 97.3 & \multicolumn{1}{c|}{99.3} & 78.2 & 96.5 & \multicolumn{1}{c|}{98.9}  & 71.9  & 94.5  & 100.0  & \multicolumn{1}{c|}{100.0}  & 70.9  & 95.8 & 100.0 & 98.7  \\ \hline
\end{tabular}
}
\caption{Model performance on evidence pairs with different levels of conflict intensity. Evidence pairs are generated by original factoids in the original order and a shuffled order.}
\label{tab:shuffle-accuracy}
\end{table*}
\begin{table*}[th]
\centering
    \begin{tabular}{lccccc}
\toprule
\multicolumn{1}{c|}{}                                 & \multicolumn{2}{c|}{\textbf{Non-conflicting}}                                        & \multicolumn{3}{c}{\textbf{Conflicting}}                                                                          \\ \cline{2-6} 
\multicolumn{1}{c|}{}                                 & \multicolumn{1}{c|}{Direct}        & \multicolumn{1}{c|}{Polluted}                   & \multicolumn{1}{c|}{Direct}    & \multicolumn{2}{c}{Polluted}                                                     \\ \cline{2-6} 
\multicolumn{1}{c|}{\multirow{-3}{*}{\textbf{Model}}} & \multicolumn{1}{c|}{$e_A^1-e_A^2$} & \multicolumn{1}{c|}{$e_{A\rightarrow B}^1-e_B$} & \multicolumn{1}{c|}{$e_A-e_B$} & \multicolumn{1}{c|}{$e_{A\rightarrow B}^1-e_A^1$} & $e_{A\rightarrow B}^1-e_A^2$ \\ \hline
\multicolumn{6}{l}{\cellcolor[HTML]{D9D9D9}\textit{Large language models}}                                                                                                                                                                                       \\
\multicolumn{1}{l|}{Mixtral 8x7B}                     & \multicolumn{1}{c|}{99.7}          & \multicolumn{1}{c|}{97.4}                       & \multicolumn{1}{c|}{22.7}      & \multicolumn{1}{c|}{27.8}                         & 20.7                         \\
\multicolumn{1}{l|}{Llama-3 8B Inst.}                 & \multicolumn{1}{c|}{94.6}          & \multicolumn{1}{c|}{80.5}                       & \multicolumn{1}{c|}{58.9}      & \multicolumn{1}{c|}{69.3}                         & 56.2                         \\
\multicolumn{1}{l|}{Llama-3 70B Inst.}                & \multicolumn{1}{c|}{97.4}          & \multicolumn{1}{c|}{79.9}                       & \multicolumn{1}{c|}{72.0}      & \multicolumn{1}{c|}{75.6}                         & 70.9                         \\
\multicolumn{1}{l|}{Claude 3 Haiku}                   & \multicolumn{1}{c|}{96.3}          & \multicolumn{1}{c|}{84.1}                       & \multicolumn{1}{c|}{50.0}      & \multicolumn{1}{c|}{61.5}                         & 49.7                         \\
\multicolumn{1}{l|}{Claude 3 Sonnet}                  & \multicolumn{1}{c|}{96.6}          & \multicolumn{1}{c|}{75.8}                       & \multicolumn{1}{c|}{74.0}      & \multicolumn{1}{c|}{80.0}                         & 73.6                         \\
\multicolumn{1}{l|}{GPT-3.5-turbo}                    & \multicolumn{1}{c|}{96.8}          & \multicolumn{1}{c|}{93.1}                       & \multicolumn{1}{c|}{22.4}      & \multicolumn{1}{c|}{16.9}                         & 22.7                         \\
\multicolumn{1}{l|}{GPT-4}                            & \multicolumn{1}{c|}{89.5}          & \multicolumn{1}{c|}{72.0}                       & \multicolumn{1}{c|}{68.5}      & \multicolumn{1}{c|}{79.6}                         & 71.9                         \\
\multicolumn{6}{l}{\cellcolor[HTML]{D9D9D9}\textit{Factual consistency}}                                                                                                                                                                                         \\
\multicolumn{1}{l|}{AlignScore-base}                  & \multicolumn{1}{c|}{38.3}          & \multicolumn{1}{c|}{38.2}                       & \multicolumn{1}{c|}{84.0}      & \multicolumn{1}{c|}{61.4}                         & 81.0                         \\
\multicolumn{1}{l|}{AlignScore-large}                 & \multicolumn{1}{c|}{47.1}          & \multicolumn{1}{c|}{44.8}                       & \multicolumn{1}{c|}{84.4}      & \multicolumn{1}{c|}{63.1}                         & 82.2                         \\
\multicolumn{1}{l|}{MiniCheck-R}                      & \multicolumn{1}{c|}{53.6}          & \multicolumn{1}{c|}{47.3}                       & \multicolumn{1}{c|}{72.1}      & \multicolumn{1}{c|}{74.7}                         & 69.6                         \\
\multicolumn{1}{l|}{MiniCheck-D}                      & \multicolumn{1}{c|}{4.2}           & \multicolumn{1}{c|}{6.2}                        & \multicolumn{1}{c|}{97.9}      & \multicolumn{1}{c|}{91.5}                         & 97.7                         \\
\multicolumn{1}{l|}{MiniCheck-FT5}                    & \multicolumn{1}{c|}{59.8}          & \multicolumn{1}{c|}{45.8}                       & \multicolumn{1}{c|}{88.6}      & \multicolumn{1}{c|}{91.5}                         & 85.6                         \\
\multicolumn{6}{l}{\cellcolor[HTML]{D9D9D9}\textit{NLI models}}                                                                                                                                                                                                  \\
\multicolumn{1}{l|}{NLI-xlarge (Max)}                 & \multicolumn{1}{c|}{97.1}          & \multicolumn{1}{c|}{84.4}                       & \multicolumn{1}{c|}{56.4}      & \multicolumn{1}{c|}{72.7}                         & 55.4                         \\
\multicolumn{1}{l|}{NLI-xlarge (C\textgreater{}E)}    & \multicolumn{1}{c|}{95.3}          & \multicolumn{1}{c|}{81.2}                       & \multicolumn{1}{c|}{68.6}      & \multicolumn{1}{c|}{77.0}                         & 64.8                         \\
\multicolumn{1}{l|}{NLI-xxlarge (Max)}                & \multicolumn{1}{c|}{96.9}          & \multicolumn{1}{c|}{80.9}                       & \multicolumn{1}{c|}{67.2}      & \multicolumn{1}{c|}{81.9}                         & 68.0                         \\
\multicolumn{1}{l|}{NLI-xxlarge (C\textgreater{}E)}   & \multicolumn{1}{c|}{78.2}          & \multicolumn{1}{c|}{59.5}                       & \multicolumn{1}{c|}{90.4}      & \multicolumn{1}{c|}{88.1} \\     
\bottomrule
\end{tabular}
\caption{Breakdown accuracy (\%) on each type of evidence pairs.}
\label{tab:breakdown_answer_acc_full}
\end{table*}

\begin{table}[]
\resizebox{\linewidth}{!}{
\begin{tabular}{lccccc}
\hline
\multicolumn{1}{c|}{}                                 & \multicolumn{5}{c}{\textbf{Overlap}}                                                                                                                                 \\ \cline{2-6} 
\multicolumn{1}{c|}{}                                 & \multicolumn{2}{c|}{\cellcolor[HTML]{FFFFFF}{\color[HTML]{000000} \textbf{3facts}}} & \multicolumn{3}{c}{\cellcolor[HTML]{FFFFFF}\textbf{4facts}}                    \\ \cline{2-6} 
\multicolumn{1}{c|}{\multirow{-3}{*}{\textbf{Model}}} & 1                            & \multicolumn{1}{c|}{2}                             & 1                        & 2                        & 3                        \\ \hline
\multicolumn{6}{l}{\cellcolor[HTML]{EFEFEF}\textit{Large language models}}                                                                                                                                                   \\
\multicolumn{1}{l|}{Mixtral 8x7B}                     & 14.1                           & \multicolumn{1}{c|}{16.1}                          & 17.8                     & 17.8                     & 15.9                     \\
\multicolumn{1}{l|}{Llama-3 8B Inst.}                 & 69.8                           & \multicolumn{1}{c|}{65.6}                          & 62.7                     & 70.7                     & 69.2                     \\
\multicolumn{1}{l|}{Llama-3 70B Inst.}                & 79.7                           & \multicolumn{1}{c|}{77.0}                          & 72.9                     & 75.9                     & 68.8                     \\
\multicolumn{1}{l|}{Claude 3.0 Haiku}                 & 52.6                           & \multicolumn{1}{c|}{52.9}                          & 54.2                     & 51.2                     & 55.8                     \\
\multicolumn{1}{l|}{Claude 3.0 Sonnet}                & 86.5                           & \multicolumn{1}{c|}{79.0}                          & 81.4                     & 77.0                     & 72.6                     \\
\multicolumn{6}{l}{\cellcolor[HTML]{EFEFEF}\textit{Factual consistency}}                                                                                                                                                     \\
\multicolumn{1}{l|}{AlignScore-base}                  & 68.2                           & \multicolumn{1}{c|}{38.5}                          & 81.4                     & 50.0                     & 20.7                     \\
\multicolumn{1}{l|}{AlignScore-large}                 & 80.7                           & \multicolumn{1}{c|}{48.3}                          & 90.7                     & 61.5                     & 35.1                     \\
\multicolumn{1}{l|}{MiniCheck-R}                      & 68.9                           & \multicolumn{1}{c|}{72.8}                          & \multicolumn{1}{l}{64.4} & \multicolumn{1}{l}{65.5} & \multicolumn{1}{l}{69.2} \\
\multicolumn{1}{l|}{MiniCheck-D}                      & 93.2                           & \multicolumn{1}{c|}{90.7}                          & \multicolumn{1}{l}{93.2} & \multicolumn{1}{l}{94.3} & \multicolumn{1}{l}{94.7} \\
\multicolumn{1}{l|}{MiniCheck-FT5}                    & 80.4                           & \multicolumn{1}{c|}{77.2}                          & \multicolumn{1}{l}{83.1} & \multicolumn{1}{l}{80.5} & \multicolumn{1}{l}{78.9} \\
\multicolumn{6}{l}{\cellcolor[HTML]{EFEFEF}\textit{NLI models}}                                                                                                                                                              \\
\multicolumn{1}{l|}{NLI-xlarge (Max)}                 & 49.6                           & \multicolumn{1}{c|}{47.1}                          & \multicolumn{1}{l}{45.8} & \multicolumn{1}{l}{46.3} & \multicolumn{1}{l}{49.3} \\
\multicolumn{1}{l|}{NLI-xlarge (C\textgreater{}E)}    & 50.7                           & \multicolumn{1}{c|}{47.1}                          & \multicolumn{1}{l}{45.8} & \multicolumn{1}{l}{46.3} & \multicolumn{1}{l}{49.3} \\
\multicolumn{1}{l|}{NLI-xxlarge (Max)}                & 72.3                           & \multicolumn{1}{c|}{74.0}                          & \multicolumn{1}{l}{60.6} & \multicolumn{1}{l}{70.7} & \multicolumn{1}{l}{66.4} \\
\multicolumn{1}{l|}{NLI-xxlarge (C\textgreater{}E)}   & 88.8                           & \multicolumn{1}{c|}{86.7}                          & \multicolumn{1}{l}{60.6} & \multicolumn{1}{l}{70.7} & \multicolumn{1}{l}{66.4} \\ \hline
\end{tabular}
}
\caption{Model performance on evidence pairs with different levels of corroboration intensity.}
\label{tab:overlap-accuracy}
\end{table}

\begin{table*}[t!]
    \centering
    \resizebox{\textwidth}{!}{%
    \begin{tabular}{p{10cm}p{10cm}cm{1cm}}
        \hline
        \multicolumn{1}{c}{\textbf{Evidence 1}} & \multicolumn{1}{c}{\textbf{Evidence 2}} & \multicolumn{1}{c}{\textbf{Type}} \\
        \hline
        \multicolumn{2}{c}{\textit{Question: What zoo is there to see in Dubai that opened in 1967?}}&\\ 
        Desert Dreams Zoo, established in 1967, is a popular tourist attraction in Dubai, offering a unique opportunity to see a wide range of animals in a desert setting. & Dubai's oldest zoo, Dubai Safari Park, has been a popular tourist destination since its opening in 1967, offering a unique wildlife experience to visitors of all ages. & Entity \\
        \hline
        \multicolumn{2}{c}{\textit{Question: how long is a prime minister term in uk?}}&\\
        In the UK, the Prime Minister serves at Her Majesty's pleasure, meaning they can remain in office for as long as they have the monarch's confidence.&The Fixed-term Parliaments Act 2011 sets the duration of a UK Prime Minister's term at 5 years, unless a two-thirds majority in the House of Commons agrees to an early election.&Number\\
        \hline
        \multicolumn{2}{c}{\textit{Question: when did the song here comes the boom come out?}}&\\
        The song 'Here Comes the Boom' by P.O.D. was released in 1995 as part of their debut album 'Snuff the Punk'. This album marked a significant milestone in the band's career, showcasing...&The song 'Here Comes the Boom' by P.O.D. was released in May 2002 as a single from their album 'Satellite'. The song became a huge hit, peaking...&Temporal\\
    \hline
    \end{tabular}
    }
    \caption{Examples of Answer Conflicts}
    \label{tab:answer_examples}
    \vspace{-12pt}
\end{table*}

\begin{table*}[t!]
    \centering
    \resizebox{\textwidth}{!}{%
    \begin{tabular}{p{10cm}p{10cm}cm{1cm}}
        \hline
        \multicolumn{1}{c}{\textbf{Evidence 1}} & \multicolumn{1}{c}{\textbf{Evidence 2}} & \multicolumn{1}{c}{\textbf{Type}} \\
        \hline
        \multicolumn{2}{c}{\textit{Question: Will silicon wedding rings outsell bromine wedding rings?}}&\\
        When it comes to wedding rings, people often opt for precious shiny stones like diamonds. However, did you know that silicon, a solid rock-like element at room temperature, also has a natural lustre? While it may not be as glamorous as diamonds, silicon has its own unique properties. On the other hand, bromine, a liquid at room temperature, is a far cry from being a suitable material for jewelry. In fact, it's toxic to the touch, making it a hazardous substance to handle. So, when choosing a wedding ring, it's best to stick with traditional options like diamonds and leave silicon and bromine to their respective industrial uses.&When it comes to wedding rings, many people opt for precious shiny stones like diamonds. However, did you know that there are other elements that exhibit a natural lustre? Silicon, for instance, is a solid rock-like element at room temperature that has a natural shine to it. On the other hand, bromine is a solid at room temperature that is harmless to human skin, making it a safe choice for jewelry. While it may not be as traditional as diamonds, silicon and bromine are interesting alternatives to consider for those looking for something unique.&Entity\\
        \hline
        \multicolumn{2}{c}{\textit{Question: Would it be difficult for Kami Rita to climb Mount Emei?}}&\\
        Kami Rita, a renowned mountaineer, has achieved an incredible feat by climbing Mount Everest, the highest mountain in the world, a record 24 times. Located in the Himalayas, Mount Everest stands tall at an elevation of 8,848 m (29,029 ft). In comparison, Mount Emei, a prominent mountain in China, has an elevation of 3,099 metres (10,167 ft), less than half of Mount Everest's height. Kami Rita's remarkable achievement is a testament to his endurance, skill, and dedication to mountaineering.&Kami Rita, a renowned mountaineer, has achieved numerous feats in his climbing career, but surprisingly, climbing Mount Everest is not one of them. Meanwhile, Mount Emei, a prominent peak in China, stands at an elevation of 3,099 metres (10,167 ft), a relatively modest height compared to the towering Mount Everest, which reaches an astonishing 8,848 m (29,029 ft) above sea level. Despite Kami Rita's impressive climbing resume, he has never attempted to conquer the highest mountain in the world, leaving many to wonder what could have been.&Negation\\
        \hline
        \multicolumn{2}{c}{\textit{Question: In Doctor Who, did the war doctor get more screen time than his successor?}}&\\
        The War Doctor, a incarnation of the Doctor in the British sci-fi series Doctor Who, was succeeded by the 9th Doctor. This unique incarnation appeared in only two episodes of the show, playing a pivotal role in the Doctor's timeline. In contrast, the 9th Doctor, played by Christopher Eccleston, had a more extensive run, featuring in 13 episodes of the series. Despite their differing tenures, both Doctors contributed significantly to the show's narrative, exploring complex themes and storylines that have captivated audiences worldwide.&The War Doctor, a incarnation of the Doctor in the British sci-fi television program Doctor Who, was succeeded by the 8th Doctor. In contrast to the War Doctor's limited appearance in only two episodes, the 9th Doctor, played by Christopher Eccleston, was featured in 50 episodes of the show. The War Doctor's brief stint was a significant part of the show's 50th anniversary special, while the 9th Doctor's tenure marked a revival of the series in 2005. Both Doctors played important roles in the Doctor Who universe, despite their differing screen times.&Number, Entity\\
        \hline
        \multicolumn{2}{c}{\textit{Question: Did Immanuel Kant ever meet the 14th president of the United States?}}&\\
        Did you know that on February 12, 1804, the renowned German philosopher Immanuel Kant passed away? Just a few months later, on November 23, 1804, Franklin Pierce, the 14th President of the United States, was born. Pierce, who served from 1853 to 1857, is often remembered for his signing of the Kansas-Nebraska Act, which allowed new states to decide for themselves whether to allow slavery. Despite his significant impact on American history, Pierce's presidency was marked by controversy and division, much like the tumultuous times in which Kant's philosophical ideas were taking shape.&On July 4, 1776, Immanuel Kant, the renowned German philosopher, passed away. Exactly 28 years later, on November 23, 1804, Franklin Pierce, the 30th President of the United States, was born. Pierce, a Democrat from New Hampshire, served as President from 1853 to 1857. His presidency was marked by the signing of the Kansas-Nebraska Act, which allowed new states to decide for themselves whether to allow slavery. Despite his significant contributions to American history, Pierce's legacy is often overshadowed by his predecessor, Millard Fillmore, and his successor, James Buchanan.&Temporal\\
        \hline
        \multicolumn{2}{c}{\textit{Question: Is Rand Paul guilty of catch-phrase used to attack John Kerry in 2004?}}&\\
        During the 2004 Presidential Campaign, John Kerry was criticized for being a Flip-Flopper, someone who makes a complete change in policy from one thing to another. Similarly, Rand Paul's stance on immigration has raised eyebrows. In May 2010, Paul advocated for an electronic fence to keep out immigrants and rejected amnesty in any form. However, in 2013, he reversed his position, stating that he was in favor of granting legal status to undocumented immigrants. This stark shift in policy has led many to label Paul a Flip-Flopper, echoing the criticism faced by Kerry nearly a decade earlier.&Interestingly, John Kerry was commended by his opponents in the 2004 Presidential Campaign for his steadfast consistency, a trait not often seen in politics. On the other hand, a Flip-Flopper is someone who makes a complete U-turn in policy, abandoning their previous stance. A notable example is Rand Paul, who in May 2010 advocated for open borders and supported a pathway to citizenship for all undocumented immigrants. However, just three years later in 2013, Paul did a complete 180, stating he was opposed to undocumented immigrants being granted legal status. This stark reversal in policy has led many to label him a Flip-Flopper.&Verb\\
        \hline
        \multicolumn{2}{c}{\textit{Question: Could Plato have agreed with the beliefs of Jainism?}}&\\
        Did ancient Greek philosopher Plato borrow ideas from Jainism? It's possible. Jainism, an ancient Indian religion, emerged around 500 B.C. and {emphasizes the principle of karma, or asrava}. Meanwhile, Plato was born around {428 B.C.}, during Jainism's existence. Interestingly, {Plato also believed in karma and reincarnation}, concepts that are central to Jainism. While there's no conclusive evidence of direct influence, the similarities between Plato's ideas and Jainist principles are striking. Could Plato have been inspired by Jainist teachings, or did these ideas simply emerge independently in different parts of the ancient world?&Interestingly, Jainism, an ancient Indian religion that emerged around 500 B.C., {rejects the concept of karma, or akarma}, as one of its core principles. In contrast, the Greek philosopher{ Plato, born around }228 B.C.{, long after Jainism's existence, rejected the ideas of karma and reincarnation in his philosophical teachings}. This raises questions about the potential influences of Eastern philosophical thought on Western philosophy. Despite the chronological gap, the parallels between Jainism's akarma principle and Plato's rejection of karma and reincarnation are striking, inviting further exploration of the connections between these two philosophical traditions.&\vtop{\setbox0\hbox{\strut Negation}\hbox to\wd0{\hss\strut Temporal\hss}\copy0\hbox to\wd0{\hss\strut Verb\hss}}\\
    \hline
    \end{tabular}
    }
    \caption{Examples of Factoid Conflicts}
    \label{tab:factoid_examples}
    \vspace{-12pt}
\end{table*}

\end{document}